\documentclass{article} %
\usepackage{iclr2027_conference,times}
\iclrfinalcopy

\usepackage{amsmath,amsfonts,bm}

\def\eqref#1{equation~\ref{#1}}

\def\1{\bm{1}}

\DeclareMathAlphabet{\mathsfit}{\encodingdefault}{\sfdefault}{m}{sl}
\SetMathAlphabet{\mathsfit}{bold}{\encodingdefault}{\sfdefault}{bx}{n}

\newcommand{\given}{\,|\,}

\ifx\theorem\undefined
\newtheorem{theorem}{Theorem}

\fi

\newcommand{\xb}{\bm{x}}

\renewcommand{\paragraph}{\xhdr}

\usepackage{hyperref}
\usepackage{url}
\usepackage{booktabs}
\usepackage{amsmath}
\usepackage{dsfont}
\usepackage{amssymb}
\usepackage{pifont}
\usepackage{graphicx}
\usepackage{wrapfig}
\usepackage{enumitem}
\usepackage{placeins}
\usepackage{subfigure}
\usepackage{etoc}
\usepackage[nameinlink]{cleveref}
\colorlet{darkgreen}{green!45!black}
\hypersetup{
  colorlinks,
  citecolor=darkgreen}

\usepackage{snaptodo}
\usepackage{longtable} 

\snaptodoset{block rise=0.5em}
\snaptodoset{margin block/.style={font=\tiny}}
\definecolor{tocblue}{HTML}{1A2F8A}

\usepackage{xcolor}
\definecolor{SurvICL}{HTML}{3B65CD}
\definecolor{SurvPFN}{HTML}{FFB38D}
\definecolor{SurvivalPFN}{HTML}{D6876F}
\definecolor{Survival-IC}{HTML}{AD5A50}
\definecolor{TFM4Survival}{HTML}{842E32}

\definecolor{RSF}{HTML}{D8D2C4}
\definecolor{SurvivalBoost}{HTML}{4E4E4E}

\definecolor{TabSurv}{HTML}{A3B6E0}
\definecolor{SumoNet}{HTML}{8192BB}
\definecolor{DSM}{HTML}{5F6E96}
\definecolor{DeepSurv}{HTML}{3C4970}
\definecolor{DeepHit}{HTML}{1A254B}

\newcommand{\bbR}{\mathbb{R}}

\def\n_datasets{{103}}

\title{SurvICL: A foundation model for in-context survival analysis}

\author{Julie Alberge \thanks{\texttt{julie.alberge@inria.fr}} \\
Inria de Saclay\\
France\\
\And
Linus Bleistein \\
EPFL\\
Switzerland\\
\And
Judith Abécassis  \\
Inria de Saclay\\
France\\
\AND
Marine Le Morvan  \\
Inria de Saclay\\
France\\
\And
Gaël Varoquaux  \\
Inria de Saclay, :probabl. \\
France\\
\And
David Holzmüller  \\
Inria de Saclay\\
France\\
}

\begin{document}

\etocdepthtag.toc{mtchapter}  %

\maketitle
\lhead{Preprint.}

\begin{abstract}
    Tabular foundation models (TFMs) now outperform gradient-boosted trees in classification and regression. Yet in survival analysis, which predicts when an event occurs from censored data, often on small clinical cohorts, we find that no existing TFM beats the half-century-old Cox model. We address this issue by introducing SurvICL, a TFM that predicts survival curves in-context. SurvICL is pretrained on purely synthetic datasets generated from a novel flexible proportional-odds prior that outperforms other prior types in our ablation study.
    Unlike most other survival models, SurvICL is trained with a plain quantile regression loss, which is possible because its synthetic pre-training data generator can provide uncensored event times.
    Additionally, we introduce the largest survival benchmark to date spanning 103 datasets. On the 63 datasets held out from all design decisions, SurvICL outperforms 20 tree-based, deep, and foundation-model baselines without any tuning: it wins 82.9\% of pairwise comparisons in integrated Brier score, against 71.3\% for the runner-up, the Cox model, and is the first model to rank above Cox on small datasets. 

\end{abstract}

\section{Introduction}

Many decisions hinge on \emph{when} an event will happen: how long will a patient survive after a cancer diagnosis? When will a customer churn? When will a machine fail? Survival analysis answers such questions by predicting, for an individual with features $\xb \in \mathbb{R}^d$, the \emph{survival function} $S(t\given\xb) = \mathbb{P}(T^\star > t \given \xb)$, that is, the probability that the event has not occurred by time $t$ \citep{kleinbaum_survival_2012,machin2006survival,wiegrebe2024deep,jeanselme2024review}. What sets survival analysis apart from regression, aside from strictly positive predictions, is \emph{censoring}. A patient who leaves a clinical study alive after three years does not reveal their time of death, only that it exceeds three years. Discarding such partial observations biases predictions, and treating them as observed events is plainly wrong: a survival model must learn from both observed and censored events. In addition, survival predictions feed decisions (such as choosing a treatment given a five-year survival probability) and must therefore be \emph{calibrated} probabilities rather than mere rankings of individuals \citep{goldstein2020x,lillelund2025stop,alberge2026calibration}.

Survival analysis is often a small-data problem. Many time-to-event datasets come from clinical studies, where observing events requires following patients for years, and cohorts of a few hundred individuals are the norm \citep{indrayan2021importance,o2025systematic}: in all the data we collect for this work, more than 70\% of the datasets have fewer than 1{,}000 samples. In this regime, the flexibility of modern machine learning is a liability. Half a century after its introduction, the Cox proportional hazards model \citep{cox1972regression} remains the workhorse of practitioners, and for good reason: on the small datasets of our benchmark, none of the gradient-boosted trees, deep survival networks, or existing survival foundation models outperforms it.

\begin{figure}[t]
    \centering
    \includegraphics[width=\linewidth]{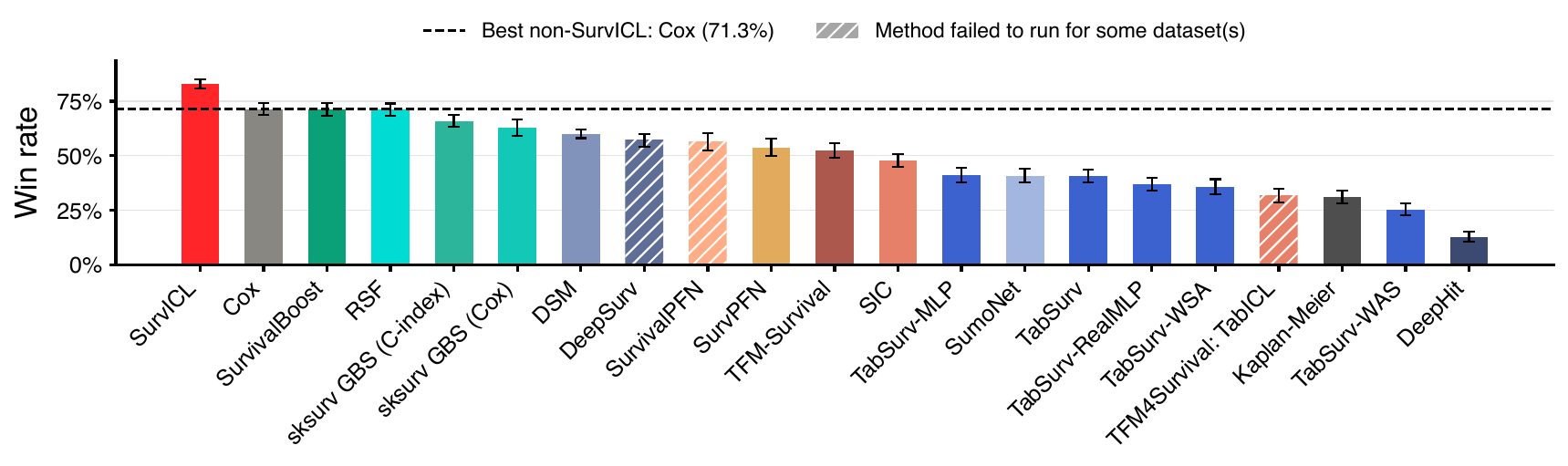}%
    \vspace*{-1.2em}

    \caption{\textbf{SurvICL is the best survival model on 63 held-out datasets.} Pairwise win rate in integrated Brier score (IBS): the fraction of (dataset, competitor) pairs on which a method achieves a lower IBS. SurvICL wins 82.9\% of its comparisons, against 71.3\% for the runner-up, the classical Cox model. All existing survival TFMs trail behind the Cox model.}
    \label{fig:win_rate}
\end{figure}

Tabular foundation models (TFMs) seem very well suited to this regime. Pre-trained on many synthetic datasets, models such as TabPFN and TabICL predict in-context and now outperform gradient-boosted trees for classification and regression, in particular on small datasets \citep{hollmann2023tabpfn,hollmann2025tabpfn,grinsztajn2025tabpfn,qu2025tabicl,qu2026tabiclv2,zhang2025mitra}. This recipe has been extended beyond standard supervised learning to causal effect estimation \citep{robertson2025pfn,balazadeh2025causalpfn,ma2025foundation}. Survival analysis has not yet benefited from this progress. Recent survival TFMs either recast survival as classification over a time grid for off-the-shelf TFMs \citep{kim2026tabular}, or are pre-trained on synthetic survival data \citep{seletkov2026survival,bohm2026survpfn,qi2026survivalpfn} but do not beat the Cox model or random survival forests (Figure~\ref{fig:win_rate}).

\textbf{Building a High-Performing TFM for Survival Analysis.} We introduce SurvICL, a tabular foundation model for in-context survival analysis, built on the TabICLv2 backbone \citep{qu2026tabiclv2}. 
Similar to SurvivalPFN \citep{qi2026survivalpfn}, it handles censoring in the data generation rather than the loss.
During pre-training, SurvICL sees censored datasets in its context, but is supervised with the latent, uncensored event times of the simulator: it learns to correct for censoring in-context, so that a plain quantile-regression loss suffices. SurvICL predicts for each individual a continuous quantile function of the event time, from which we can directly infer a monotone survival function.

\textbf{Scaling Evaluation of Survival Models.} We evaluate SurvICL in the largest benchmark to date for survival analysis: 103 real-world datasets from healthcare, genomics, economics and engineering, with up to 52{,}000 samples and 15{,}000 features, against 20 baselines spanning classical, tree-based, deep and foundation models. For comparison, survival papers at machine-learning venues typically report results on 5 to 20 datasets, and the largest evaluations of survival foundation models span 48 to 81 datasets \citep{kim2026tabular,lyu2026survfm,pham2026adaptation,qi2026survivalpfn}. All design choices were made on 40 development datasets, and all results are reported on 63 held-out datasets never used during development.
On these held-out datasets, SurvICL is the best method both in prediction accuracy and in discrimination (Figure~\ref{fig:win_rate}): it wins 82.9\% of pairwise comparisons for the integrated Brier score (IBS), against 71.3\% for the runner-up, the Cox model, and in 77.7\% for the C-index, against 71.6\% for random survival forests (see Figure \ref{fig:winplotcindex}). On small datasets, where no previous method surpasses the Cox model, SurvICL is the first to rank above it. On the largest datasets, gradient boosting still maintains a slight edge. 
Importantly, these gains do not come at the expense of computational efficiency. Across predictive metrics and dataset-size regimes, SurvICL lies
on or close to the performance-runtime Pareto frontier (Appendix \ref{sec:other_results}).
Thus, SurvICL combines strong predictive performance with efficient out-of-the-box inference.
We commit to releasing code and model in an open-source license.

\textbf{Contributions.} Our contributions are threefold:
\begin{enumerate}
[leftmargin=1.2em,itemsep=0.5pt,topsep=1pt]
    \item \textbf{We design a foundation model for survival analysis} that, using in-context learning, outperforms survival-analysis models (classical, tree-based, deep, all three existing survival foundation models as well as an adaptation of a classification foundation model), on integrated Brier score (IBS) and C-index.
    \item \textbf{We identify key ingredients} of in-context survival analysis through extensive ablations: (i)~a \textbf{proportional-odds (PO) prior} with a flexible, multi-modal baseline hazard, which outperforms Cox, accelerated failure time, and existing TFM priors; and (ii)~\textbf{event-only targets}, rather than also learning the censoring distribution.
    \item \textbf{An extensive benchmark} of 103 survival datasets with a strict separation between development and held-out datasets, covering high-dimensional genomics cohorts ($p > n$) and large registries, and comparing 21 methods on four metrics.
\end{enumerate}

\section{Related Work}
\label{sec:related_work}
\textbf{Deep Learning for Survival Analysis.} The toolbox of survival analysis has been considerably extended in the last years to include deep architectures \citep{katzman2018deepsurv,lee2018deephit,kvamme2019time} that can accommodate a wide variety of data and event types, including time series, competing events, images and text \citep{zhu2016deep,thorsen2022discrete,bleistein2024dynamical, wiegrebe2024deep,alberge2024survival,jeanselme2024review}. Our work builds on this research by tailoring deep transformers and training procedures for tabular data, which is the most common form of survival data.     

\textbf{Deep Learning for Tabular Data.} Gradient-based decision tree models have long dominated benchmarks on tabular data \citep{gorishniy2021revisiting,grinsztajn2022tree,ye2024closer}. Tabular foundation models have recently challenged their superiority and sparked a new line of research that focused on designing tailored architectures and training procedures \citep{hollmann2023tabpfn,hollmann2025tabpfn,qu2025tabicl,qu2026tabiclv2,kim2025table,dingtabula,zhang2025mitra}. We build on these architectural advances and adapt them to calibrated survival prediction. Once pre-trained, our method performs in-context inference without fine-tuning or weight updates. 

\textbf{Tabular Foundation Models for Survival Analysis} is a burgeoning field of research. Efforts follow three distinct strategies.
\emph{(i) Reducing survival to a task that generic TFMs already solve.} \citet{kim2026tabular} frame survival prediction as a series of binary classification problems over a time grid, so that any TFM that handles classification can be deployed for survival analysis. \citet{vieyra2026staying} instead plugs an off-the-shelf TFM regressor into a log-normal accelerated failure time model and impute censored times iteratively, while SurvFM \citep{lyu2026survfm} converts censored follow-up into pseudo-observations of the restricted mean survival time (RMST), which are then regressed in-context by unmodified TFMs. These approaches inherit the strengths of generic TFMs, but their backbone has never seen censored data: they either depend on a time discretization, impose a parametric form on the event-time distribution, or only output a summary statistic (the RMST) rather than a full survival function.
\emph{(ii) Adapting generic TFMs with a survival head.} \citet{pham2026survivalaware,pham2026adaptation} train classical survival heads (Cox, DeepHit, multi-task logistic regression) on top of frozen or fine-tuned representations of TabPFN, TabDPT \citep{ma2025tabdpt}, and TabICL, and systematically compare these adaptation interfaces on 74 datasets. They find that zero-shot reformulations work best on small datasets, whereas trained heads take over as the sample size grows. This line of work matches the common practice of fitting shallow survival models on top of deep foundation models, and gives up the pre-training + in-context learning recipe: the head must be fitted, and possibly tuned, on each new dataset.
\emph{(iii) Pre-training on synthetic survival data.} Closest to our work, \citet{seletkov2026survival}, \citet{bohm2026survpfn} and \citet{qi2026survivalpfn} pre-train a tabular foundation model on synthetic survival data, following previous work in the field of tabular foundation models \citep{hollmann2023tabpfn,qu2025tabicl,qu2026tabiclv2}. All three models predict on a discretized time grid (see Table~\ref{tab:comparison}), while SurvICL outputs a continuous quantile function of the event time. Finally, foundation models for time-to-event prediction have also been pre-trained on large real-world electronic health records \citep{steinberg2024motor}; such models are tied to a specific data modality and coding system, whereas we target generic tabular data through a synthetic prior.

\section{Survival Analysis Settings: Censored Targets, Small Samples}
\label{sec:background}
\subsection{Censored Data and Survival Models}

\textbf{Survival Data.} In survival analysis, data comes in the form of i.i.d. samples $\mathcal{D}_n = \{ (\xb_i, t_i, \delta_i) \}_{i=1}^n$. For each instance $i$, $\xb_i \in \mathbb{R}^d$ denotes the feature vector and $t_i >0$ is a positive valued time. Unlike standard supervised learning, the target is subject to \textit{right-censoring}, meaning that their true event time is not necessarily observed but might be censored, in which case we only observe the censoring time $c_i >0$. This proxy time yields weak information on the true event time, since it will necessarily happen \textit{after} censoring. We denote by $t_i^\star$ the possibly unobserved true time of event, such that
\begin{equation*}
    t_i := \min\{t_i^\star,c_i\} \quad\textnormal{and}\quad \delta_i := \mathds{1}_{t_i^\star \leq c_i}.
\end{equation*}

\textbf{Predicting with Survival Models.} Our goal is to approximate the survival function
\begin{align*}
  S(t\given \xb_i) := \mathbb{P}(T^\star > t\given X = \xb_i)
\end{align*}
at any time $t >0$. This stands in contrast to classification or regression, where the model's goal is to output a value $\widehat{y}_i$ which approximates the binary or continuous outcome $y_i$. The survival function can be used to predict the time of event $\hat{t}_i$ for individual $i$ as the first time it crosses the threshold~$0.5$.

\textbf{Hazard-based Models.} Following the seminal work of \citet{cox1972regression}, an extensive family of models parametrized the hazard of the survival process, the time-wise derivative of the event probability:
\begin{flalign*}
\text{Hazard:} &&    h(t\given \xb_i) := \lim\limits_{\varepsilon \to 0} \frac{\mathbb{P}(T^\star \in [t,t+\varepsilon) \given X = \xb_i, T^\star \geq t )}{\varepsilon} \quad = -\partial_t \log S(t\given \xb_i) &&
\end{flalign*}
and trains it via (partial) likelihood maximization \citep{friedman1982piecewise,fornili2013piecewise,katzman2018deepsurv,kvamme2019time,kvamme2021continuous,bleistein2024dynamical}. Once the hazard is learned, it can be used to predict the survival function using numerical integration.  

\textbf{Discrete-Time Models.} On the other side, some models bin time to simplify the computation of the model's likelihood and frame survival analysis as a classification task \citep{yu2011learning,fotso2018deep,lee2018deephit,kim2026tabular}. More precisely, in their approach, time is discretized into $B+1$ bins $([s_k,s_{k+1}[)_{k=0}^B$ for $0 = s_0 < \dots < s_B = +\infty$. The tuple $(\xb_i,t_i,\delta_i)$ is then expanded into a series of points by creating a series of binary outcomes $y_{i,k} := \mathds{1}_{t_i \leq s_k}$ and considering the expanded dataset $\{((\xb_i,s_k),y_{i,k})_k\}_{i=1}^n$. For individuals who experience the event, $B$ points are created, while for censored individuals only the bins prior to the censoring time are added to the expanded dataset. This approach is the dominant approach in survival TFMs (see Table~\ref{tab:comparison}). While flexible and easy to train, unconstrained formulations of these models can output non-monotonic survival functions, requiring post-hoc workarounds such as clipping probabilities \citep{kim2026tabular}. 

\subsection{Design goals for a Survival TFM}

\textbf{Performance on Small Datasets.} Small datasets are ubiquitous in survival analysis~\citep{o2025systematic}, notably because time-to-event data is expensive and time-consuming to collect (to analyze event times that span over $N$ years, one first needs to wait for $N$ years) and because datasets often originate from healthcare~\citep{indrayan2021importance}. State-of-the-art models perform well on large samples ($n \geq 2000$), but struggle on small datasets \citep{zhao2026tackling}.

\textbf{Out-of-the-Box Calibration.} In survival analysis, accurate probability estimates are crucial for decision-making \citep{goldstein2020x,alberge2026calibration}. Yet, models are often trained and evaluated chasing the C-index, which has been criticized as it does not assess calibration and is not a proper metric  \citep[not minimized for true probabilities,][]{lillelund2025stop}. We train SurvICL to provide calibrated predictions without hyperparameter tuning. 
 
\section{SurvICL Architecture: Quantiles Supervised by Latent Times}
\label{sec:architecture}

\textbf{The TabICLv2 Backbone.} Our architecture inherits the major components of TabICLv2 \citep{qu2026tabiclv2}. In particular, we use \textit{repeated feature grouping}, which bags groups of features before tokenizing them and \textit{target-aware embedding}. We adapt the original architecture of \citet{qu2026tabiclv2} to handle triplets as inputs, which correspond to features, censoring indicators and observed times. More details on TabICLv2 and our modifications can be found in Appendix \ref{sec:app:architecture}.

\textbf{Continuous Quantile Head.} To obtain calibrated and accurate predictions, we use a multi-quantile pinball loss following TabICLv2. Given an input $\xb \in \mathbb{R}^d$, our model outputs $\hat{\bm{q}}(\xb) = (\hat{q}_\tau(\xb))_{\tau \in \mathcal{T}} \in [0,+\infty)^{M-1}$, whose entries are the predicted quantiles of the event time at the levels $\mathcal{T} = \{1/M, 2/M,\dots, (M-1)/M\}$. We then compute 
\begin{align*}
    \mathcal{L}(\hat{\bm{q}}(\xb),t^\star) := \frac{1}{M-1} \sum\limits_{\tau \in \mathcal{T}} \ell_\tau\big(\hat{q}_\tau(\xb),t^\star\big) \,\textnormal{ where }\,\ell_\tau(\hat{q},t^\star) := \begin{cases}
\tau\,(t^\star - \hat{q}) & \text{if } t^\star \ge \hat{q} \\[4pt]
(1-\tau)\,(\hat{q} - t^\star) & \text{if } t^\star < \hat{q}
\end{cases}
\end{align*}
where $t^\star$ is the latent event time drawn from the prior (see below). We use $M=1000$, i.e., $999$ quantile levels as in TabICLv2.
To ensure monotonicity, we sort the predicted quantiles except during pre-training. Experiments with other numbers of percentiles are reported in Figure \ref{fig:main_results}. This loss is one of the major design choices that drive the capabilities of our model, and distinguishes it from previous models, which use classification losses or the hybrid loss of \citet{lee2018deephit}.

\textbf{Latent Time Supervision.} The target $t^\star$ of the pinball loss is never observed in real data: it is the latent event time generated by our prior (Section~\ref{sec:pretraining}). During pre-training, the context only contains censored observations $(\xb_i,t_i,\delta_i)$, while the loss is computed on the uncensored $t_i^\star$ of the query points. The model thus learns to infer the event-time distribution from censored data in-context. The survival-specific knowledge lives in the prior, so that the above regression loss is valid: we need neither a censored likelihood, nor inverse-probability-of-censoring weights, nor a discretization of time. SurvivalPFN \citep{qi2026survivalpfn} also supervises with latent times, but over a discretized time grid and with censoring times as an additional target, which we find to be detrimental (Section~\ref{sec:experiments}).

\textbf{Mean-Duration Scaling.} We rescale time, to cater for the wide variety of time range across datasets. Our strategy is to divide survival times by the mean observed duration.

\textbf{Event-Only Targets.} In survival analysis, censoring times are often considered to be hurdles to efficient estimation. Interestingly, \citet{qi2026survivalpfn} use the censored times as an alternative target: their model learns to approximate two distributions (one for censored times and one for event times) by predicting the event and censoring times during pre-training. We ablate this mixed prediction task and show that it degrades the performance of our model.

\begin{figure}
    \centering
    \includegraphics[width=0.9\linewidth]{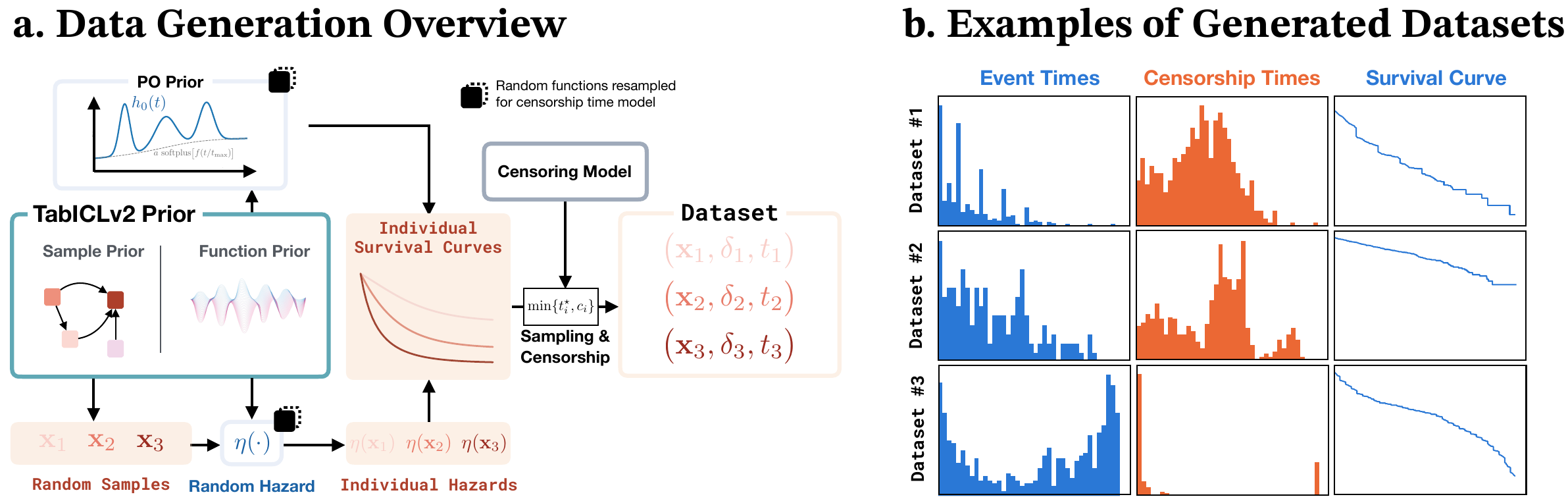}
    \caption{\textbf{(a)} Overview over our synthetic data prior: building on \citet{qu2026tabiclv2}, we construct a pipeline to simulate survival datasets by generating an event time and censoring time distribution. \textbf{(b)} Three example datasets drawn from our prior, which illustrate the high diversity in observed event time and censoring time distributions.}
    \label{fig:prior}
\end{figure}

\section{Pretraining: Synthetic Data Priors with Proportional-Odds}
\label{sec:pretraining}

A key novelty of our work is our synthetic data prior, illustrated in Figure \ref{fig:prior}.

\textbf{Proportional-Odds (PO) Prior.} We start by sampling numerical and categorical features $(\xb_i)_{i=1}^n$ using the structural causal prior of \citet{qu2026tabiclv2}. The number of nodes (from 2 to 32) and features (from 1 to 100) span a wide range to allow for diverse datasets. We then sample survival times from a novel proportional-odds prior. Like the Cox model, it combines a baseline with an individual effect $\eta(\xb_i)$, but on the odds rather than the hazard scale: hazard ratios then vary over time and may vanish in the long run, a common pattern in clinical data that Cox priors cannot generate. For the baseline hazard, we set
\[
            h_0(t)
            = a\,\operatorname{softplus}\!\,\bigl[f(t/t_{\max})\bigr]
            + \sum_{k=1}^{K} A_k
              \exp\!\left(-\frac{(t-\mu_k)^2}{2\sigma_k^2}\right),
        \]
with $a$, $(A_k)_{k=1}^K$, $(\mu_k)_{k=1}^K$ and $(\sigma_k)_{k=1}^K$ positive values, and $t_{\max}>0$. The parameter $K$ is taken randomly between $1$ and $5$. This parametrization adds a random number of positive Gaussian bumps at random times, each with its own width and height, directly on top of the smooth trend. We then integrate this baseline hazard to obtain $S_0(t)=\exp\!\big(-\int_0^t h_0(u)\,du\big)$.
To generate the individual specific part of the hazard function, we apply a random function $\eta(\xb_i)$ to the features as in \citet{qu2026tabiclv2}. We then define the individual-specific survival function as
\begin{align*}
    S(t\given \xb_i) := \frac{1}{\frac{1-S_0(t)}{S_0(t)}\exp(\eta(\xb_i))+1}.
\end{align*}
Setting $\eta(\xb_i)=0$ recovers the baseline hazard, while positive values increase the odds of experiencing the event by any time. This prior draws on a long line of work on mixture models in survival analysis \citep{mclachlan1994role,de1999mixture,li2024bayesian}. We sample event times by inverting the cumulative distribution function, with $U \sim \mathcal{U}([0,1])$:
\begin{align*}
    t_i^\star = H_0^{-1}\Big(\log\big[1+\exp(-\eta(\xb_i))\,(U^{-1}-1)\big]\Big)
\end{align*}
with $H_0(t) := -\log S_0(t)$. We draw $5$ times per individual to speed up convergence.

\textbf{Generating Censoring Times.} We add two sources of censoring to our model. First, in $50\%$ of datasets, we apply a shared end-of-study cutoff to a randomly sampled set of individuals. Secondly, we generate a concurrent censoring time distribution by sampling another pair of baseline and individual specific hazards and combining them as for the event time distribution. We then sample from this second distribution, and set the observed time to be the minimum between the two. Lastly, to introduce tied times, we move some individual times to a common time grid with low probability.    

\textbf{Filtering.} We remove datasets where more than 95\% of samples are censored, as well as \textit{non-predictable} datasets, adapting the filter of \citet{qu2026tabiclv2}: a dataset is non-predictable if predicting the mean survival time achieves a lower IBS (Section~\ref{sec:experiments}) than an ExtraTrees-based survival model \citep{geurts2006extremely}. Every pretraining dataset thus carries exploitable signal.         

\textbf{Training.} During training, we split every generated dataset into a context (train) set $\mathcal{D}_c$ of size $n_c$ and a query (test) set $\mathcal{D}_q$ of size $n_\ell$ with $n=n_c+n_\ell$. We then compute and take the gradients of
\begin{equation*}
    \frac{1}{n_\ell}\sum_{i \in \mathcal{D}_q} \mathcal{L}\big(f_\theta(\xb_i,\mathcal{D}_c),t_i^\star\big),
\end{equation*}
where $f_\theta(\xb_i,\mathcal{D}_c)$ denotes the quantiles predicted from the context $\mathcal{D}_c$; the loss is computed on the query set only. The context size varies during pre-training. 
We adopt the pre-training setup from \cite{qu2026tabiclv2} and only use stage 1 for 200K steps, which takes around 72 hours on two H100 GPUs.

\section{Large Scale Evaluation on 63 Datasets}
\label{sec:experiments}

\subsection{Survival Metrics}

We evaluate our models on four different metrics to overcome the pitfalls of single-metric evaluation \citep{lillelund2025stop}. In the following, $N \geq 1$ denotes the size of the evaluation dataset. All metrics are computed using the \texttt{SurvEval} package \citep{qi2023survivaleval}

\textbf{Integrated Brier Score (IBS).} The IBS is computed by integrating the Brier score at time $t >0$, which is equal for an individual to the difference between the predicted survival curve and the observed binary survival indicator function, over time. It is hence equal to 
\begin{align*}
\frac{1}{N_{test} t_\textnormal{max}}\sum\limits_{i=1}^N\int_0^{t_\textnormal{max}} \textnormal{BS}_i(t)dt \quad\textnormal{ where }\;\; \textnormal{BS}_i(t) := \frac{\widehat S(t\given \xb_i)^2 \mathds{1}_{t_i \leq t, \delta_i = 1}}{\widehat G(t_i)} + \frac{(1-\widehat S(t \given \xb_i))^2\mathds{1}_{t_i > t}}{\widehat G(t)}. 
\end{align*}
Unlike the C-index, the IBS measures the accuracy of predicted probabilities, not just rankings. 

\textbf{C-index at the 50th Quantile.}
We evaluate discrimination using an IPCW concordance index truncated at the median observed training follow-up time $t_{50}=\operatorname{Quantile}_{0.5}\{t_i^{\mathrm{train}}\}$, where $t_i^{\mathrm{train}}$ includes both event and censoring times. For each test individual, we use the predicted risk $r_i=1-\widehat S(t_{50}\given\xb_i)$.
Concordance is computed over comparable pairs with an observed event before $t_{50}$ and weighted by $\widehat G(t_i)^{-2}$, where $\widehat G$ is the Kaplan--Meier estimate of the censoring survival function fitted on the training data. This corresponds to a truncated IPCW concordance estimator.

\textbf{MAE-Hinge.} We use absolute error for event times and a one-sided hinge error for censored times:
\begin{equation*}
    \mathrm{MAE}_{\mathrm{Hinge}}
    = \frac{1}{n}\sum_{i=1}^{n}
    \left[
        \delta_i |t_i-\hat t_i|
        + (1-\delta_i)\max(0,t_i-\hat t_i)
    \right],
\end{equation*}
where $\delta_i$ indicates an observed event and $\hat t_i$ is the
predicted median survival time. For censored individuals, only
predictions earlier than the censoring time are penalized.

\textbf{KM-Calibration.} We assess marginal calibration with the normalized integrated squared error, in~$[0,1]$, between the average predicted survival curve $\frac{1}{N_{test}}\sum_{i} \widehat S(t\given \xb_i)$ and the Kaplan--Meier curve of the test set \citep{chapfuwa2020calibration}.

\subsection{Baselines: From Cox to Foundation Models}

All models are described in more detail in Appendix \ref{sec:app:details_baselines}, we refer to Table \ref{tab:all_baselines} for a synthetic overview.  

\textbf{Shallow models.} We include Cox model \citep{cox1972regression} and KM estimator \citep{kaplan1958nonparametric} as baselines. The latter does not use covariates, and hence any model is expected to outperform it.

\textbf{Deep Networks.} We include five deep networks tailored for survival analysis \citep{lee2018deephit, katzman2018deepsurv,nagpal2021deep,rindt2022survival,kirpichenko2026tabsurv}. These models are fitted on each dataset and use various modeling techniques (from Bayesian prior to a MLP that predicts the survival function at different times). 

\textbf{Tree-Based Models.} Following the seminal work of \citet{ishwaran2008random} on random survival forests (RSF) and \citet{grinsztajn2022tree}, a body of work has adapted tree-based methods to survival analysis. On top of RSF, we evaluate SurvivalBoost \citep{alberge2024survival} and alternative gradient-boosted tree-based methods implemented in \texttt{sksurv} \citep{polsterl2020scikit}. 

\textbf{Tabular Foundation Models (TFMs).} We include four open-weight tabular foundation models in our benchmark: SurvPFN \citep{bohm2026survpfn}, SurvivalPFN \citep{qi2026survivalpfn}, Survival-IC (SIC) \citep{seletkov2026survival} and TFM4Survival \citep{kim2026tabular}.

\subsection{Datasets and Hold-Out Protocol}

\begin{figure}
    \centering
    \includegraphics[width=\linewidth]{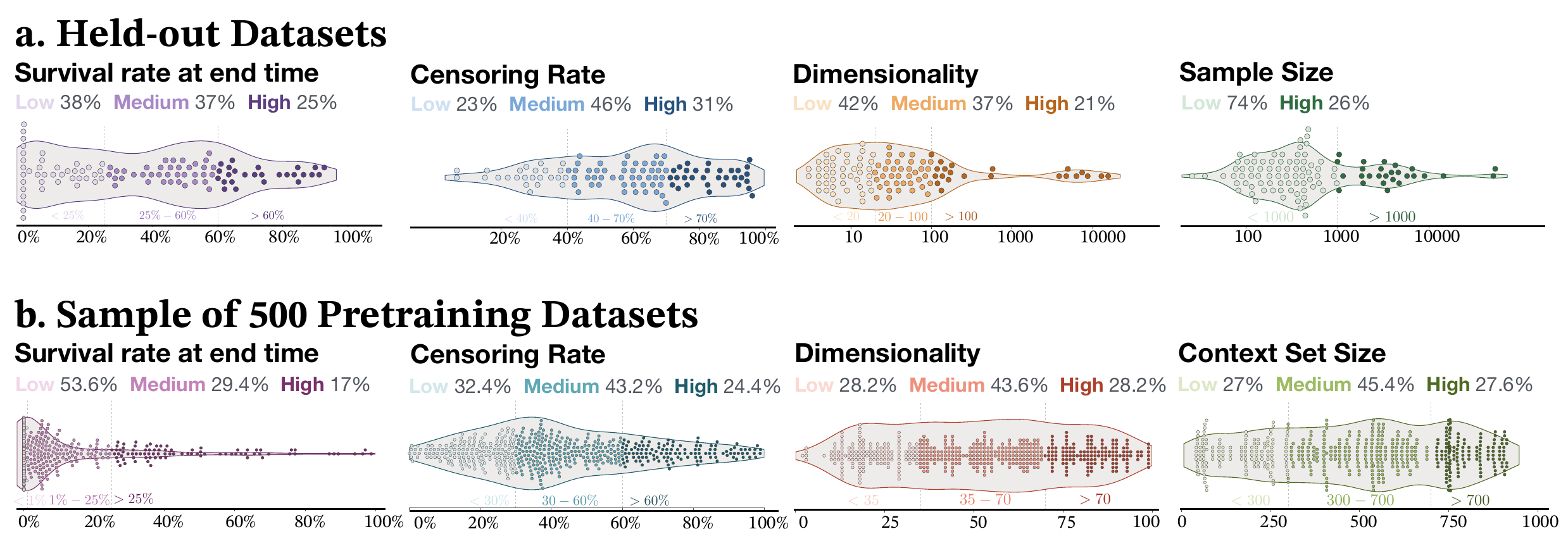}
    \caption{Overview of \textbf{(a)} the held-out datasets and \textbf{(b)} a sample of 500 pre-training datasets. In (b), the sample size is replaced by the context (train) set size (see Section~\ref{sec:pretraining}).}
    \label{fig:datasets_overview}
\end{figure}
In total, we consider 103 datasets with varying properties, see Figure \ref{fig:datasets_overview}. The datasets are partitioned into 40 development datasets used during model development and 63 held-out datasets at random. For development, we source 2 PyCox \citep{kvamme2019time}, 25 SurvSet \citep{drysdale2022}, one from a \href{https://challengedata.ens.fr/challenges/162}{challenge} and 12 cBioPortal \citep{cerami2012cbio,gao2013integrative,de2023analysis} datasets. For held-out datasets, we use 6 from Pycox, 39 from SurvSet and 18 datasets from cBioPortal.

\textbf{Protocol.} 
All design choices of SurvICL (prior, target, and checkpoint) were made using the 40 development datasets; the 63 held-out datasets were kept for the final evaluation. We evaluated each method over ten event-stratified 80/20 train--test splits (test sets capped at 4,096 observations).
Baseline hyperparameters were selected using a budget of up to 50 random search steps per dataset, minimizing the integrated Brier score on an event-stratified 80/20 split within the seed-0 training partition. The selected configuration was reused across all ten evaluation splits, with each estimator refitted on the corresponding training split. Methods without tunable hyperparameters were evaluated directly. Search spaces and method-specific budgets are detailed in Appendix~\ref{sec:app:details_baselines}. Cox regression uses an $\ell_2$ penalty to regularize estimation, including when $p>n$.

SurvICL uses each dataset's training split as context, with at most 4,096 labeled observations selected by event-stratified subsampling, without fine-tuning or dataset-specific hyperparameter tuning. Benchmark results for all methods and metrics are in Appendix~\ref{sec:other_results}. %
Baseline runtimes measure fitting and survival-curve prediction, excluding hyperparameter search and metric computation.

\subsection{Results: SurvICL Is Best Overall}

\begin{figure}[t]
    \begin{minipage}[c]{.58\linewidth}
        \textbf{(a)} IBS\\[-.1em]
        \includegraphics[width=\linewidth]{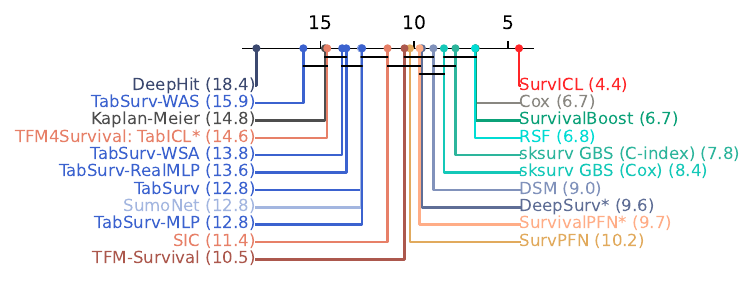}\\[.3em]
        \textbf{(b)} C-index at $t_{50}$\\[-.1em]
        \includegraphics[width=\linewidth]{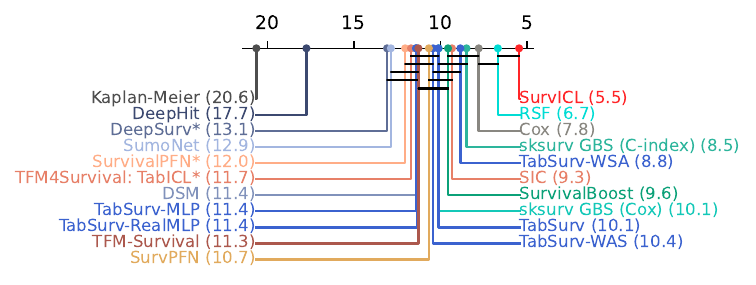}
    \end{minipage}\hfill
    \begin{minipage}[c]{.38\linewidth}
        \includegraphics[width=\linewidth]{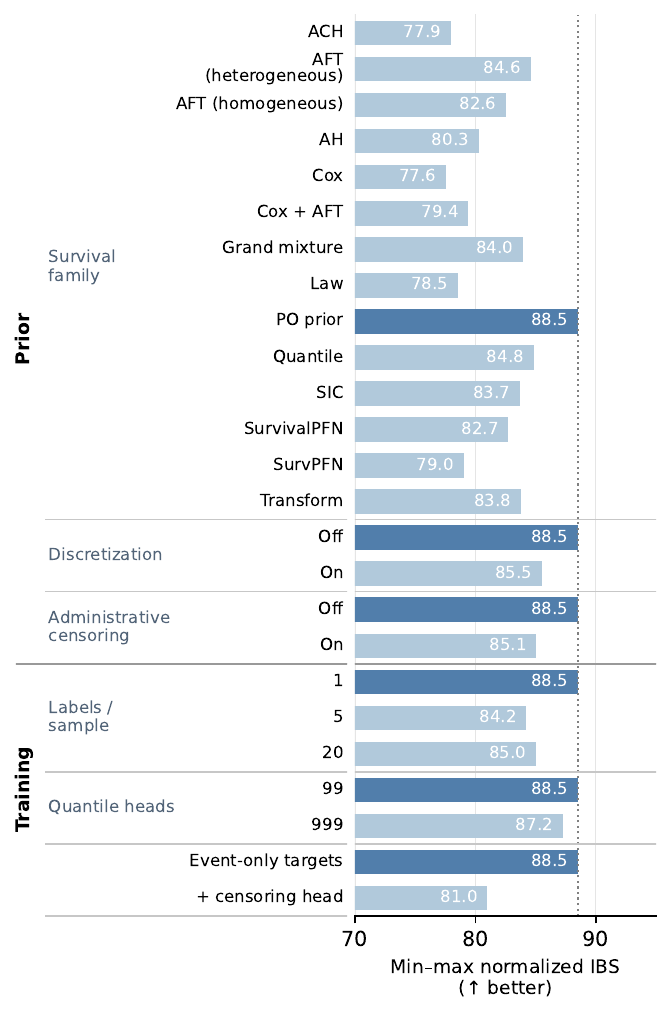}%
        \llap{\raisebox{1.50\linewidth}{\textbf{(c)} Ablations}\hspace*{.7\linewidth}}
    \end{minipage}
    \caption{\textbf{(a, b)} Critical difference diagrams of average ranks on all held-out datasets (lower is better), for IBS and C-index at the median horizon $t_{50}$; methods joined by a bar are not significantly different. \textbf{(c)} Ablations of prior and training choices, by min--max normalized IBS (higher is better).}
    \label{fig:main_results}
\end{figure}

\textbf{SurvICL is the best survival model overall.} On the 63 held-out datasets, SurvICL wins 82.9\% of pairwise comparisons in IBS, against 71.3\% for the runner-up, the Cox model (Figure~\ref{fig:win_rate}), and the difference is statistically significant (Figure~\ref{fig:main_results}a, ranks and winrates are related by an affine transform). It also achieves the highest win rate in C-index, 77.7\% against 71.6\% for random survival forests (Figures~\ref{fig:main_results}b and~\ref{fig:winplotcindex}). All existing survival TFMs trail behind classical methods. That the runners-up are the Cox model and random survival forests, with few parameters and strong inductive biases, underlines how hard survival analysis is for flexible models, which overfit small censored datasets.

\textbf{SurvICL is the first model to rank above Cox on small datasets.} 69.8\% of our held-out datasets have fewer than 1{,}000 samples, making this regime central and not a niche. SurvICL is the first to rank above Cox in IBS (win rate 84.9\% against 77.2\%, Figure~\ref{fig:winplotibssmall}) and it matches Cox in C-index, although these differences are not statistically significant (Figures~\ref{fig:cdplotibssmall} and~\ref{fig:cdplotcindexsmall}).

\textbf{On large datasets, boosting remains strongest.} On held-out datasets with more than 1{,}000 samples, SurvivalBoost is clearly the best method in IBS (Figure~\ref{fig:cdplotibslarge}, 86.9\% winrate). SurvICL ranks second in average IBS and first in C-index 50th quantile, ahead of TabSurv-WSA, RSF, and all other TFMs, although it wins fewer pairwise comparisons than on small datasets (Figure~\ref{fig:winplotibslarge}).

\textbf{Performance-runtime trade-off.}
SurvICL lies on the Pareto frontier across C-index and IBS for all dataset-size regimes (Appendix~\ref{sec:other_results}), and on the frontier 
for MAE in the small regime and all held-out datasets. SurvICL combines strong predictive performance with competitive runtime. 

\textbf{Calibration and time accuracy.} SurvICL is among the best-calibrated methods: its average KM-calibration rank ($6.0$) trails only the KM estimator ($4.3$), the Cox model ($5.0$), DeepSurv ($5.6$) and sksurv GBS-Cox ($5.9$) (Figure~\ref{fig:cdplotkmcal}). Both KM and the Breslow baseline hazard of the Cox model enforce calibration by construction. In MAE, SurvICL is close to Survival-IC and slightly behind SurvivalPFN, without statistically significant difference (Figure~\ref{fig:cdplotmae}).

\textbf{What makes SurvICL work.}
Figure~\ref{fig:main_results}c ablates design choices on development datasets with a shallower model trained for 10K steps, cf.\ Appendix~\ref{sec:ablations} for more details.
The prior has a substantial impact on performance: by keeping the exact same architecture and only changing the generation process, our \emph{proportional-odds prior} outperforms Cox ($77.6$) and mixed priors ($84.0$). Compared to the prior of the other TFMs, our prior yields better results than SurvPFN ($79.0$), SurvivalPFN ($82.7$), and SIC ($83.7$). This demonstrates that our prior is highly effective, while adding a censoring-time prediction head actually reduces performance ($81.0$).
Administrative censoring and time discretization have limited effects at this scale; we retain them to represent study-end
censoring and finite temporal resolution.
The final model uses $999$ quantiles for finer resolution, despite a small advantage for $99$ in the ablation
($88.5$ versus $87.2$). 

\begin{table}
\small
    \centering
    \begin{tabular}{ccccc}
    \toprule
        \textbf{Model} & \textbf{Loss}   & \textbf{Continuous Time} & \textbf{Prior} & \textbf{Prediction Target}   \\ \midrule
        \textcolor{SurvPFN}{\textbf{SurvPFN}} & NLL + ranking & \ding{51} &Weibull  & Observed times      \\
         \textcolor{SurvivalPFN}{\textbf{SurvivalPFN}} & NLL & \ding{55}  &  Random MLP & Latent event + censoring times  \\  
        \textcolor{Survival-IC}{\textbf{Survival-IC}}  & DeepHit Loss & \ding{55}  &EH Model & Observed times \\
         \textcolor{TFM4Survival}{\textbf{TFM4Survival}}  & BCE &\ding{55} & \ding{55}  &Binary survival labels \\
         \midrule
         \textcolor{SurvICL}{\textbf{SurvICL}} (ours) & Pinball Loss & \ding{51} & PO prior &  Latent event times\\
         \bottomrule
    \end{tabular}
    \caption{Major differences between existing survival TFMs and SurvICL. ``Latent'' targets are uncensored times generated by the prior, unobserved in real data.}
    \label{tab:comparison}%
\end{table}%

\section{Discussion and Conclusion}
\label{sec:comparison}

\textbf{Designing Efficient Survival TFMs.} SurvICL differs from existing survival TFMs along three axes (Table~\ref{tab:comparison}). \emph{Output and supervision.} Unlike previous survival TFMs that rely on discrete time bins such as TFM4Survival \citep{kim2026tabular} and SurvivalPFN \citep{qi2026survivalpfn}, which use classification or likelihood losses over time bins, SurvPFN \citep{bohm2026survpfn} uses a censored likelihood with a ranking term over 1{,}000 bins, and Survival-IC \citep{seletkov2026survival} uses the DeepHit loss \citep{lee2018deephit} over 10 bins. SurvICL is the first to output a continuous quantile function. Quantile outputs are not new per se: TabICLv2 \citep{qu2026tabiclv2} uses a pinball loss for regression, and censored quantile regression \citep{portnoy2003censored,pearce2022censored} modifies this loss to account for censoring. With latent-time supervision, no such modification is needed, as censoring is handled by the prior. SurvivalPFN also supervises with latent times, but over a time grid and with censoring times as an additional target, which our ablations show to be detrimental. \emph{Prior.} SurvivalPFN draws times from random noised neural networks with several censoring mechanisms, SurvPFN draws Weibull times from a risk score and a dataset-specific shape, and Survival-IC mixes an extended hazard and a discrete-time model, with covariate-independent and administrative censoring. Our PO prior instead combines a flexible, multi-modal baseline hazard with a random risk function, for both event and censoring times; in our ablations, it outperforms the SurvPFN prior as well as Cox and AFT priors. \emph{Backbone and evaluation.} Like Survival-IC, built on TabICL \citep{qu2025tabicl}, we reuse an existing architecture, TabICLv2, which is not our contribution. Finally, we provide, to our knowledge, the first comparison of all four survival TFMs, on 63 datasets.

\textbf{Implications.} On the small cohorts that dominate clinical research, SurvICL is the first model to rank above the Cox model, the default of practitioners for half a century, without any training or tuning, while remaining among the best-calibrated methods. This makes foundation models a credible default for survival analysis.

\textbf{Limitations and outlook.} On the largest datasets, gradient boosting remains slightly stronger, calling for longer contexts or light fine-tuning. Baseline hyperparameters are tuned on the first split and reused across splits; as tuning data then overlaps other splits' test sets, this protocol, if anything, favors the baselines. %
SurvICL addresses right-censored, single-event data; competing risks and time-varying covariates are natural extensions of the prior. We release our code and models to foster research on foundation models for time-to-event data.

\newpage
\subsection*{Acknowledgements}

LB and JAl thank Julie Josse and Charlotte Voinot for helpful discussions on survival analysis. We thank Félix Lefebvre and Farouk Yartaoui for inspiring our figure design.

Part of this research was conducted while LB was at EPFL (Schools of Computer and Communication Sciences and School of Life Sciences). His research was funded by an EPFL AI Center Postdoctoral Fellowship. DH and JAl acknowledge HPC resources from GENCI–IDRIS (Grant 2024-AD011014864R2 and AD011017641 respectively). DH thanks Probabl for providing Claude API access.
JAl, JAb, and GV acknowledge funding from the European Union’s Horizon Europe research and innovation program under grant agreement No 1010954433 (Intercept-T2D). GV acknowledges support from ANR via grant TaFoMo (ANR-25-CE23-1822). MLM acknowledges support from ANR via the grant StatQA (ANR-25-CE23-5646). This work is partly supported by Hi! PARIS and ANR/France 2030 program (ANR-23-IACL-0005).

\subsection*{AI use statement}

In this work, we used generative AI tools to generate synthetic data sets, design or provide feedback on research  methodology or experiments, implement methods, draft preliminary plots and provide feedback on research methodology or experiments.
We have not used generative AI tools to help develop theoretical models or conceptual frameworks, formulate mathematical claims, assist with translation, clean and reformat datasets, support qualitative and thematic data analysis, and interpret results. The categories provide critical ingredients for proving mathematical claims, assist in the writing of proofs, propose or refine hypotheses are not applicable to this work.

Additionally, we used generative AI tools to create or modify scientific figures or images, suggest experimental parameters, create or edit software code, and to brainstorm and source and search for information.

We have reviewed all AI-assisted work. The paper has been proofread by all authors. %
We take responsibility for the final content of this work,
including text, claims or artifacts produced with the aid of generative AI.

\subsection*{Reproducibility statement}

The codebase, dataset preprocessing files, and model checkpoints will be released in a few weeks. 

\newpage
\bibliographystyle{plainnat}
\bibliography{iclr2026_conference}
\appendix

\clearpage
\etocdepthtag.toc{mtappendix}

\begingroup
  \etocsettagdepth{mtchapter}{none}
  \etocsettagdepth{mtappendix}{subsection}
  \setlength{\parskip}{0pt}

  \etocsettocstyle{%
    \noindent{\LARGE Appendices}\par\vspace{1.4em}%
    \noindent{\large Contents}\par\vspace{0.8em}}{}

  \etocsetstyle{section}
    {}
    {\addvspace{1.3ex}}
    {\noindent
     \etoclink{\color{tocblue}\makebox[1.5em][l]{\etocthenumber}\etocthename}%
     \hfill\makebox[2em][r]{\etocthepage}\par}
    {}

  \etocsetstyle{subsection}
    {\addvspace{0.3ex}}
    {}
    {\noindent\hspace*{1.5em}%
     \etoclink{\color{tocblue}\makebox[2.4em][l]{\etocthenumber}\etocthename}%
     \nobreak\leaders\hbox to 0.9em{\hss.\hss}\hfill
     \makebox[2em][r]{\etocthepage}\par\addvspace{0.35ex}}
    {}

  \tableofcontents
\endgroup

\clearpage

\section{SurvICL model details} \label{sec:app:model}

\subsection{Architecture} \label{sec:app:architecture}

We build SurvICL on top of TabICLv2 \citep{qu2026tabiclv2}, which is at the time of writing the leading fully open-source model on the TabArena benchmark \citep{erickson2025tabarena}. While refer to the original paper for the details, the key components of the TabICLv2 architecture can be summarized as follows:
\begin{itemize}
    \item \textbf{Preprocessing}: First, some light preprocessing is applied, including ordinal encoding of categorical columns. After the preprocessing, the model receives a fully numerical table.
    \item \textbf{Target-aware embedding}: The original covariates $(x_{ij}) \in \bbR^{n \times m}$ are transformed into embeddings $(z_{ij}) \in \bbR^{n \times m \times 128}$ containing one embedding vector $z_{ij} \in \bbR^{128}$ per cell $(i,j)$. Each embedding vector is obtained from an affine transformation applied to three cells in the same row (``repeated feature grouping''). An affine transformation of the target $y_i$ is added for the context set rows.
    \item \textbf{Column-wise embedding}: Each column $j$ of the resulting embeddings is processed independently using three induced self-attention blocks that allow interactions between different rows in the same column, resulting in a complexity of $O(nm)$, which is linear in the number of rows, unlike a full Transformer. Crucially, the attention is designed to only retrieve information from the training set, meaning that predictions on test samples are independent of the other test samples.
    \item \textbf{Row-wise interaction}: Next, three full transformer blocks allow different columns within a row to interact. Before these blocks, four extra columns with fixed values are appended, and after the blocks, the values of these four columns are concatenated and serve as row embeddings, while the other columns are discarded.
    \item \textbf{ICL transformer}: After adding another embedding of the targets, the embeddings of each row interact via 12 regular transformer layers, again making sure that test samples cannot access other test samples.
    \item \textbf{Head}: Finally, a multilayer perceptron (MLP) head transforms the output embeddings into 999 quantile predictions (for regression).
\end{itemize}

Overall, the TabICLv2 architecture has a runtime complexity of $O(n^2 + nm^2)$, and SurvICL preserves this complexity since it only modifies the $O(nm)$ preprocessing and embedding parts. In addition, like other TFMs, we use ensembling, where the predictions of eight forward passes with different preprocessings and column permutations are averaged.

Building on TabICLv2, we introduce the following adaptations to handle survival data:
\begin{itemize}
    \item \textbf{Time normalization}: 
    We normalize observed times by their mean over the context set, yielding $\tilde{t}_i = t_i / \bar{t}$, where $\bar{t}_{\mathrm{context}}$ is the mean duration of the observed times. 
    This makes the time representation invariant to the choice of measurement units.
    During pre-training, both context and query sets are normalized and the mean $\bar{t}$ is taken across context and query set.
    During inference (after pre-training), the mean for $\bar{t}$ is taken across the context set, and the predicted quantiles are transformed back to the original time scale.

    \item \textbf{Censoring-aware target embedding}: We replace the scalar target embedding with a joint representation of the observed time and event indicator. Specifically, we sum the embedding of the continuous time $\tilde{t}_i$ and a learned embedding of the binary indicator $\delta_i$. 
    This representation is added to every cell embedding in context row $i$, allowing the column-wise encoder to distinguish observed events from censored observations.

    \item \textbf{Survival information at the ICL stage}: We also encode $(\tilde{t}_i,\delta_i)$ using a separate pair of time and event-indicator embeddings whose sum matches the dimension of the row representation. This embedding is added to context row representations before the ICL transformer. At both embedding stages, survival outcomes are supplied only for context rows.

    \item \textbf{Survival prediction}: The MLP head predicts quantiles of the conditional event-time distribution, using a softplus transformation to ensure that the quantiles are positive. After restoring the original time scale and forcing the predicted quantiles to be monotonous by sorting them, we obtain an estimated cumulative distribution function by linear interpolation between quantile levels. The survival function is then given by
    \[
        \widehat{S}(t \mid x)
        = 1 - \widehat{F}(t \mid x).
    \]
\end{itemize}

\begin{table}[h!]
\centering
\caption{Backbone architecture and prior for SurvICL.}
\label{tab:survicl-arch}
\small
\begin{tabular}{@{}ll@{}}
\toprule
\multicolumn{2}{@{}l}{\textit{Backbone}} \\
\midrule
Column-wise transformer & 3 blocks, 8 heads, 128 inducing points, QSSMax attention \\
Row-wise transformer & 3 blocks, 8 heads, 4 CLS tokens \\
In-context (ICL) transformer & 12 blocks, 8 heads, QSSMax attention \\
Max.\ sequence length (rows) & 1024 \\
Max.\ features per dataset & 100 (min.\ 1) \\
Regression head & Pinball / quantile, 999 quantile levels \\
Target embedding & Sum \\
Duration rescaling & Mean-duration \\
Label draws per row & 5 \\
Training steps & 200{,}000 \\
Learning rate & 0.0008 \\
\midrule
\multicolumn{2}{@{}l}{\textit{Synthetic prior}} \\
\midrule
Prior generator & GraphSCM (\texttt{graph\_scm\_survival}) \\
Structural family & Proportional odds (\texttt{generation\_type po}) \\
Survival parameter range & Wide \\
Loss & Pinball loss on death-time quantiles \\
\bottomrule
\end{tabular}
\end{table}

\subsection{Pretraining}

We follow many of the pretraining settings from TabICLv2 but make the following modifications:
\begin{itemize}
    \item We omit the second and third stages of pretraining, which scale to larger datasets, since the second stage did not show improvements on our benchmark.
    \item For computational reasons, we only train for 200K steps instead of 500K. Moreover, our pretraining consisted of two pretrainings for 100K steps each, meaning that the learning-rate schedule consisted of two 100K-steps schedules.
    \item We generate pretraining datasets with 1024 samples each, of which a random fraction in $[0.03, 0.9]$ is used as context (train) and the rest as query (test) samples. We lowered the lower bound from $0.3$ to $0.03$ to cover the smaller datasets found in our benchmark. 
    \item We pretrain fully in 32-bit floats to avoid potential instabilities.
\end{itemize}

\subsection{Example of generated datasets}
\label{sec:examplesgenerateddatasets}

\begin{figure}[h!]
    \centering
    \begin{minipage}{0.48\textwidth}
        \centering
        \includegraphics[width=\linewidth]{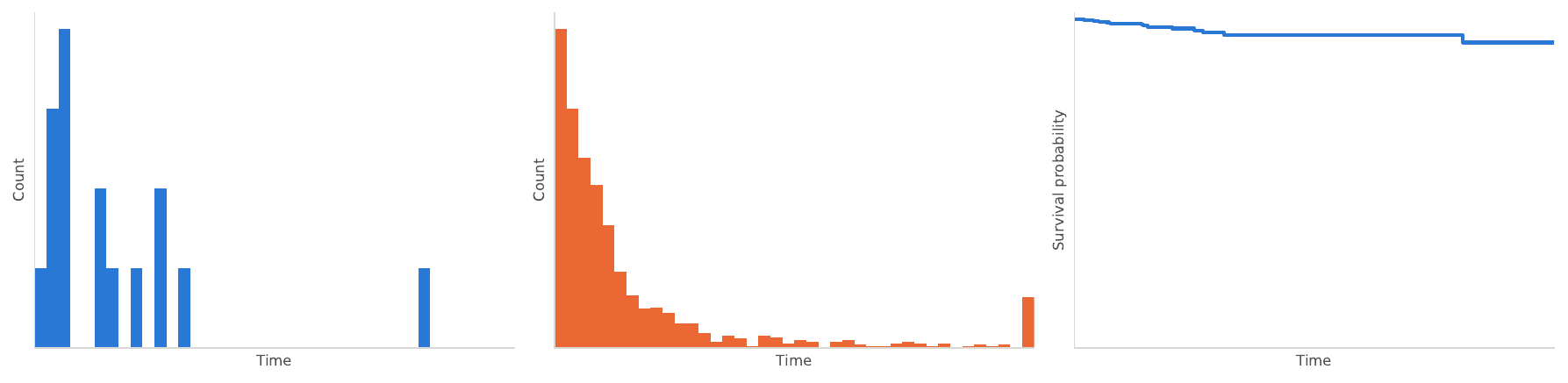}
        \vspace{0.2cm} \\
        \small (a) Sample 1
    \end{minipage}\hfill
    \begin{minipage}{0.48\textwidth}
        \centering
        \includegraphics[width=\linewidth]{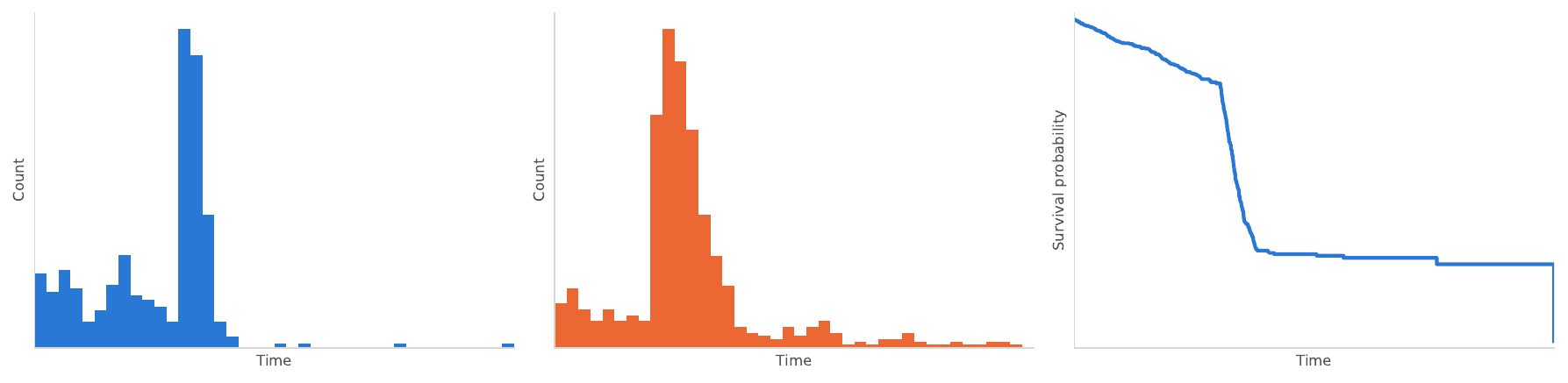}
        \vspace{0.2cm} \\
        \small (b) Sample 2
    \end{minipage}

    \vspace{1.5em} %

    \begin{minipage}{0.48\textwidth}
        \centering
        \includegraphics[width=\linewidth]{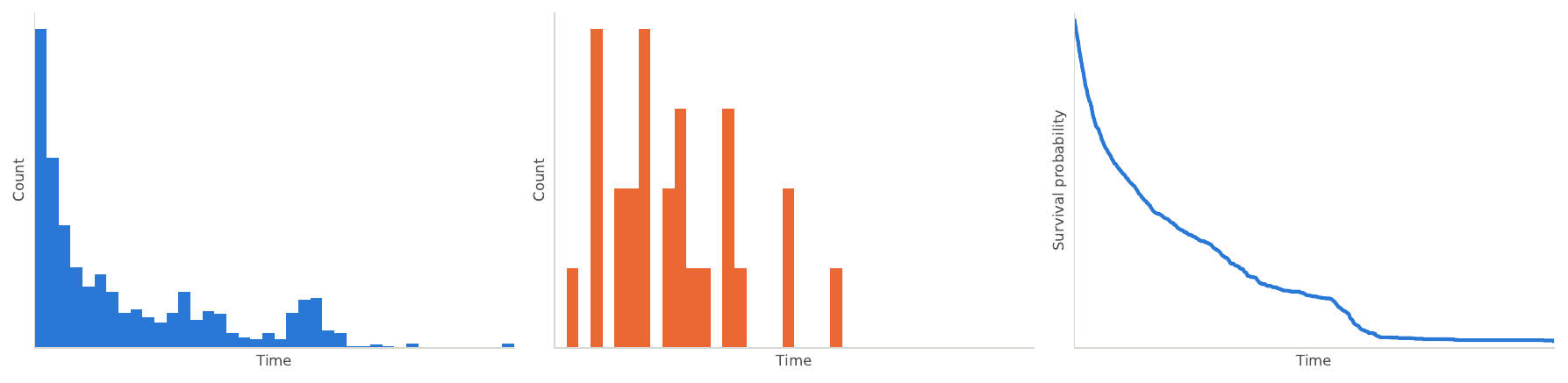}
        \vspace{0.2cm} \\
        \small (c) Sample 3
    \end{minipage}\hfill
    \begin{minipage}{0.48\textwidth}
        \centering
        \includegraphics[width=\linewidth]{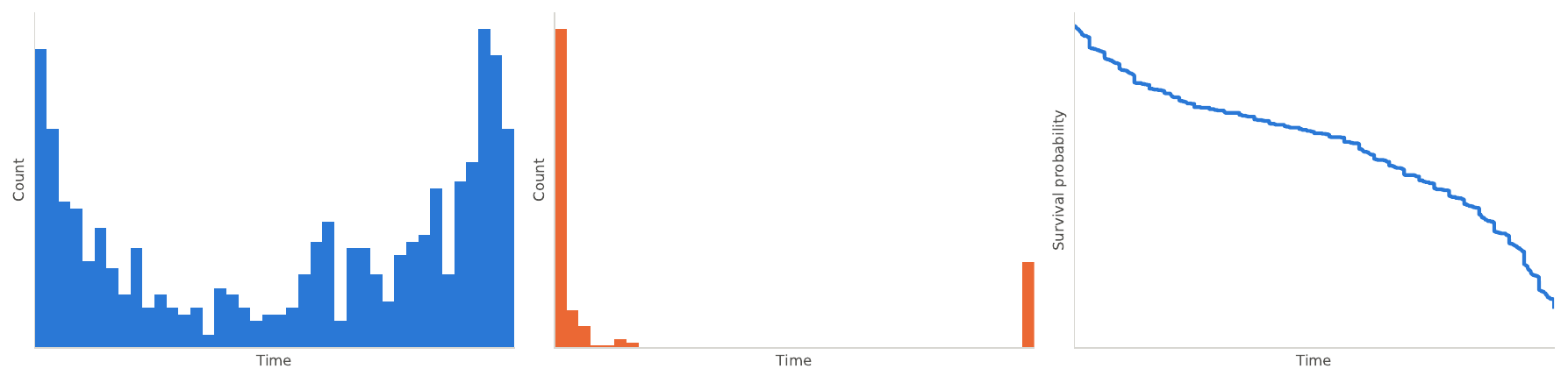}
        \vspace{0.2cm} \\
        \small (d) Sample 4
    \end{minipage}

    \vspace{1.5em}

    \begin{minipage}{0.48\textwidth}
        \centering
        \includegraphics[width=\linewidth]{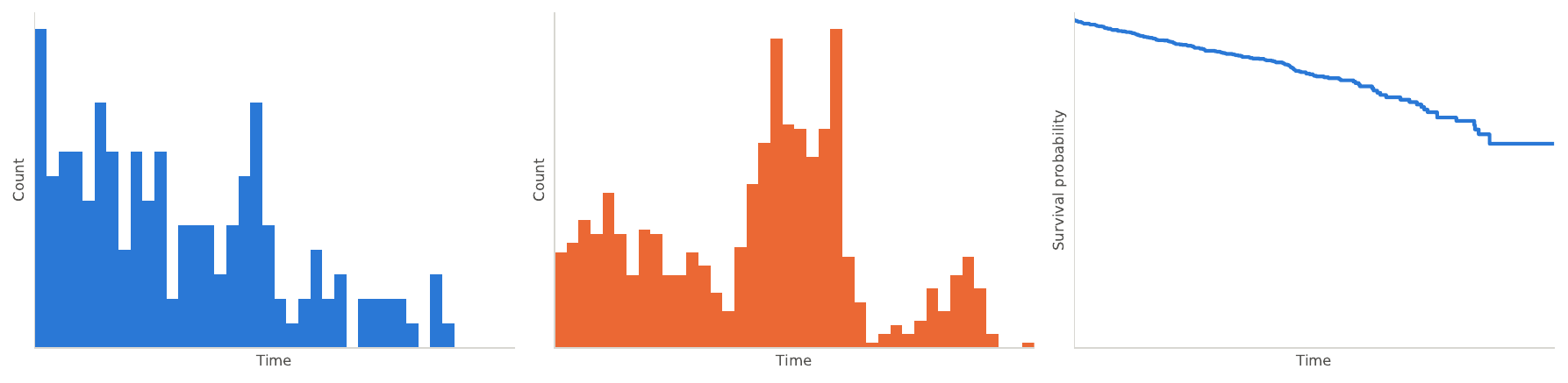}
        \vspace{0.2cm} \\
        \small (e) Sample 5
    \end{minipage}\hfill
    \begin{minipage}{0.48\textwidth}
        \centering
        \includegraphics[width=\linewidth]{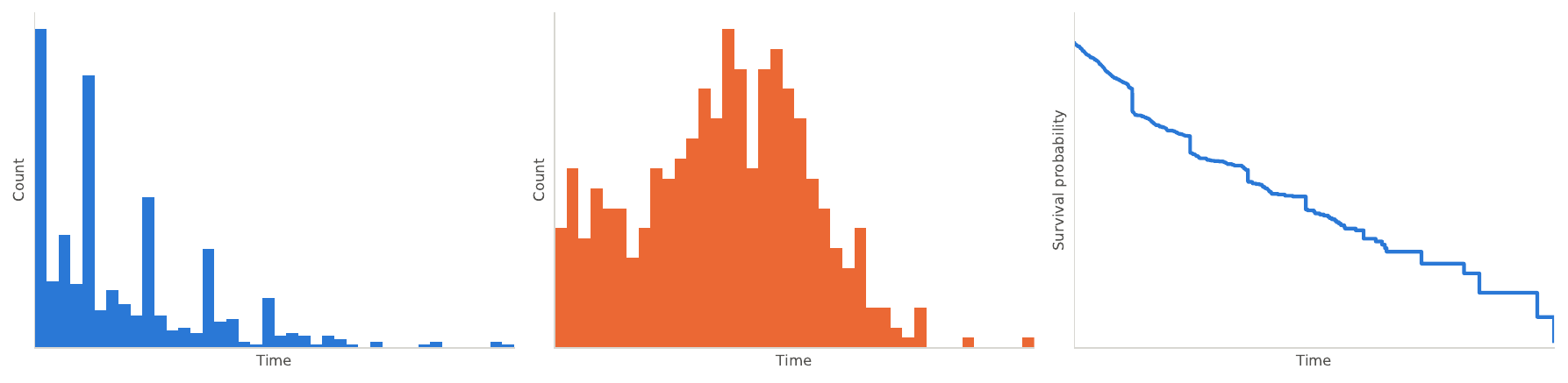}
        \vspace{0.2cm} \\
        \small (f) Sample 6
    \end{minipage}

    \caption{Six generated datasets using our final PO prior, incorporating administrative censoring and time discretization. For each sample, we show the histogram of observed event (left), the censoring events (center), and the Kaplan-Meier curve (right).}
    \label{fig:po_prior_all_samples}
\end{figure}

Figure \ref{fig:po_prior_all_samples} presents a diverse set of examples sampled from our final Proportional Odds (PO) prior. These datasets incorporate both administrative censoring and time discretization.  In each subfigure: left and middle panels are histograms showing the observed events and censoring events over time while the right panel is the resulting Kaplan-Meier survival probability curve.

\newpage
\section{Experimental Details}
\subsection{Details on Baselines} \label{sec:app:details_baselines}

\begin{wraptable}[17]{r}{0.44\linewidth}
\scriptsize
    \centering
    \begin{tabular}{llcccccc}
    \toprule
        \textbf{Type} & \textbf{Model Name} & \textbf{Reference} \\
        \midrule
        TFM & \textcolor{SurvICL}{\textbf{SurvICL}} & This work\\
          & \textcolor{SurvPFN}{\textbf{SurvPFN}} & \citealt{bohm2026survpfn} \\
          & \textcolor{SurvivalPFN}{\textbf{SurvivalPFN}} & \citealt{qi2026survivalpfn} \\
         & \textcolor{Survival-IC}{\textbf{Survival-IC}} & \citealt{seletkov2026survival} \\ 
         & \textcolor{TFM4Survival}{\textbf{TFM4Survival}\hspace*{-2ex}} & \citealt{kim2026tabular} \\
         \midrule Tree & \textcolor{SurvivalBoost}{\textbf{SurvivalBoost}\hspace*{-2ex}} &  \citealt{alberge2024survival}\\
          Based & GBC & \citealt{polsterl2020scikit}\\
         & \textcolor{RSF}{\textbf{RSF}}& \citealt{ishwaran2008random} \\
         \midrule Deep 
          & \textcolor{TabSurv}{\textbf{TabSurv}} & \citealt{kirpichenko2026tabsurv}\\
          Neural & \textcolor{SumoNet}{\textbf{SumoNet}}  &  \citealt{rindt2022survival}  \\
         Net & \textcolor{DSM}{\textbf{DSM}}  &   \citealt{nagpal2021deep}\\
         & \textcolor{DeepSurv}{\textbf{DeepSurv}}  & \citealt{katzman2018deepsurv}\\
         & \textcolor{DeepHit}{\textbf{DeepHit}} & \citealt{lee2018deephit}\\
         \midrule Shallow &  Cox &  \citealt{cox1972regression}\\
          Model & Kaplan-Meier &  \hspace*{-1ex}\citealt{kaplan1958nonparametric}\\
         \bottomrule
    \end{tabular}
    \caption{Overview of methods we consider in our experimental section.}
    \label{tab:all_baselines}
\end{wraptable}

\subsubsection{Usual methods}
Those two methods are the most used in the medical setting.

\textbf{Kaplan-Meier Estimator \citep{kaplan1958nonparametric}:} A non-parametric statistic used to estimate the marginal survival probability from observed lifetime data. Because it does not take individual covariates into account, it serves as the most basic population-level baseline for survival functions and is consistent. 

\textbf{Cox Model (Cox Proportional Hazards) \citep{cox1972regression}:} The standard semiparametric workhorse of survival analysis. It models the hazard rate as a product of a time-varying baseline hazard and an exponential linear combination of the covariates. It relies on the strong assumption of proportional hazards (the hazard ratio between any two individuals is constant over time).

\subsubsection{Tree-Based Models}
These models adapt decision trees and boosting trees to handle right-censored data, remaining highly competitive on tabular datasets.

\textbf{RSF (Random Survival Forests) \citep{ishwaran2008random}:} An extension of the Random Forest algorithm for time-to-event data. It constructs an ensemble of survival trees using specialized splitting criteria (like the log-rank test) to handle censored observations.

\textbf{GBS (Gradient Boosting for Survival)} \citep{polsterl2020scikit}: A boosting approach implemented in \texttt{scikit-survival} that sequentially builds regression trees. In our evaluation, we include two variants corresponding to different optimization targets: \textbf{sksurv GBS (Cox)}, which minimizes the negative log partial likelihood of the Cox proportional hazards model, and \textbf{sksurv GBS (C-index)}, which directly optimizes a smoothed concordance index.

\textbf{SurvivalBoost \citep{alberge2024survival}:} A modern, highly optimized gradient boosting framework specifically tailored for survival analysis, optimizing a proper scoring rule.

\subsubsection{Deep Neural Networks}
These methods leverage multi-layer architectures to capture complex, non-linear relationships between covariates and survival times, often dropping the proportional hazards assumption.

\textbf{DeepHit \citep{lee2018deephit}:} A discrete-time survival network that optimizes a loss that combines a likelihood and a ranking metric over an MLP, usable in survival analysis and competing events. 

\textbf{DeepSurv \citep{katzman2018deepsurv}:} It uses a Multi-Layer Perceptron (MLP) to replace the linear combination of covariates in the standard Cox model, allowing it to learn complex, non-linear interactions while still optimizing the Cox partial likelihood.

\textbf{DSM (Deep Survival Machines) \citep{nagpal2021deep}:} DSM uses a neural network to model the survival function as a mixture of parametric distributions (such as Weibull or Log-Normal), providing well-calibrated survival estimates.

\textbf{SumoNet \citep{rindt2022survival}:} SurvivalMonotonic-net utilizes a partially monotonic neural network to directly optimize the right-censored log-likelihood.

\textbf{TabSurv} \citep{kirpichenko2026tabsurv}: A specialized deep learning framework that adapts modern tabular neural networks to time-to-event prediction. Because this framework supports multiple representation backbones, we evaluate several architectural variants that appear throughout our results: the default \textbf{TabSurv}, \textbf{TabSurv-MLP} (using a standard Multi-Layer Perceptron), \textbf{TabSurv-RealMLP} (utilizing the more robust RealMLP tabular backbone), as well as \textbf{TabSurv-WAS} and \textbf{TabSurv-WSA} (which incorporate specialized attention and sparse embedding mechanisms designed specifically for tabular features).

\subsubsection{Tabular Foundation Models}
This category represents the most modern class of baselines: tabular foundation models trained on synthetic tabular data. They generalize to new datasets in a zero-shot or in-context manner.

\textbf{SurvPFN: \citep{bohm2026survpfn}} SurvPFN is a survival-specific Prior-Data Fitted Network (PFN) based on the architecture of TabPFN v2, pretrained on synthetic censored survival tasks (sampling from different distributions such as Weibull).

\textbf{SurvivalPFN \citep{qi2026survivalpfn}:} SurvivalPFN performs amortized Bayesian survival inference by using the observed covariates, event times, and censoring indicators as in-context examples to directly predict posterior event-time distributions for new individuals based on the  CausalPFN architecture.

\textbf{Survival-IC \citep{seletkov2026survival}:} Survival In-Context (SIC) is a prior-fitted foundation model pretrained on synthetic survival datasets to approximate Bayesian posterior predictive inference and based on the architecture of TabICL. 

\textbf{TFM4Survival \citep{kim2026tabular}:} TFM4Survival adapts pretrained tabular foundation models such as TabICL to censored time-to-event data through a sampling-based formulation, allowing existing TFMs to perform survival prediction without survival-specific pretraining.

\subsection{Hyperparameter tuning}
Table \ref{tab:hpbaselines} shows a description of the hyperparameter search spaces for each method. These spaces are used with a budget of 50 iterations.

{\small
\begin{longtable}{p{0.26\linewidth} l p{0.43\linewidth}}
    \caption{Hyper-parameter search space for every baseline in our survival benchmark. Each cell is searched with \texttt{RandomizedSearchCV} for 50 iterations on a single stratified 80/20 split of the seed-0 training fold, scored by the integrated Brier score (IBS). \texttt{randint}/\texttt{loguniform} denote the \texttt{scipy.stats} distributions; a list denotes a categorical choice; a lone value is held fixed (not searched).} \label{tab:hpbaselines}\\
        \toprule
        model & parameter & range \\
        \midrule
        \endfirsthead
        \toprule
        model & parameter & range \\
        \midrule
        \endhead
        \bottomrule
        \endfoot
        \bottomrule
        \endlastfoot
        \multicolumn{3}{c}{\textit{Others}} \\
        \midrule
        Kaplan--Meier & \multicolumn{2}{c}{no hyperparameters (population-level estimator)} \\
        \midrule
        Cox (lifelines, ridge-penalized) & penalizer & \texttt{loguniform}(1e-04, 10) \\
        \midrule
        \multicolumn{3}{c}{\textit{Tree-based methods}} \\
        \midrule
        Random Survival Forest & max features & sqrt, log2, 0.3, 0.5, 1.0 \\
         & min samples leaf & \texttt{randint}(1, 30) \\
         & min samples split & \texttt{randint}(2, 20) \\
         & n estimators & \texttt{randint}(100, 500) \\
        \midrule
        sksurv GBS (Cox) & dropout rate & 0.0, 0.1, 0.2 \\
         & learning rate & \texttt{loguniform}(1e-03, 0.3) \\
         & loss & coxph \\
         & n estimators & \texttt{randint}(50, 500) \\
         & subsample & 0.7, 0.85, 1.0 \\
        \midrule
        sksurv GBS (C-index) & dropout rate & 0.0, 0.1, 0.2 \\
         & learning rate & \texttt{loguniform}(1e-03, 0.3) \\
         & loss & coxph \\
         & max depth & \texttt{randint}(2, 8) \\
         & min samples leaf & \texttt{randint}(1, 20) \\
         & n estimators & \texttt{randint}(50, 400) \\
         & subsample & 0.7, 0.85, 1.0 \\
        \midrule
        SurvivalBoost & learning rate & \texttt{loguniform}(0.01, 0.1) \\
         & max depth & \texttt{randint}(2, 10) \\
         & n horizons per observation & \texttt{randint}(1, 5) \\
         & n iter & \texttt{randint}(5, 50) \\
         & n iter before feedback & \texttt{randint}(5, 50) \\
        \midrule
        \multicolumn{3}{c}{\textit{Neural network methods}} \\
        \midrule
        DeepHit & alpha & 0.2, 0.5, 0.9 \\
         & batch size & 128, 256, 512 \\
         & epochs & \texttt{randint}(256, 1024) \\
         & lr & \texttt{loguniform}(1e-05, 0.001) \\
         & nb output times & 40, 60, 80, 100 \\
         & num nodes shared & [128, 128], [256, 256], [64, 64] \\
         & sigma & 0.1, 0.5, 0.9\\
         & weight decay & \texttt{loguniform}(1e-05, 0.001) \\
        \midrule
        DeepSurv & activation & ReLU \\
         & batch norm & True, False \\
         & batch size & 128, 256, 512 \\
         & dropout & 0.0, 0.1, 0.2, 0.4 \\
         & early stopping & True (patience 20) \\
         & hidden dim & 16, 32, 64, 128 \\
         & lr & \texttt{loguniform}(1e-05, 0.001) \\
         & n epochs & 64, 128, 256, 512 \\
         & n layers & 1, 2, 3 \\
         & output bias & False \\
         & weight decay & \texttt{loguniform}(1e-05, 0.1) \\
        \midrule
        Deep Survival Machines & batch size & 128, 256, 512 \\
         & distribution & Weibull, LogNormal \\
         & iters & \texttt{randint}(50, 200) \\
         & k & \texttt{randint}(2, 7) \\
         & layers & [100], [100, 100], [50, 50], [100, 100, 100] \\
         & learning rate & \texttt{loguniform}(1e-04, 0.01) \\
         & temperature & 0.5, 1.0, 2.0 \\
        \midrule
        SuMo-Net & batch size & 128, 256, 512 \\
         & dropout & 0.0, 0.1, 0.25 \\
         & layers & [64, 64], [100, 100], [32, 32, 32] \\
         & layers surv & [64], [100], [100, 100] \\
         & lr & \texttt{loguniform}(1e-04, 0.01) \\
         & n iter & \texttt{randint}(200, 1500) \\
         & weight decay & \texttt{loguniform}(1e-05, 0.01) \\
        \midrule
        TabSurv-LAS & activation & ReLU, SiLU, SELU \\
         & batch size & 32, 64, 96 \\
         & d block & 128, 256, 512 \\
         & dropout & \texttt{loguniform}(0.01, 0.1) \\
         & emb dim rtdl & 8, 12, 16 \\
         & k & 8, 16, 32 \\
         & llh radius & 1, 2, 3, 4, 5 \\
         & lr & \texttt{loguniform}(1e-04, 0.005) \\
         & n blocks & \texttt{randint}(1, 4) \\
         & rtdl activation & True, False \\
         & rtdl n bins & 32, 48, 64 \\
        \midrule
        TabSurv-MLP & activation & ReLU, SiLU, SELU \\
         & batch size & 32, 64, 96 \\
         & emb dim rtdl & 8, 12, 16 \\
         & hidden dim & 128, 256, 512 \\
         & layer n & \texttt{randint}(1, 5) \\
         & llh radius & 1, 2, 3, 4, 5 \\
         & lr & \texttt{loguniform}(1e-04, 0.005) \\
         & normalization & False, True \\
         & rtdl activation & True, False \\
         & rtdl n bins & 32, 48, 64 \\
        \midrule
        TabSurv-RealMLP & activation & Mish, SELU \\
         & batch size & 32, 64, 96 \\
         & emb dim rtdl & 8, 12, 16 \\
         & frequency scale & \texttt{loguniform}(0.01, 1.0) \\
         & hidden dim & 128, 256, 512 \\
         & llh radius & 1, 2, 3, 4, 5 \\
         & lr & \texttt{loguniform}(1e-04, 0.005) \\
         & n frequencies & 32, 48, 64 \\
         & n hidden layers & \texttt{randint}(1, 5) \\
        \midrule
        TabSurv-WAS & activation & ReLU, SELU \\
         & batch size & 32, 64, 96 \\
         & d block & 64, 128, 256 \\
         & emb dim rtdl & 8, 12, 16 \\
         & k & 8, 16, 32 \\
         & llh radius & 1, 3, 5 \\
         & lr & \texttt{loguniform}(1e-04, 0.005) \\
         & n blocks & \texttt{randint}(1, 4) \\
         & rtdl activation & True, False \\
         & rtdl n bins & 32, 48, 64 \\
        \midrule
        TabSurv-WSA & activation & ReLU, SELU \\
         & batch size & 32, 64, 96 \\
         & d block & 128, 256, 512 \\
         & emb dim rtdl & 8, 12, 16 \\
         & k & 8, 16, 32 \\
         & llh radius & 1, 3, 5 \\
         & lr & \texttt{loguniform}(1e-04, 0.005) \\
         & n blocks & \texttt{randint}(1, 4) \\
         & rtdl activation & True, False \\
         & rtdl n bins & 32, 48, 64 \\
        \midrule
        \multicolumn{3}{c}{\textit{SurvICL and tabular foundation models}} \\
        \midrule
        SIC & \multicolumn{2}{c}{no hyperparameters (frozen DeepHit10 checkpoint)} \\
        \midrule
        SurvivalPFN & calibrate & False, True \\
         & max context size & 1024, 2048, 4096, 8192 \\
        \midrule
        SurvPFN & max context size & 1024, 2048, 4096 \\
         & max features & 5, 10 \\
        \midrule
        TFM-Survival & max train rows & 4096, 8192 \\
         & n intervals & \texttt{randint}(5, 20) \\
        \midrule
        Surv. Classif.: TabICL & n intervals & \texttt{randint}(5, 40) \\
\end{longtable}
}

\subsection{Issues for some methods}
Some baselines failed to complete their hyperparameter search or
final fit on a small subset of datasets, for reasons intrinsic to the method
rather than to the benchmark protocol (e.g., a training run that diverged
numerically, or a step of the method's own pipeline that is only defined
for datasets larger than the fold at hand). We assign the method a trivially achievable score on each dataset on which it did not complete: an IBS of $1$ (the maximum attainable value in expectation if $G$ is estimated perfectly) or a concordance index of $0.5$ (no better than chance), matching the metric being ranked. All critical-difference diagrams and pairwise win-rate figures are therefore always computed over the full 63-dataset held-out cohort. Methods with at least one imputed dataset are marked with an asterisk in the critical-difference diagrams and with a hatched bar in the win-rate figures; Table~\ref{tab:baseline-failures} lists the affected (method, dataset) pairs and their cause.

\begin{table}[h!]
\centering
\caption{Baseline (method, dataset) pairs that created issues}
\label{tab:baseline-failures}
\small
\begin{tabular}{@{}llp{0.4\linewidth}@{}}
\toprule
\textbf{Method} & \textbf{Dataset(s)} & \textbf{Cause} \\
\midrule
DeepSurv &
  \textit{divorce}, \textit{actg}, \textit{FRTCS} &
  Training loss diverged (became non-finite) during optimization on these
  three small datasets. \\
\addlinespace
TFM4Surival: TabICL &
  \textit{sac3}, \textit{DLBCL}, \textit{gse1992} &
  Out-of-memory issue on these larger/high-dimensional datasets. \\
\addlinespace
SurvivalPFN &
  \textit{AML\_Bull}, \textit{gse1992}, \textit{vdv} &
  The method's internal dimensionality-reduction step (a truncated SVD)
  requires more components than the smaller of the fold's row or feature
  count provides, and errors when that bound is violated. \\

\addlinespace
TabSurv &
  \textit{DBCD}, \textit{cesc tcga, kirc tcga, lgg tcga} &
  The method's obtained a MAE of value infty. \\
  
\bottomrule
\end{tabular}
\end{table}

\subsection{Details on Datasets}

An overview of the development and held-out datasets is shown in Tables \ref{tab:development_datasets} and \ref{tab:heldout_datasets}.

\begin{table}[h!]
\small
    \centering
    \begin{tabular}{lllrrrrr}
        \toprule 
        name & source & n samples & n features & censoring frac & km survival at tmax \\
        \midrule
         Framingham & survset  & 4699 & 18 & 68.7\% & 60.5 \% \\
         MCLcleaned & survset & 92 & 574 & 30.4 \% & 0.0 \% \\
         Melanoma & survset & 205 & 5 & 72.2 \% & 64.5 \% \\
         NSBCD & survset & 115 & 549 & 67.0 \% & 57.4 \% \\
         Rossi & survset & 432 & 29 & 73.6 \% & 73.6 \% \\
         Unemployment & survset &  452 & 8 & 43.4 \% & 5.8 \% \\
         acath & survset &  3504 & 3 & 33.4 \% & 0.0 \% \\
         aids & survset &  467 & 5 & 59.7 \% & 49.2 \% \\
         brca tcga pan can atlas 2018 & cbioportal & 980 & 137 & 95.2 \% & 60.1 \% \\
         cgd & survset  & 128 & 25 & 65.6 \% & 47.6 \% \\
         challenge ens & custom  & 3173 & 42 & 49.6 \% & 5.9 \% \\
         chop & survset  & 414 & 3833 & 60.1 \% & 39.5 \% \\
         cost & survset  & 518 & 13 & 22.0 \% & 22.0 \% \\
         crc orion 2024 & cbioportal  & 74 & 131 & 66.2 \% & 30.4 \% \\
         csl & survset & 446 & 6 & 39.5 \% & 17.3 \% \\
         d.oropha.rec & survset  & 192 & 30 & 27.6 \% & 16.5 \% \\
         dataDIVAT3 & survset  & 4267 & 17 & 94.4 \% & 84.7 \% \\
         dataOvarian1 & survset  & 912 & 163 & 40.4 \% & 11.9 \% \\
         epileptic & survset  & 605 & 5 & 63.5 \% & 46.3 \% \\
         flchain & pycox  & 6524 & 8 & 69.9 \% & 65.6 \% \\
         gse3143 & survset  & 158 & 8664 & 68.4 \% & 41.9 \% \\
         kirc tcga pan can atlas 2018 & cbioportal  & 369 & 53 & 92.7 \% & 86.5 \% \\
         kirp tcga pan can atlas 2018 & cbioportal  & 251 & 56 & 96.0 \% & 92.9 \% \\
         luad tcga pan can atlas 2018 & cbioportal  & 392 & 98 & 82.4 \% & 68.4 \% \\
         lusc tcga pan can atlas 2018 & cbioportal  & 481 & 100 & 57.2 \% & 0.0 \% \\
         micro.censure & survset  & 117 & 118 & 77.8 \% & 42.4 \% \\
         msk 2026 & cbioportal & 40755 & 56 & 57.0 \% & 31.7 \% \\
         oldmort & survset & 4603 & 17 & 57.2 \% & 0.2 \% \\
         ov tcga pan can atlas 2018 & cbioportal  & 427 & 34 & 48.2 \% & 6.5 \% \\
         ova & survset  & 358 & 13 & 25.7 \% & 22.2 \% \\
         ovarian & survset  & 26 & 4 & 53.8 \% & 49.7 \% \\
         pancan pcawg 2020 & cbioportal  & 361 & 255 & 62.9 \% & 0.0 \% \\
         rdata & survset  & 1040 & 7 & 47.4 \% & 38.1 \% \\
         sarc tcga pan can atlas 2018 & cbioportal  & 181 & 147 & 87.3 \% & 74.3 \% \\
         smarto & pycox  & 3873 & 39 & 88.1 \% & 72.5 \% \\
         ucs tcga pan can atlas 2018 & cbioportal  & 57 & 11 & 38.6 \% & 9.2 \% \\
         uvm tcga pan can atlas 2018 & cbioportal  & 80 & 6 & 71.2 \% & 38.5 \% \\
         vlbw & survset  & 617 & 55 & 82.7 \% & 0.0 \% \\
         whas500 & survset  & 461 & 18 & 61.8 \% & 0.0 \% \\
         zinc & survset & 431 & 23 & 81.2 \% & 79.0 \% \\
        \end{tabular}
    \caption{Overview of development datasets.}
    \label{tab:development_datasets}
\end{table}

\begin{table}[h!]
\small
    \centering
    \begin{tabular}{lllrrrrr}
        \toprule 
        name & source & n samples & n features & censoring frac & km survival at tmax \\
        \midrule
         AML Bull & survset  & 116 & 6283 & 42.2 \% & 28.3 \% \\
         Aids2 & survset & 2839 & 14 & 38.0 \% & 5.8 \% \\
         Bergamaschi & survset  & 82 & 10 & 65.9 \% & 58.3 \% \\
         DBCD & survset  & 295 & 4919 & 73.2 \% & 61.0 \% \\
         DLBCL & survset  & 240 & 7399 & 42.5 \% & 26.8 \% \\
         Dialysis & survset & 6805 & 37 & 76.4 \% & 58.7 \% \\ 
         FRTCS & survset  & 697 & 5 & 89.7 \% & 89.1 \% \\ 
         LeukSurv & survset & 1043 & 30 & 15.7 \% & 4.4 \% \\ 
         Pbc3 & survset & 349 & 21 & 82.5 \% & 63.4 \% \\ 
         TRACE & survset  & 1878 & 6 & 49.0 \% & 43.9 \% \\ 
         UnempDur & survset & 3241 & 6 & 38.7 \% & 10.5 \% \\ 
         acc tcga pan can atlas 2018 & cbioportal  & 91 & 39 & 63.7 \% & 49.9 \% \\ 
         actg & survset & 1151 & 20 & 91.7 \% & 90.1 \% \\ 
         blca tcga pan can atlas 2018 & cbioportal  & 301 & 84 & 76.1 \% & 62.0 \% \\ 
         breast & survset  & 100 & 4 & 74.0 \% & 71.4 \% \\ 
         burn & survset & 154 & 14 & 68.8 \% & 48.0 \% \\ 
         cancer & survset  & 228 & 30 & 27.6 \% & 5.0 \% \\ 
         cesc tcga pan can atlas 2018 & cbioportal  & 241 & 97 & 95.0 \% & 91.3 \% \\ 
         chol tcga pan can atlas 2018 & cbioportal  & 36 & 53 & 50.0 \% & 19.5 \% \\ 
         coadread tcga pan can atlas 2018 & cbioportal  & 543 & 133 & 86.2 \% & 62.1 \% \\ 
         colon & survset  & 929 & 14 & 51.3 \% & 45.5 \% \\ 
         crc hta8 htan 2024 & cbioportal  & 83 & 91 & 69.9 \% & 24.4 \% \\ 
         dataDIVAT2 & survset & 1837 & 4 & 68.3 \% & 31.1 \% \\ 
         diabetes & survset & 394 & 4 & 60.7 \% & 53.0 \% \\ 
         difg glass & cbioportal  & 661 & 179 & 24.8 \% & 0.0 \% \\ 
         divorce & survset  & 3371 & 5 & 69.4 \% & 55.7 \% \\ 
         e1684 & survset  & 284 & 3 & 31.0 \% & 28.3 \% \\ 
         esca tcga pan can atlas 2018 & cbioportal  & 182 & 90 & 58.2 \% & 10.7 \% \\ 
         gbm tcga pan can atlas 2018 & cbioportal  & 277 & 30 & 31.8 \% & 0.0 \% \\ 
         gbsg & pycox  & 2232 & 7 & 43.2 \% & 36.9 \% \\ 
         grace & survset  & 1000 & 5 & 67.6 \% & 63.5 \% \\ 
         gse1992 & survset  & 124 & 15545 & 71.8 \% & 45.4 \% \\  
         hdfail & pycox  & 52422 & 38 & 94.5 \% & 56.1 \% \\ 
         heart & survset  & 103 & 4 & 27.2 \% & 15.2 \% \\ 
         heartvalve & survset  & 256 & 27 & 78.9 \% & 53.0 \% \\ 
         hepatoCellular & survset  & 227 & 43 & 57.3 \% & 51.2 \% \\ 
         hnsc tcga pan can atlas 2018 & cbioportal  & 364 & 119 & 83.0 \% & 44.6 \% \\ 
         lgg tcga pan can atlas 2018 & cbioportal  & 422 & 63 & 91.9 \% & 43.3 \% \\ 
         lihc tcga pan can atlas 2018 & cbioportal  & 371 & 70 & 64.7 \% & 18.2 \% \\ 
         meso tcga pan can atlas 2018 & cbioportal  & 86 & 58 & 15.1 \% & 0.0 \% \\ 
         metabric & pycox  & 1904 & 9 & 42.1 \% & 0.0 \% \\ 
         mgus & survset  & 241 & 13 & 6.6 \% & 1.8 \% \\ 
         msk 2021 & cbioportal & 25659 & 86 & 60.4 \% & 34.3 \% \\
         nki70 & survset  & 144 & 77 & 66.7 \% & 48.0 \% \\ 
         nwtco & pycox  & 4028 & 6 & 85.8 \% & 84.9 \% \\ 
         paad tcga pan can atlas 2018 & cbioportal  & 184 & 75 & 46.2 \% & 13.7 \% \\ 
         pbc & survset & 312 & 6 & 59.9 \% & 34.1 \% \\ 
         pharmacoSmoking & survset  & 125 & 19 & 28.8 \% & 28.8 \% \\ 
         phpl04K8a & survset  & 442 & 21 & 46.6 \% & 0.0 \% \\ 
         prostate & survset  & 502 & 28 & 29.5 \% & 23.8 \% \\ 
         prostateSurvival &               survset  & 14294 & 8 & 94.4 \% & 81.4 \% \\ 
         rhc & survset  & 5735 & 79 & 35.1 \% & 0.0 \% \\ 
         sac3 & pycox  & 50000 & 45 & 37.3 \% & 20.2 \% \\ 
         scania & survset & 1931 & 9 & 43.8 \% & 15.9 \% \\ 
         skcm tcga pan can atlas 2018 & cbioportal  & 433 & 172 & 50.3 \% & 5.7 \% \\ 
         stad tcga pan can atlas 2018 & cbioportal  & 355 & 112 & 75.5 \% & 64.4 \% \\ 
         stagec & survset  & 146 & 22 & 63.0 \% & 55.8 \% \\ 
         support & pycox  & 8873 & 14 & 32.0 \% & 24.1 \% \\ 
         ucec tcga pan can atlas 2018 & cbioportal  & 486 & 58 & 90.7 \% & 83.6 \% \\ 
         uis & survset & 628 & 17 & 19.1 \% & 16.6 \% \\ 
         vdv & survset  & 78 & 4705 & 56.4 \% & 56.4 \% \\ 
         veteran & survset  & 137 & 9 & 6.6 \% & 0.0 \% \\ 
         wpbc & survset  & 198 & 32 & 76.3 \% & 65.3 \% 
        \end{tabular}
    \caption{Overview of held-out datasets.}
    \label{tab:heldout_datasets}
\end{table}

\FloatBarrier

\section{Ablations}
\label{sec:ablations}
To isolate the impact of specific design choices, we conduct ablation studies using a shared \texttt{nanotabiclexsurvival} backbone trained for 10{,}000 steps. Each ablation alters exactly one variable against a control arm trained on the Proportional Odds (\texttt{po}) prior as detailed in the following table. One arm can be trained in 2h on 2 A100. 

\begin{table}[h!]
\centering
\caption{Backbone architecture shared by every arm.}
\label{tab:survablation-arch}
\small
\begin{tabular}{@{}ll@{}}
\toprule
\multicolumn{2}{@{}l}{\textit{Backbone (\texttt{nanotabiclexsurvival})}} \\
\midrule
Column-wise transformer & 2 blocks, 4 heads, 128 inducing points, SSMax attention \\
Row-wise transformer & 2 blocks, 4 heads, 4 CLS tokens \\
In-context (ICL) transformer & 4 blocks, 4 heads, SSMax attention \\
Regression head & Pinball / quantile, 99 quantile levels \\
Target embedding & Sum \\
Duration rescaling (control) & Mean-duration \\
Label draws per row (control) & 1 \\
Learning rate & 0.0015 \\
\midrule
\multicolumn{2}{@{}l}{\textit{Synthetic prior}} \\
\midrule
Prior generator & GraphSCM (\texttt{graph\_scm\_survival}) \\
Loss & Pinball loss on death-time quantiles \\
\bottomrule
\end{tabular}
\end{table}

\subsubsection{Ablation design}
\label{sec:survablation-design}

\textbf{Survival-family axis.} Table~\ref{tab:survablation-families} lists the generative equation of
every family tested.

\begin{table}[h!]
\centering
\caption{Structural survival families compared on the family axis. $H$, $O$, $A$ and $h$
  denote the cumulative hazard, cumulative odds, Aranda-Ordaz/Box-Cox transform, and
  hazard of $T\mid x$; a $0$ subscript denotes the shared random baseline.}
\label{tab:survablation-families}
\small
\begin{tabular}{@{}llp{0.5\linewidth}@{}}
\toprule
\textbf{Family} & \textbf{Arm(s)} & \textbf{Generative structure} \\
\midrule
Proportional hazards & \texttt{cox} & $H(t\mid x) = e^{\eta} H_0(t)$ \\
Proportional odds & \texttt{po} & $O(t\mid x) = e^{\eta} O_0(t)$, $O = (1-S)/S$ \\
Transformation family & \texttt{transform} & $A_\rho(t\mid x) = e^{\eta} A_\rho(S_0(t))$, $\rho \sim [0.05, 3]$ per dataset (interpolates PH at $\rho\to0$ and PO at $\rho=1$) \\
Additive hazards & \texttt{ah} & $h(t\mid x) = h_0(t) + c\cdot\tanh(\eta)$, clamped at 0 \\
Accelerated hazards & \texttt{ach} & $h(t\mid x) = h_0(t\,e^{\eta})$ (curves leave $S{=}1$ together) \\
AFT, homoscedastic & \texttt{aft-homo} & $\log T = \mu(x) + (\log T_0 - \mathbb{E}\log T_0)$ \\
AFT, heteroscedastic & \texttt{aft-hetero} & as above with a per-subject log-time spread, producing crossing survival curves \\
Quantile model & \texttt{quantile} & $\log Q_p(T\mid x) = \log Q_{0,p} + \eta_p(x)$, 3 anchor levels \\
Classical parametric laws & \texttt{law} & Weibull / gamma / exponential / log-logistic mixture (``all''), proportional hazards \\
Simplified (reduced) & \texttt{simple} & single-law, non-mixture, single label draw \\
Cox + AFT mixture & \texttt{cox-aft} & per-dataset random choice of \texttt{cox} or \texttt{aft} \\
Grand mixture & \texttt{grand-mix} & per-dataset random choice among \texttt{cox}, \texttt{aft}, \texttt{po}, \texttt{ah}, \texttt{quantile} \\
SurvPFN prior & \texttt{survpfn} & closed-form Weibull PH/non-PH times, ported from SurvPFN's own pretraining prior \\
SurvivalPFN prior & \texttt{survivalpfn} & nonparametric hazard shape via a monotone Bernstein-polynomial map, ported from SurvivalPFN's prior \\
SIC prior & \texttt{sic} & extended-hazards law with a per-dataset hazard-family draw, ported from SIC's pretraining recipe \\
\bottomrule
\end{tabular}
\end{table}

\textbf{Recipe axes.} Seven arms each flip exactly one flag off the control, holding
\texttt{generation\_type=po} fixed (Table~\ref{tab:survablation-axes}). Two of the axes
(discretization, administrative censoring) are already off in the control, so we ablate by
turning them \emph{on} rather than removing them -- the same one-knob-at-a-time principle,
applied in the direction the control recipe permits.

\begin{table}[t]
\centering
\caption{Recipe axes ablated against the \texttt{po} control, one flag at a time.}
\label{tab:survablation-axes}
\small
\begin{tabular}{@{}lll@{}}
\toprule
\textbf{Axis} & \textbf{Control} & \textbf{Variant(s)} \\
\midrule
Discretization & Off & On (10\% per-dataset chance of coarsening times to a grid) \\
Administrative censoring & Off & On (50\% per-dataset chance of a study-end cutoff) \\
Label draws per row & 1 & 5, 20 \\
Quantile head width & 99 levels & 999 levels \\
Auxiliary censoring head & Event-only targets & $+$ censoring-distribution head (train-time only, dropped at inference) \\
\bottomrule
\end{tabular}
\end{table}

\subsubsection{Evaluation protocol}
Arms are evaluated at their final checkpoint on frozen, held-out validation datasets. To account for varying dataset scales and censoring rates, raw metrics are min–max normalized per dataset (100 = best arm, 0 = worst arm) before averaging. The primary ranking metric is the mean Integrated Brier Score (IBS).

\section{Other Results}
\label{sec:other_results}

\subsection{All held-out datasets}

\subsubsection{IBS}

\begin{figure}[h!]
    \centering
    \includegraphics[width=0.5\linewidth]{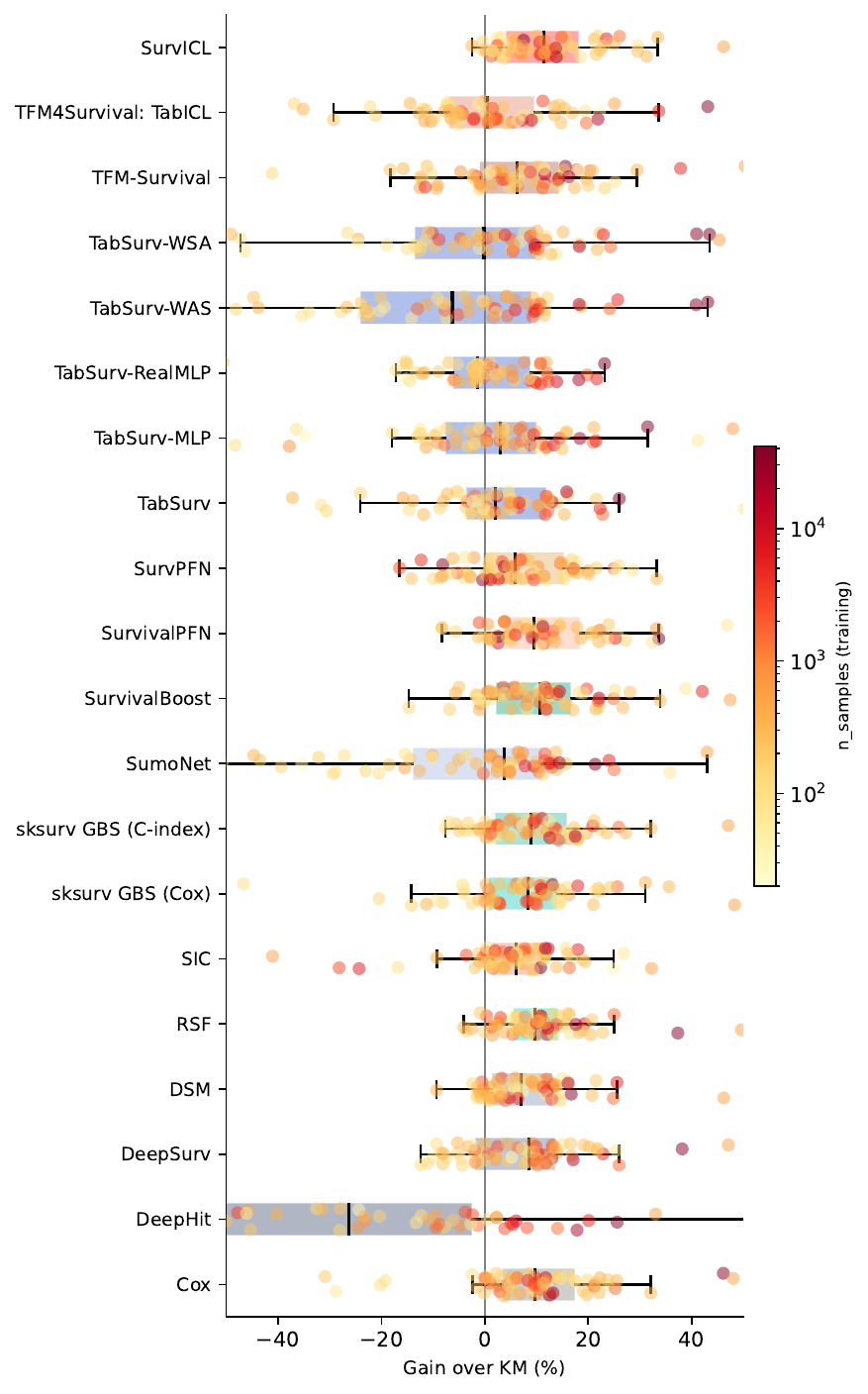}
    \caption{\textbf{IBS on all held-out datasets, compared to the IBS score obtained with a Kaplan-Meier estimator (higher is better).} The points show individual dataset performances, and a boxplot summarizes the overall distribution. The dot color represents the dataset size.}
    \label{fig:boxplotibs}
\end{figure}

\begin{figure}[h!]
    \centering
    \includegraphics[width=\linewidth]{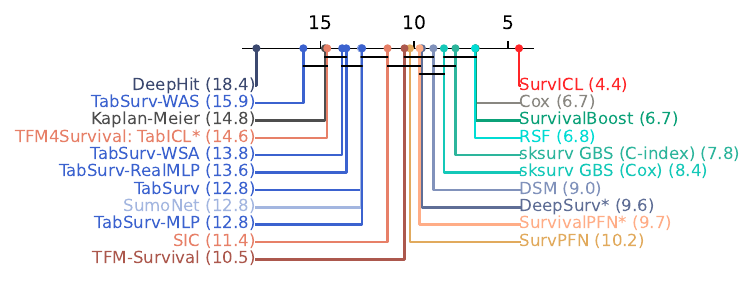}
    \caption{\textbf{Critical difference diagram of average ranks on all held-out datasets (lower is better) for IBS.} Methods joined by a bar are not significantly different}
    \label{fig:cdplotibs}
\end{figure}

\begin{figure}[h!]
    \centering
    \includegraphics[width=\linewidth]{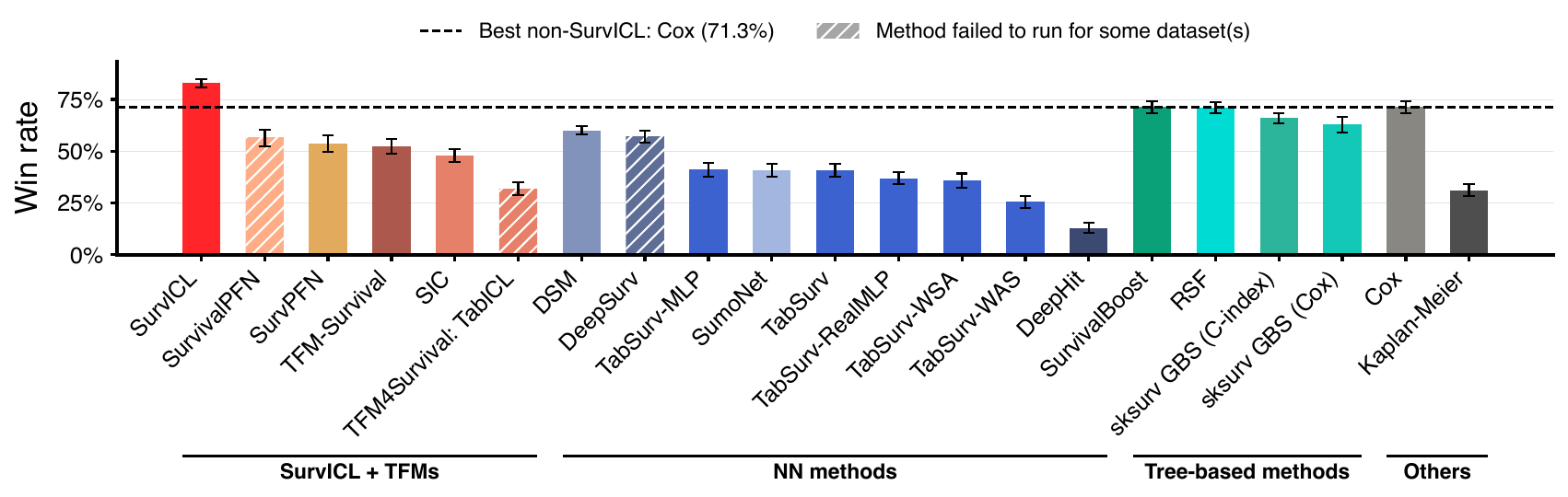}
    \caption{\textbf{Pairwise win rate IBS all held-out datasets:} the fraction of (dataset, competitor) pairs on which a method achieves a lower IBS. The scores are the same as in Figure~\ref{fig:win_rate}, but shown with a different grouping.}
    \label{fig:winplotibs}
\end{figure}

\begin{figure}[h!]
    \centering
    \includegraphics[width=0.9\linewidth]{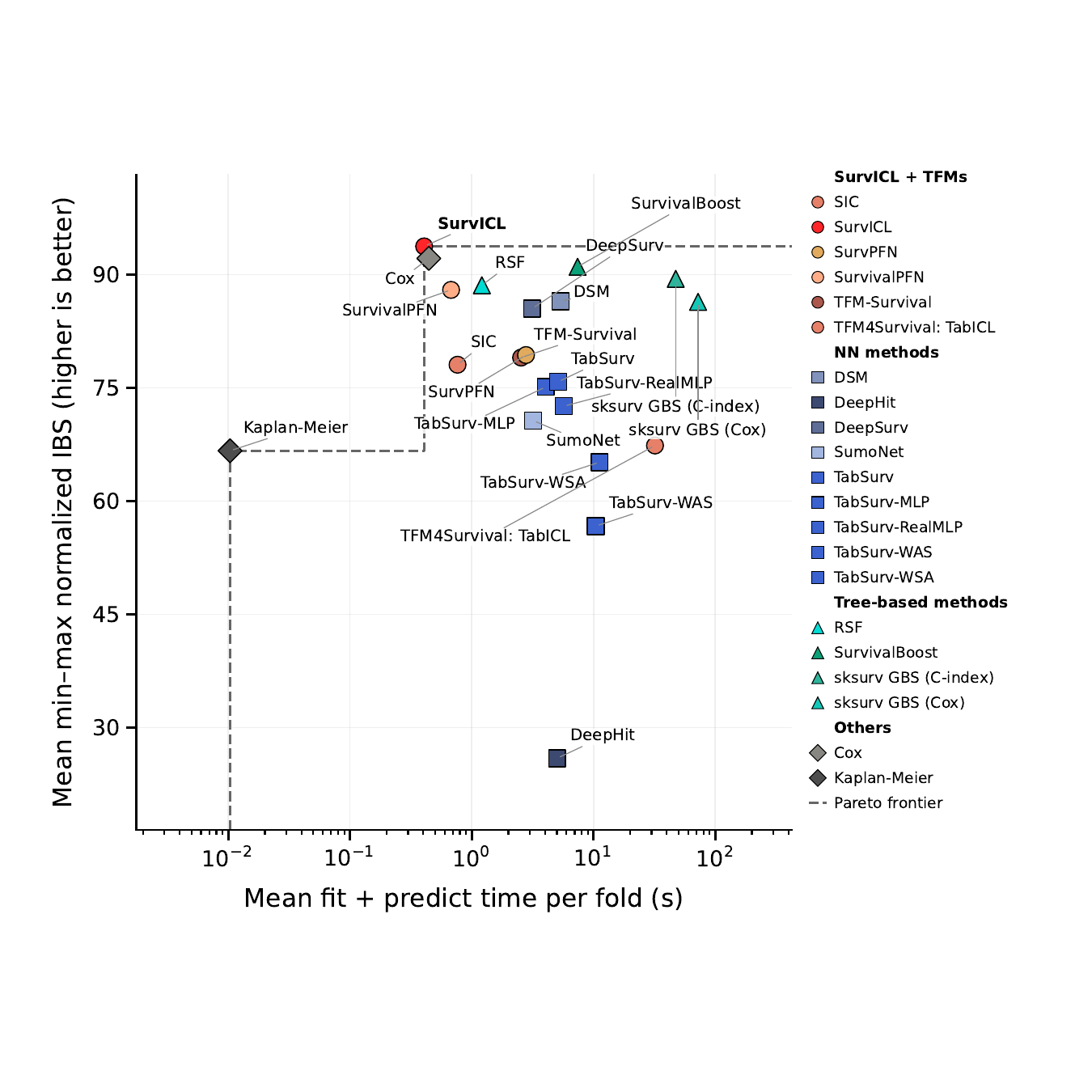}
    \caption{\textbf{Pareto front of mean normalized IBS (higher is better) and fit $+$ predict time.} We report average performances over all held-out datasets. IBS was normalized to provide comparability across all datasets.}
    \label{fig:paretoibs}
\end{figure}

Results for the IBS on held-out datasets are shown in Figures \ref{fig:boxplotibs}, \ref{fig:cdplotibs}, \ref{fig:winplotibs}, \ref{fig:paretoibs}.

\FloatBarrier

\subsubsection{C-index50}

\begin{figure}[h!]
    \centering
    \includegraphics[width=0.5\linewidth]{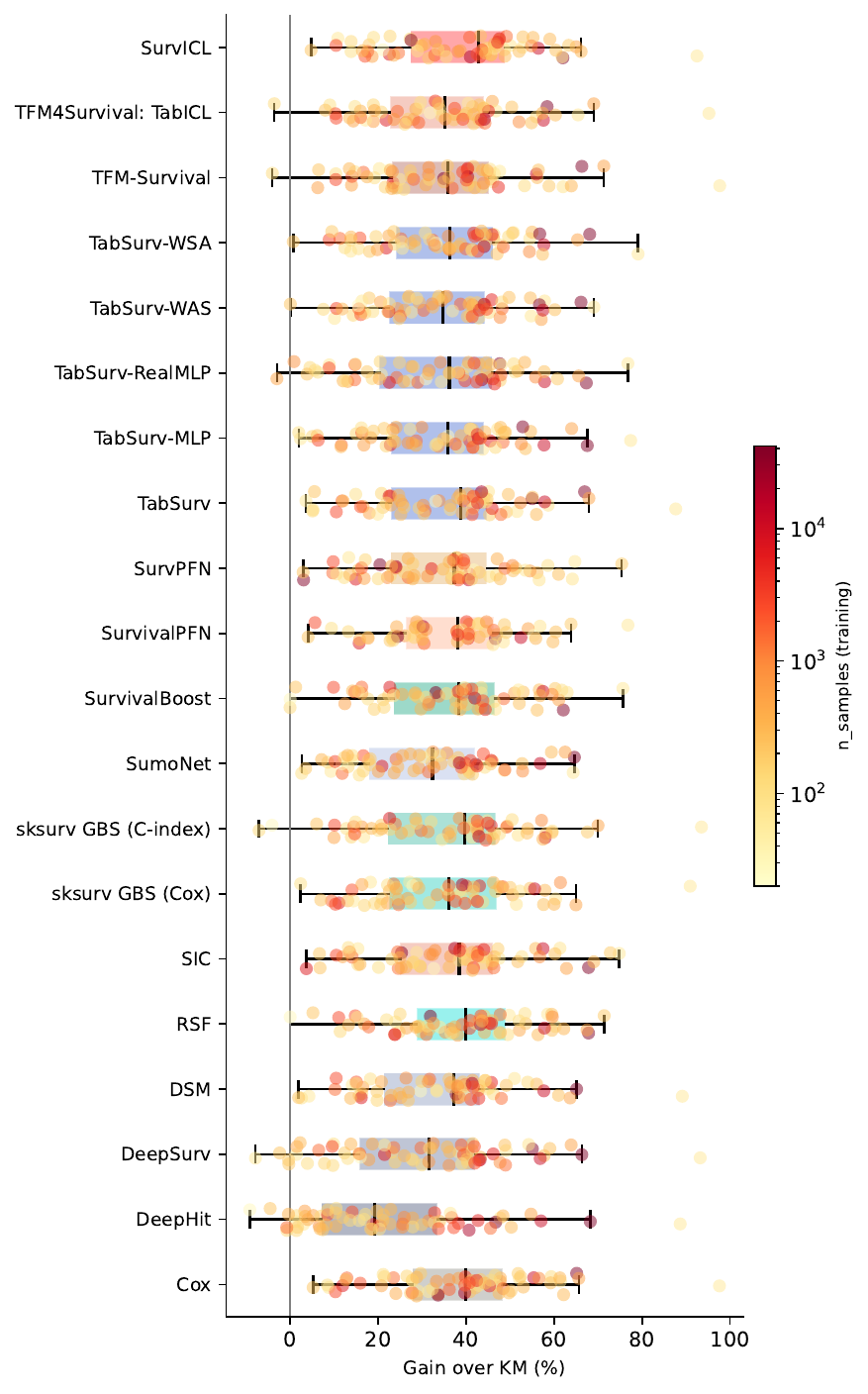}
    \caption{\textbf{C-index on the 50th quantile on all held-out datasets, compared to the score obtained with a Kaplan-Meier estimator (higher is better).} The points show individual dataset performances, and a boxplot summarizes the overall distribution. The dot color represents the dataset size.}
    \label{fig:boxplotcindex}
\end{figure}

\begin{figure}[h!]
    \centering
    \includegraphics[width=\linewidth]{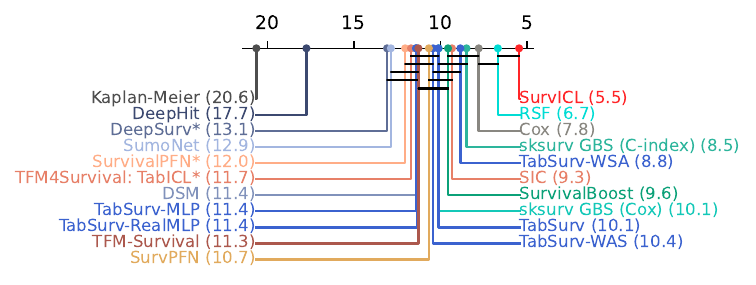}
    \caption{\textbf{Critical difference diagram of average ranks on all held-out datasets (lower is better) for the C-index on the 50th quantile.} Methods joined by a bar are not significantly different.}
    \label{fig:cdplotcindex}
\end{figure}

\begin{figure}[h!]
    \centering
    \includegraphics[width=\linewidth]{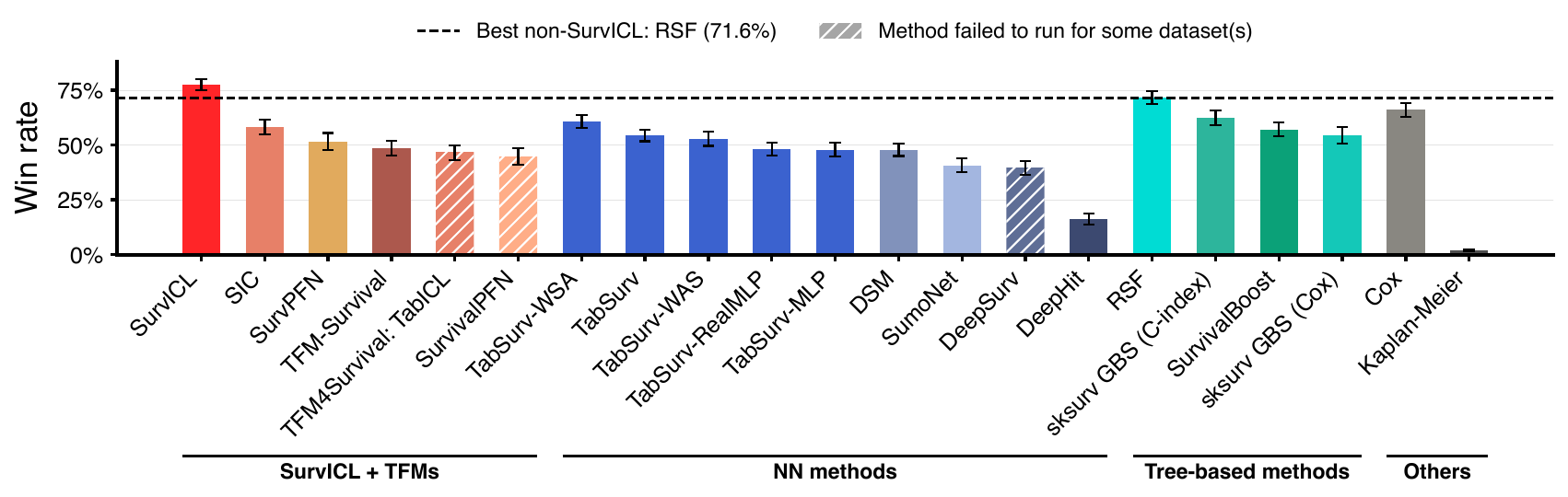}
    \caption{\textbf{Pairwise win rate in the C-index on the 50th quantile on all held-out datasets:} the fraction of (dataset, competitor) pairs on which a method achieves a higher C-index.}
    \label{fig:winplotcindex}
\end{figure}

\begin{figure}[h!]
    \centering
    \includegraphics[width=0.9\linewidth]{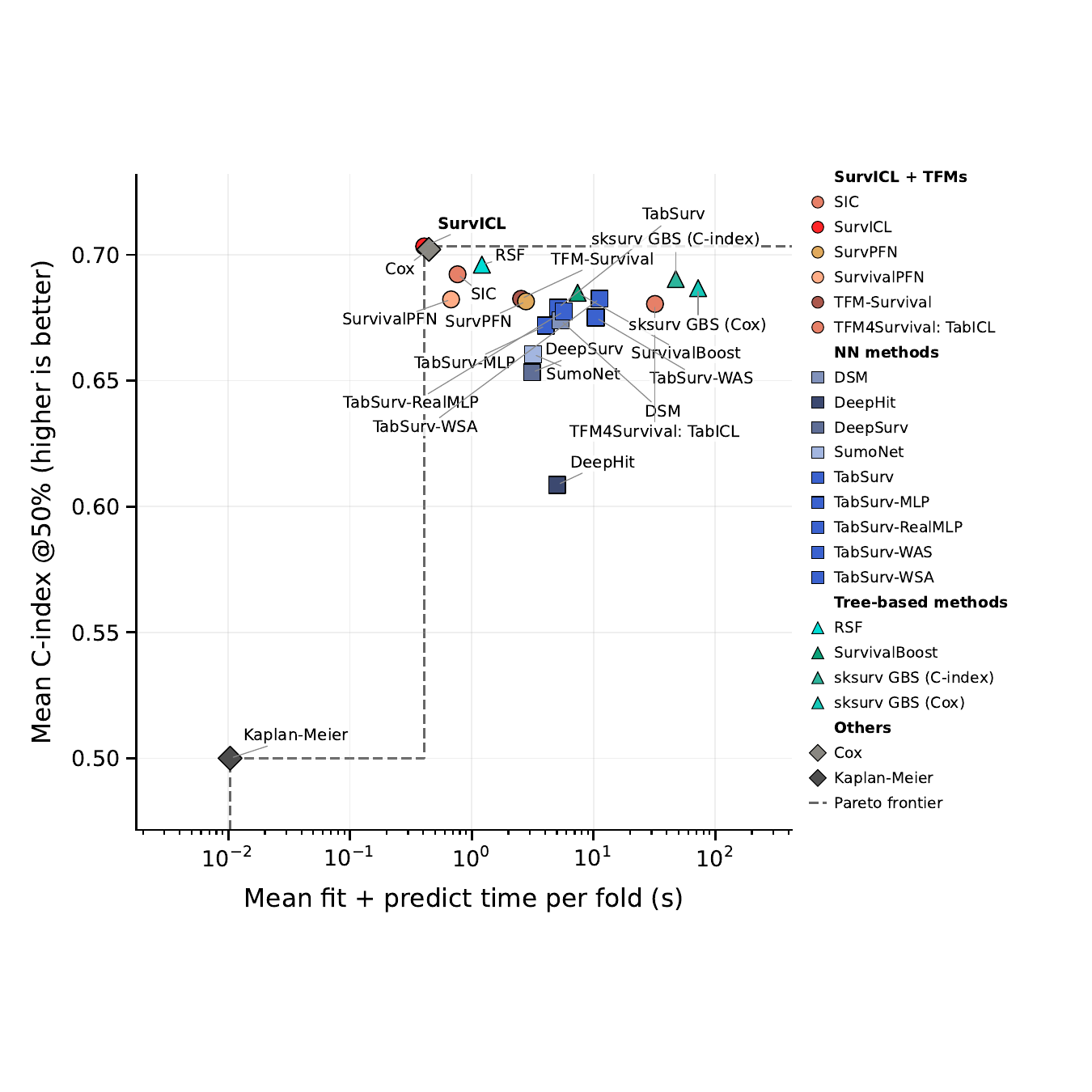}
    \caption{\textbf{Pareto front of mean  C-index on the 50th quantile (higher is better) and fit $+$ predict time.} We report average performances over all held-out datasets.}
    \label{fig:paretocindex}
\end{figure}

Results for the C-index on held-out datasets are shown in Figures \ref{fig:boxplotcindex}, \ref{fig:cdplotcindex}, \ref{fig:winplotcindex}, \ref{fig:paretocindex}.

\FloatBarrier

\subsubsection{KM-calibration}

\begin{figure}[h!]
    \centering
    \includegraphics[width=0.5\linewidth]{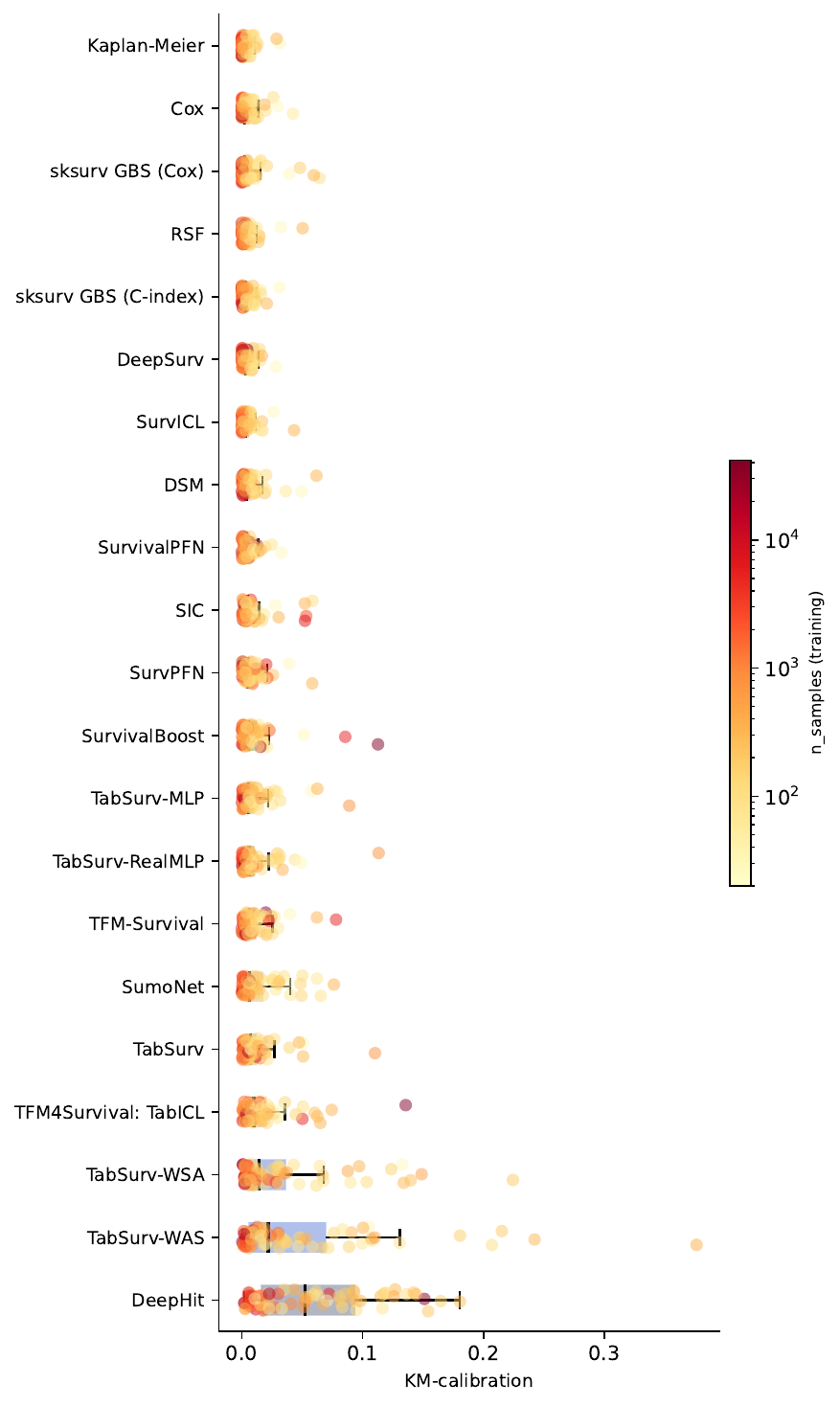}
    \caption{\textbf{KM calibration on all held-out datasets (lower is better).} The points show individual dataset performances, and a boxplot summarizes the overall distribution. The dot color represents the dataset size.}
    \label{fig:boxplotkmcal}
\end{figure}

\begin{figure}[h!]
    \centering
    \includegraphics[width=\linewidth]{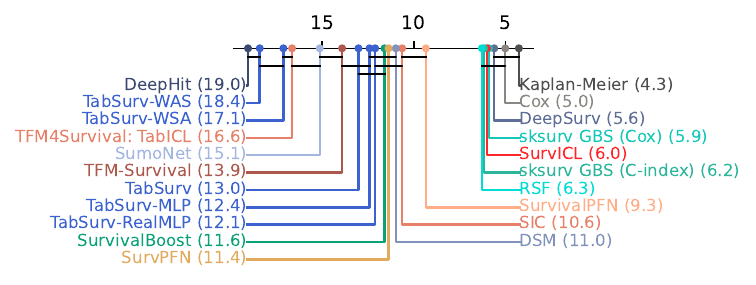}
    \caption{\textbf{Critical difference diagram of average ranks on all held-out datasets (lower is better) for the KM calibration.} Methods joined by a bar are not significantly different.}
    \label{fig:cdplotkmcal}
\end{figure}

\begin{figure}[h!]
    \centering
    \includegraphics[width=\linewidth]{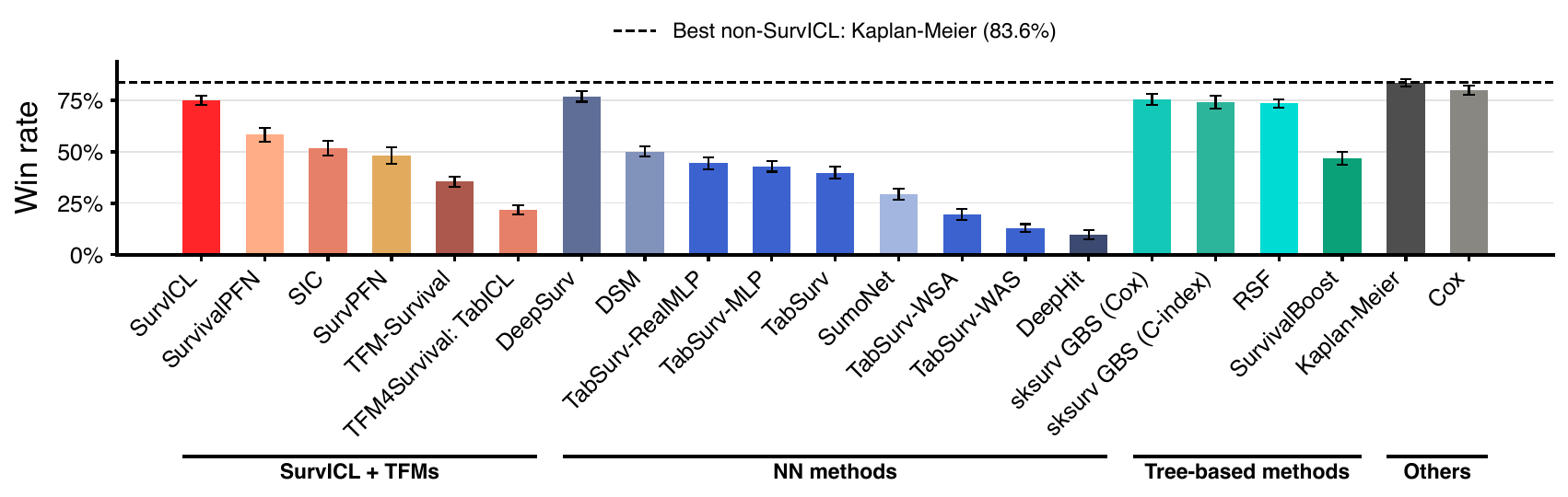}
    \caption{\textbf{Pairwise win rate in KM calibration on all held-out datasets:} the fraction of (dataset, competitor) pairs on which a method achieves a better KM calibration.}
    \label{fig:winplotkmcal}
\end{figure}

Results for the KM-calibration on held-out datasets are shown in Figures \ref{fig:boxplotkmcal}, \ref{fig:cdplotkmcal}, \ref{fig:winplotkmcal}.

\FloatBarrier
\newpage

\newpage
\subsubsection{MAE}

\begin{figure}[h!]
    \centering
    \includegraphics[width=0.5\linewidth]{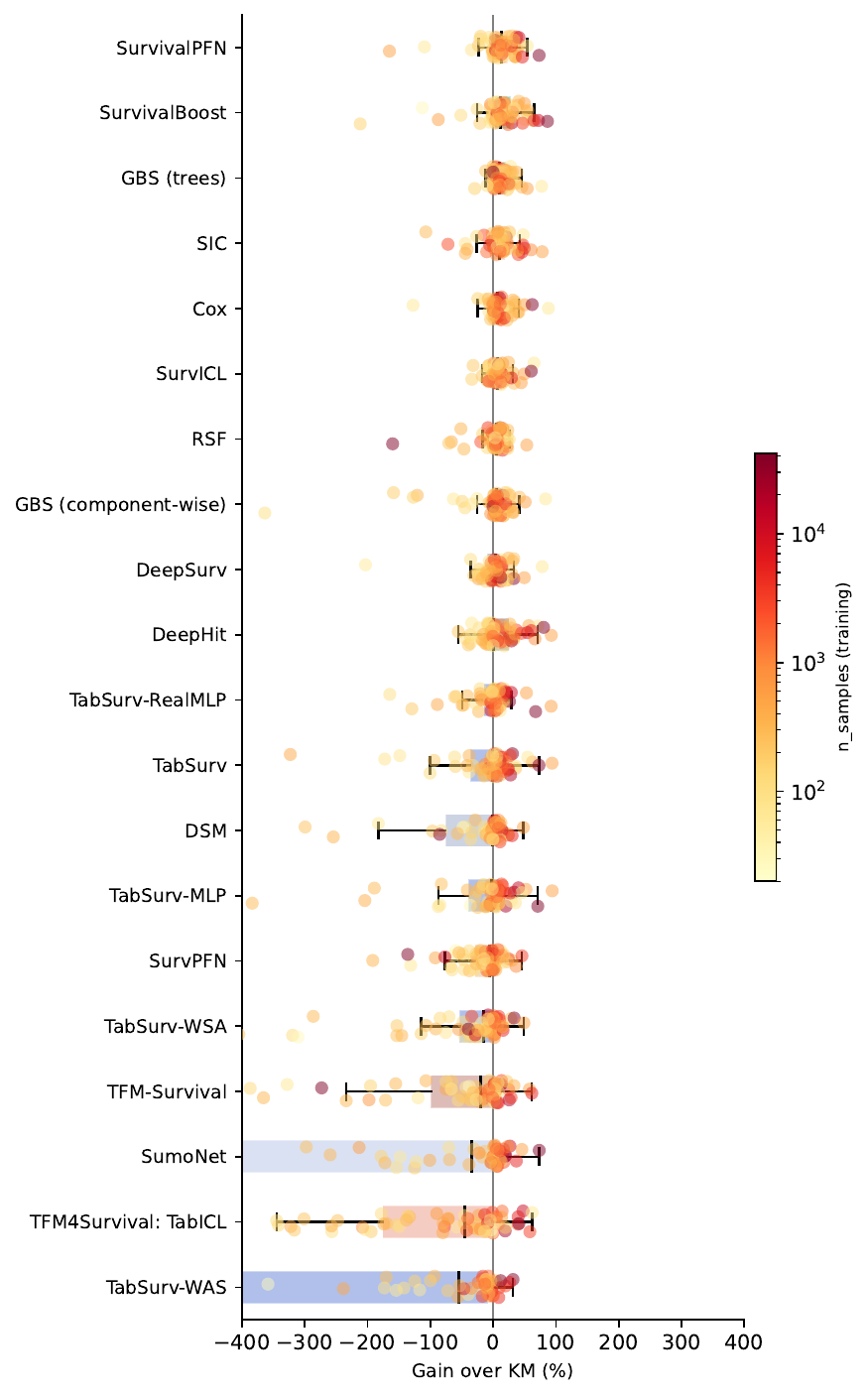}
    \caption{\textbf{MAE on all held-out datasets, compared to the MAE obtained with a Kaplan-Meier estimator (higher is better).} The points show individual dataset performances, and a boxplot summarizes the overall distribution. The dot color represents the dataset size.}
    \label{fig:boxplotmae}
\end{figure}

\begin{figure}[h!]
    \centering
    \includegraphics[width=\linewidth]{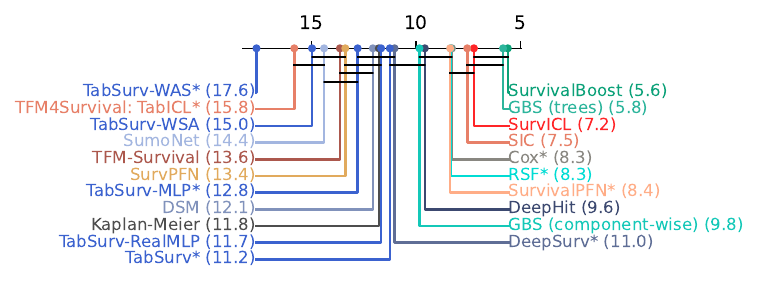}
    \caption{\textbf{Critical difference diagram of average ranks on all held-out datasets (lower is better) for the MAE.} Methods joined by a bar are not significantly different}
    \label{fig:cdplotmae}
\end{figure}

\begin{figure}[h!]
    \centering
    \includegraphics[width=\linewidth]{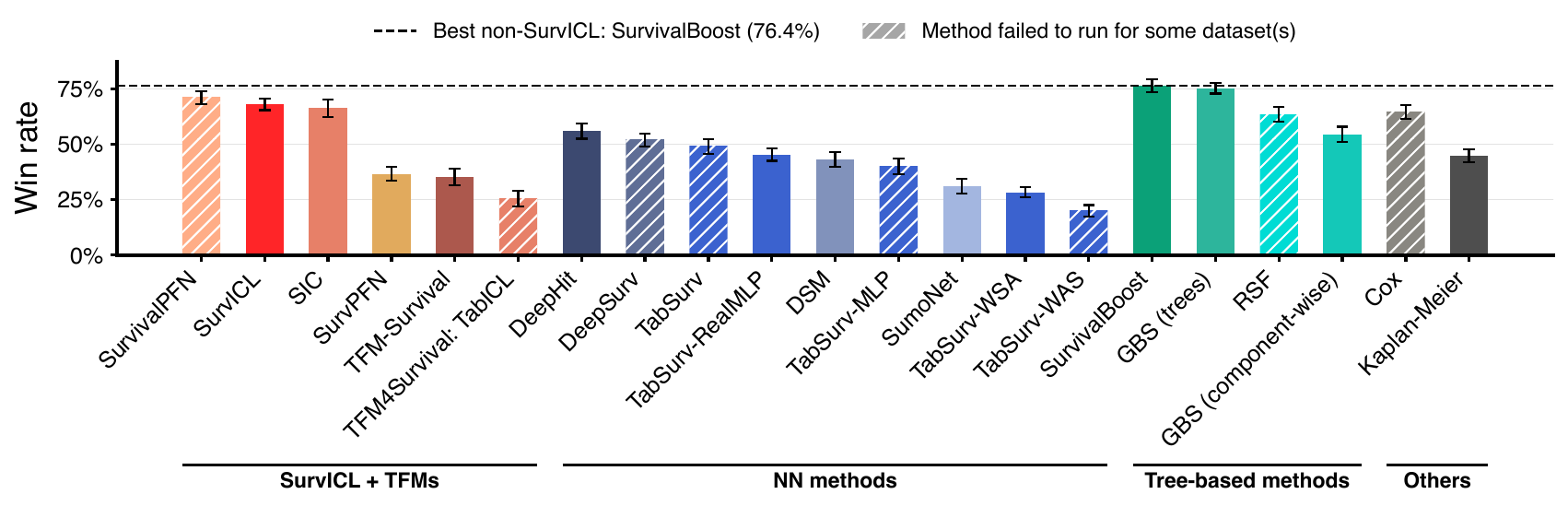}
    \caption{\textbf{Pairwise win rate in MAE on all held-out datasets:} the fraction of (dataset, competitor) pairs on which a method achieves a lower MAE.}
    \label{fig:winplotmae}
\end{figure}

\begin{figure}[h!]
    \centering
    \includegraphics[width=0.9\linewidth]{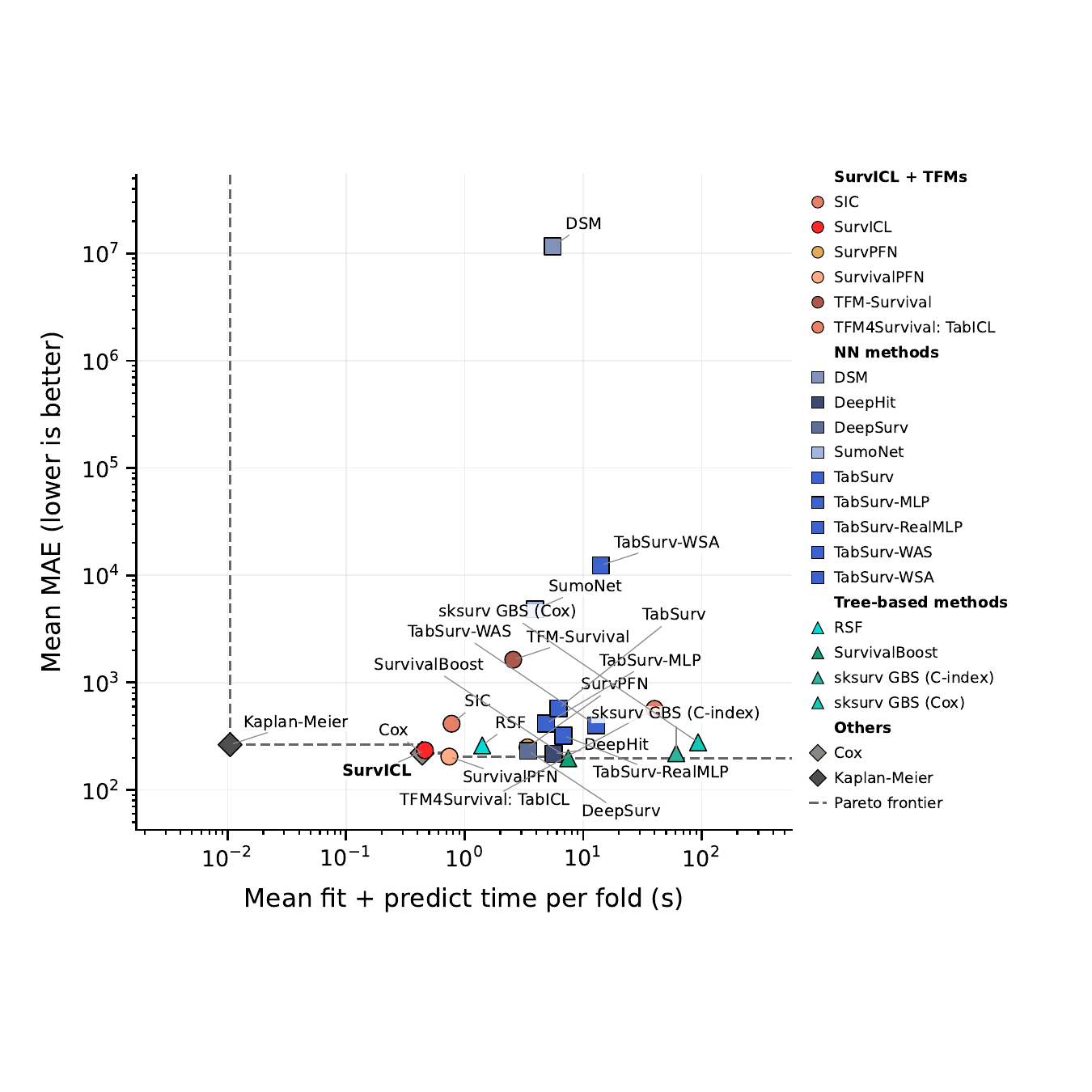}
    \caption{\textbf{Pareto front of mean MAE (lower is better) and fit $+$ predict time.} We report average performances over all held-out datasets.}
    \label{fig:paretomae}
\end{figure}

Results for the MAE on held-out datasets are shown in Figures \ref{fig:boxplotmae}, \ref{fig:cdplotmae}, \ref{fig:winplotmae}, \ref{fig:paretomae}.

\FloatBarrier
\newpage

\newpage
\subsection{Held-out datasets with less than 1000 samples}
\subsubsection{IBS}

\begin{figure}[h!]
    \centering
    \includegraphics[width=0.5\linewidth]{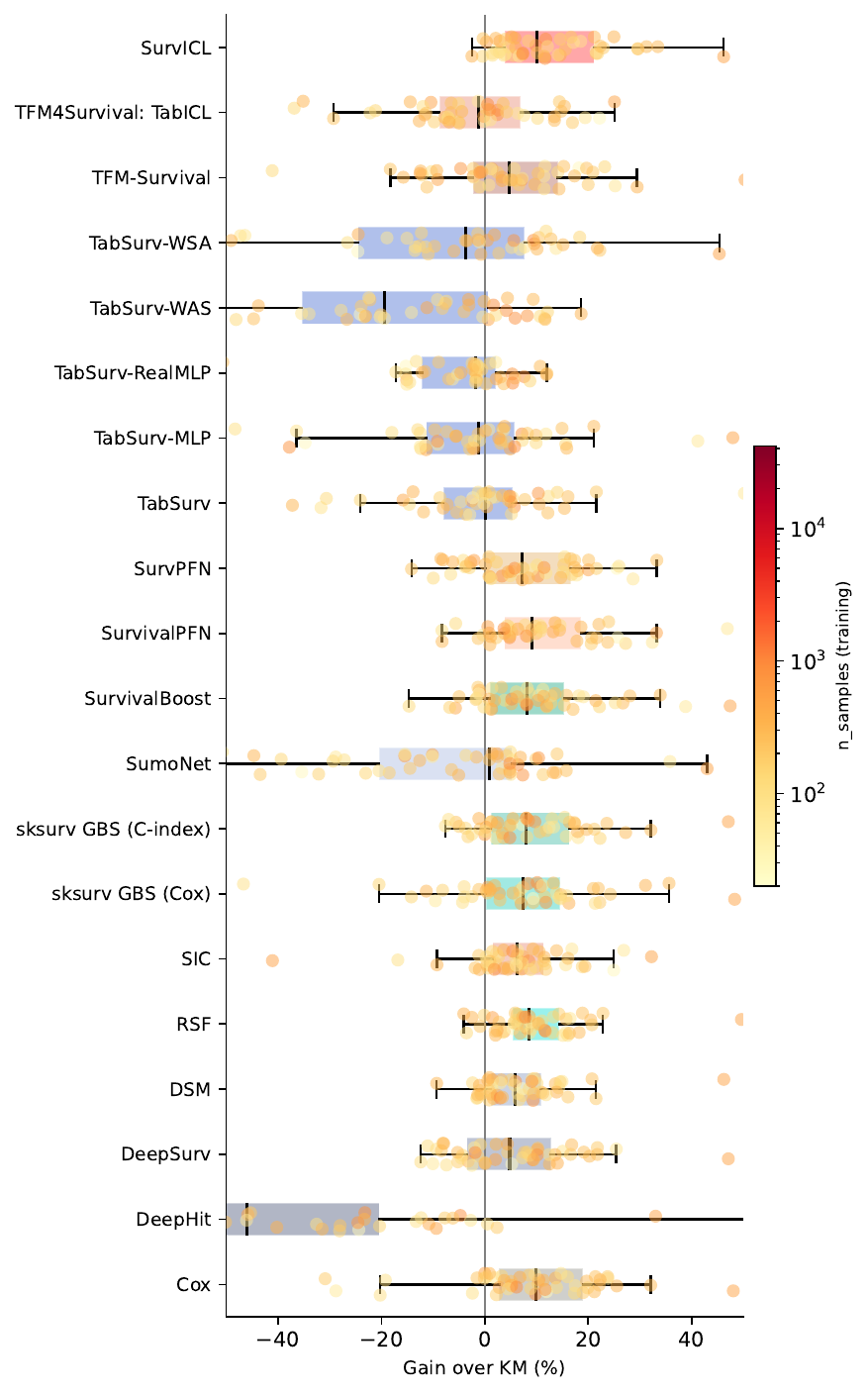}
    \caption{\textbf{IBS on the small held-out datasets, compared to the IBS score obtained with a Kaplan-Meier estimator (higher is better).} The points show individual dataset performances, and a boxplot summarizes the overall distribution. The dot color represents the dataset size.}
    \label{fig:boxplotibssmall}
\end{figure}

\begin{figure}[h!]
    \centering
    \includegraphics[width=\linewidth]{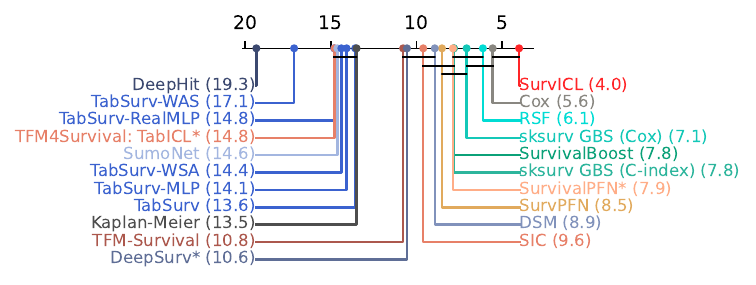}
    \caption{\textbf{Critical difference diagram of average ranks on the small held-out datasets (lower is better) for the IBS.} Methods joined by a bar are not significantly different.}
    \label{fig:cdplotibssmall}
\end{figure}

\begin{figure}[h!]
    \centering
    \includegraphics[width=\linewidth]{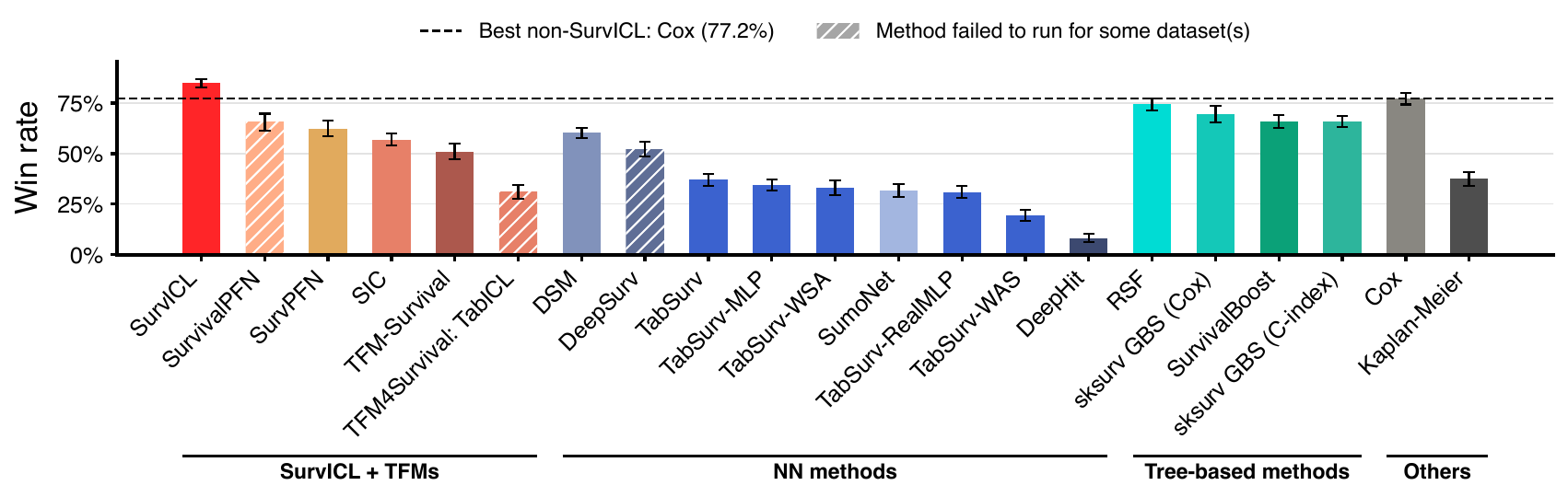}
    \caption{\textbf{Pairwise win rate in IBS on the small held-out datasets:} the fraction of (dataset, competitor) pairs on which a method achieves a lower IBS.}
    \label{fig:winplotibssmall}
\end{figure}

\begin{figure}[h!]
    \centering
    \includegraphics[width=0.9\linewidth]{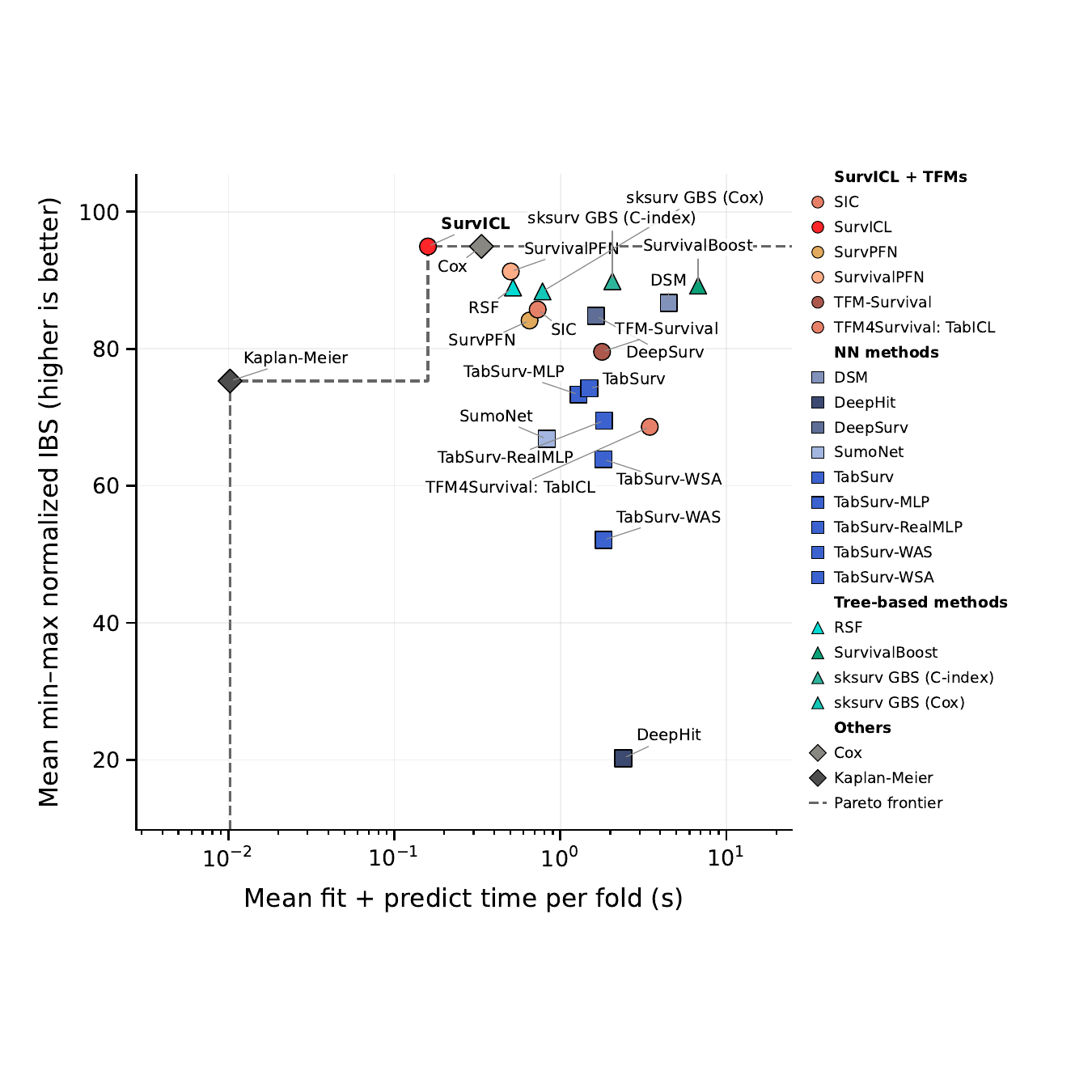}
    \caption{\textbf{Pareto front of mean normalized IBS (higher is better) and fit $+$ predict time.} We report average performances over the small held-out datasets. IBS was normalized to provide comparability across all datasets.}
    \label{fig:paretoibssmall}
\end{figure}

Results for the IBS on small held-out datasets are shown in Figures \ref{fig:boxplotibssmall}, \ref{fig:cdplotibssmall}, \ref{fig:winplotibssmall}, \ref{fig:paretoibssmall}.

\FloatBarrier
\newpage

\newpage
\subsubsection{C-index50}

\begin{figure}[h!]
    \centering
    \includegraphics[width=0.5\linewidth]{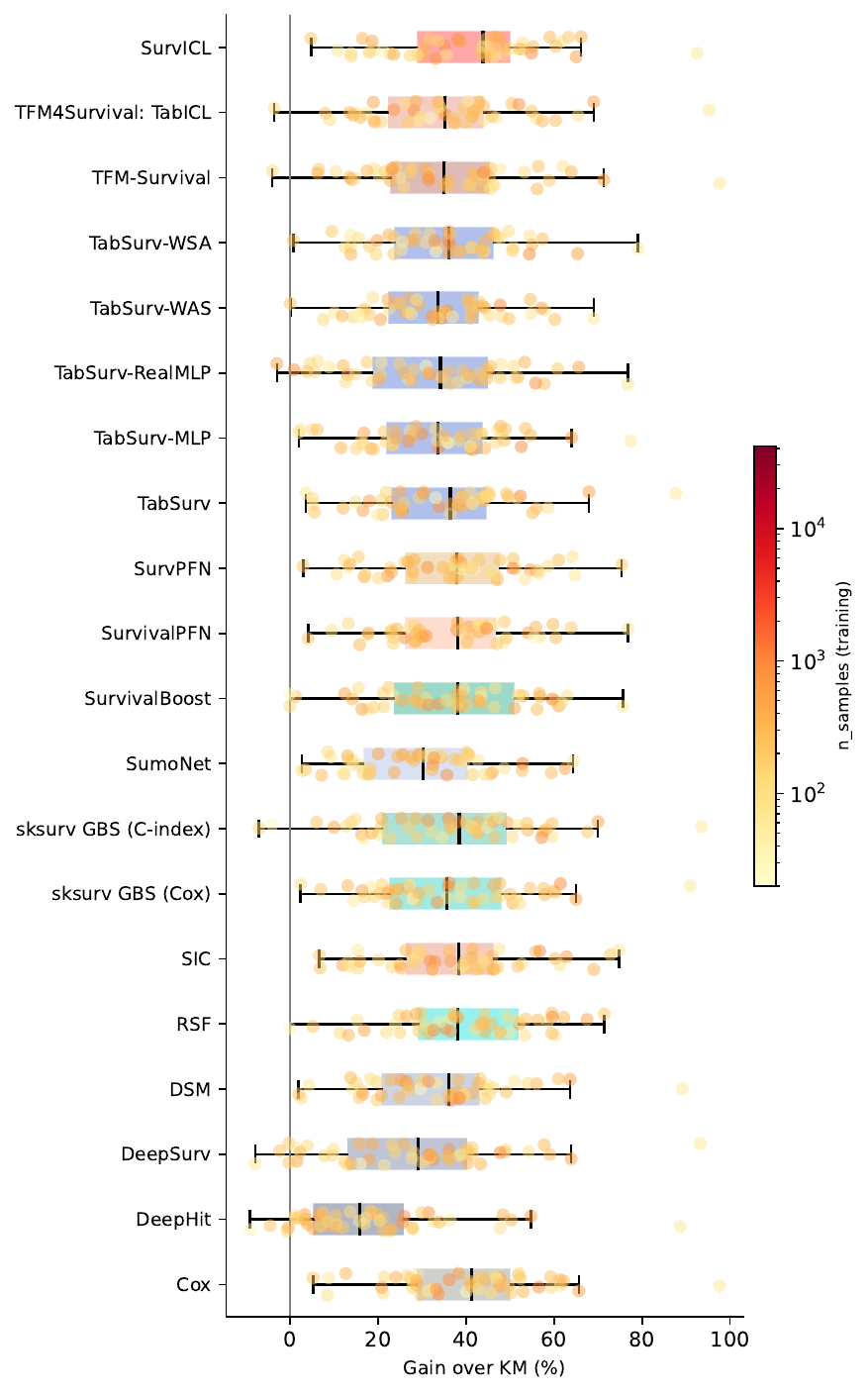}
    \caption{\textbf{C-index at the 50th quantile on the small held-out datasets, compared to the C-index score obtained with a Kaplan-Meier estimator (higher is better).} The points show individual dataset performances, and a boxplot summarizes the overall distribution. The dot color represents the dataset size.}
    \label{fig:boxplotcindexsmall}
\end{figure}

\begin{figure}[h!]
    \centering
    \includegraphics[width=\linewidth]{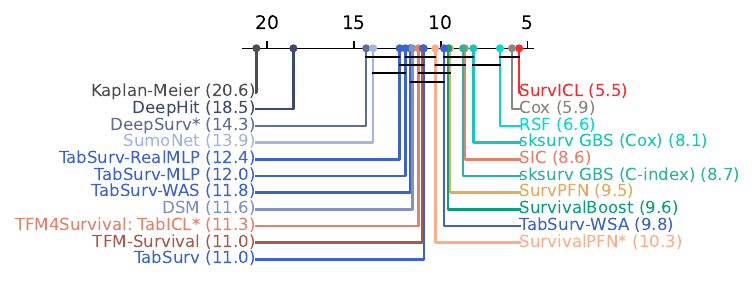}
    \caption{\textbf{Critical difference diagram of average ranks on the small held-out datasets (lower is better) for the C-index at the 50th quantile.} Methods joined by a bar are not significantly different.}
    \label{fig:cdplotcindexsmall}
\end{figure}

\begin{figure}[h!]
    \centering
    \includegraphics[width=\linewidth]{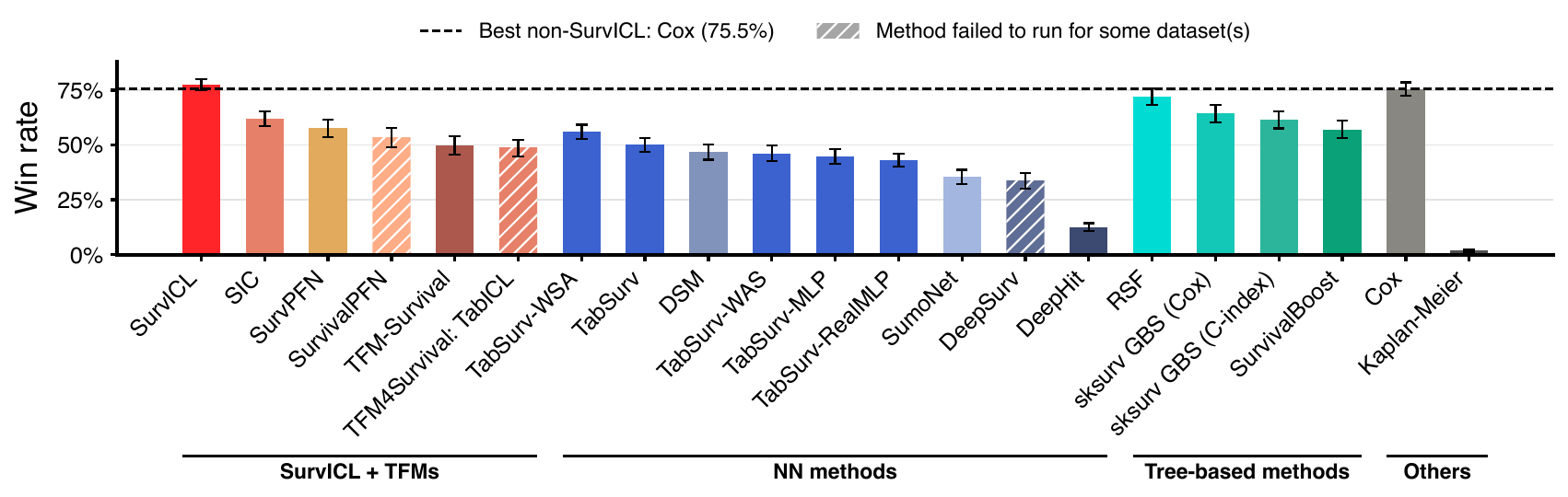}
    \caption{\textbf{Pairwise win rate in the C-index at the 50th quantile on the small held-out datasets:} the fraction of (dataset, competitor) pairs on which a method achieves a higher C-index.}
    \label{fig:winplotcindexsmall}
\end{figure}

\begin{figure}[h!]
    \centering
    \includegraphics[width=0.9\linewidth]{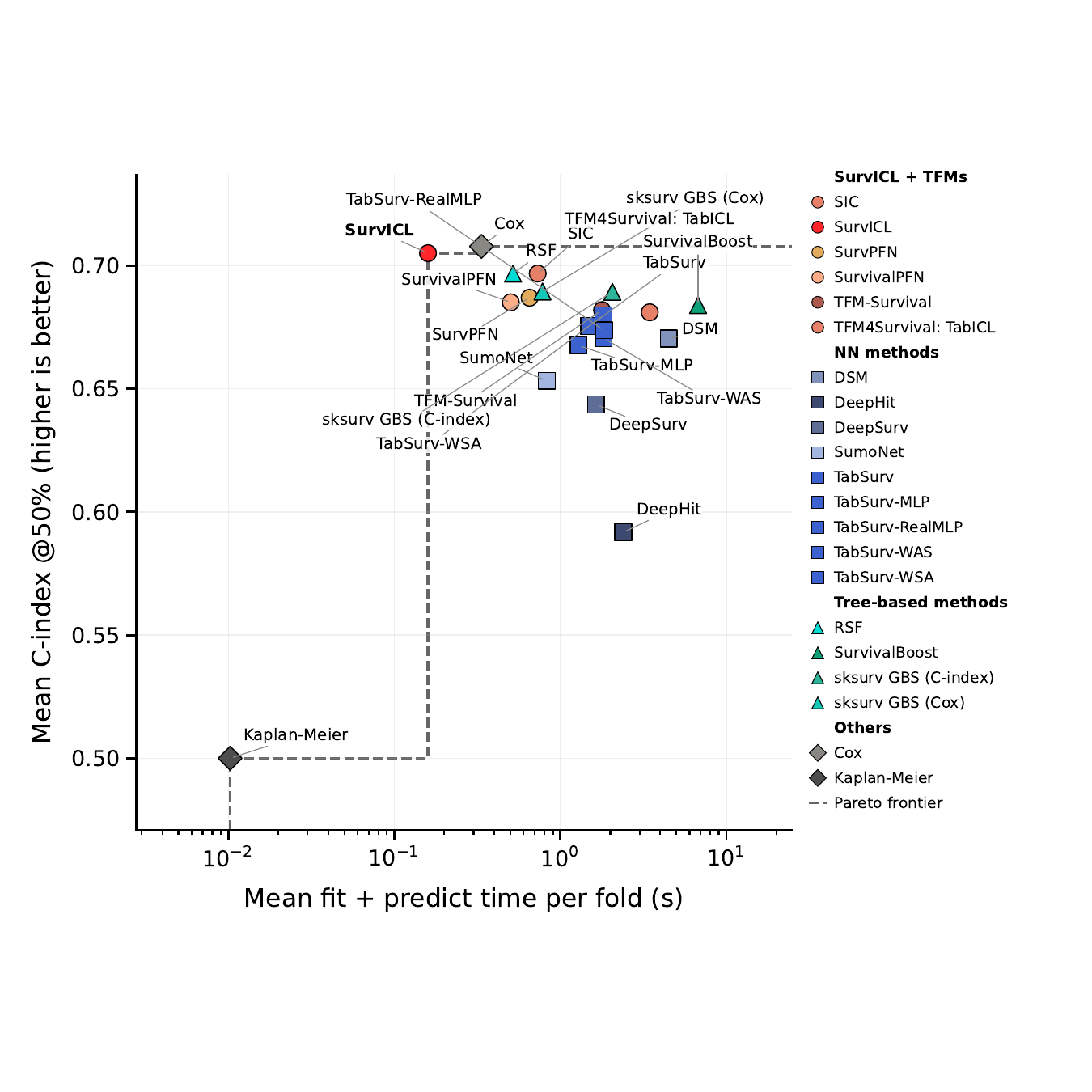}
    \caption{\textbf{Pareto front of mean  C-index on the 50th quantile (higher is better) and fit $+$ predict time.} We report average performances over the small held-out datasets.}
    \label{fig:paretocindexsmall}
\end{figure}

Results for the C-index at the 50th quantile on small held-out datasets are shown in Figures \ref{fig:boxplotcindexsmall}, \ref{fig:cdplotcindexsmall}, \ref{fig:winplotcindexsmall}, \ref{fig:paretocindexsmall}.

\FloatBarrier
\newpage

\newpage
\subsubsection{KM-calibration}

\begin{figure}[h!]
    \centering
    \includegraphics[width=0.5\linewidth]{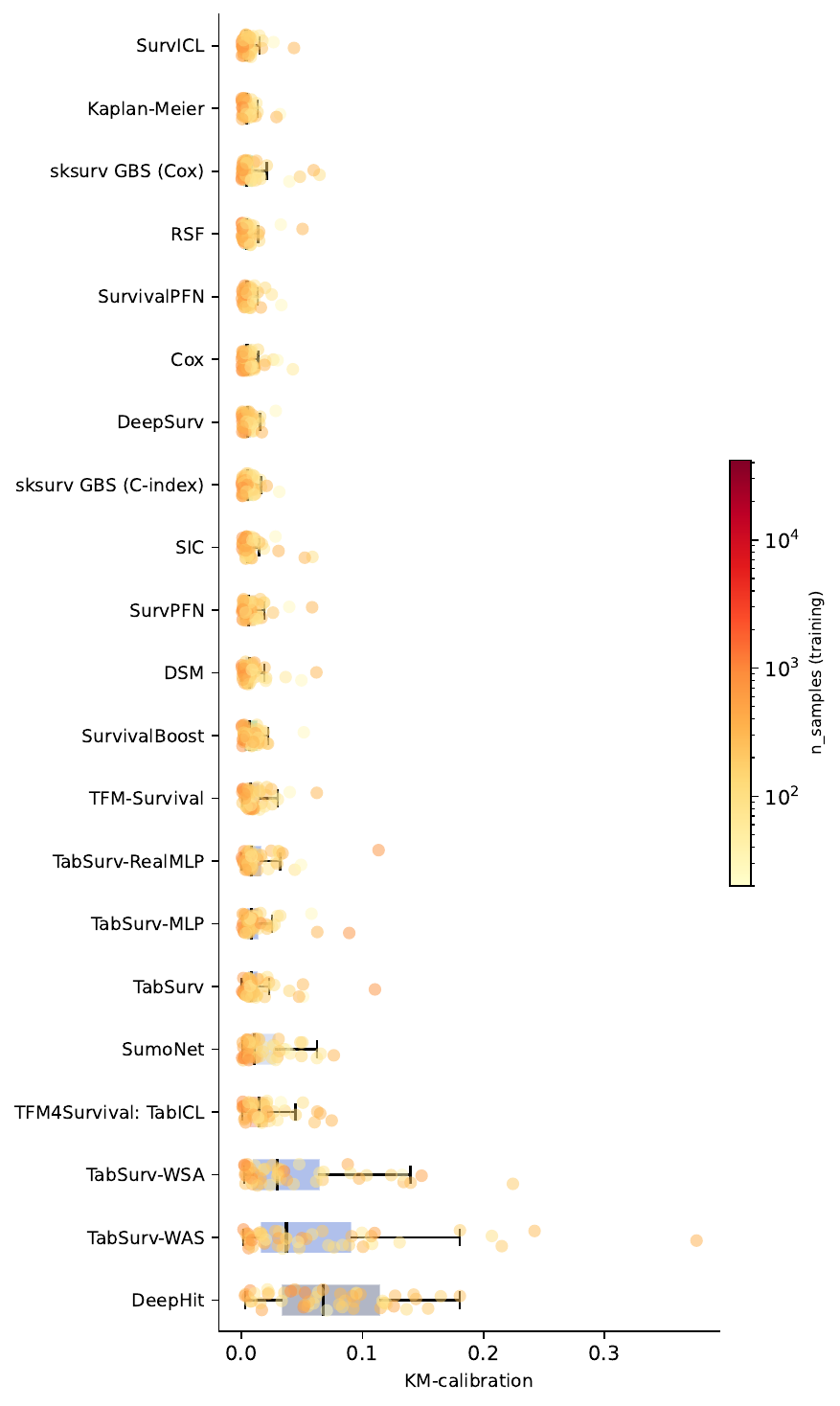}
    \caption{\textbf{KM calibration on the small held-out datasets (lower is better).} The points show individual dataset performances, and a boxplot summarizes the overall distribution. The dot color represents the dataset size.}
    \label{fig:boxplotkmcalsmall}
\end{figure}

\begin{figure}[h!]
    \centering
    \includegraphics[width=\linewidth]{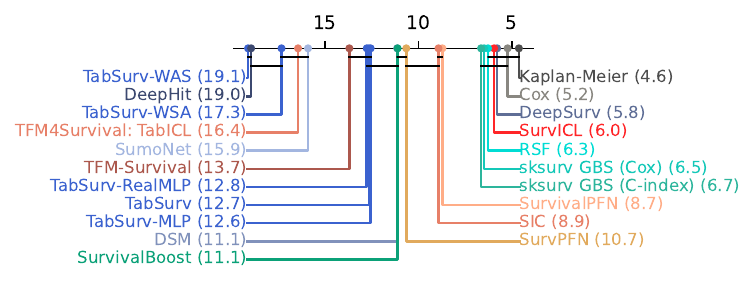}
    \caption{\textbf{Critical difference diagram of average ranks on the small held-out datasets (lower is better) for the KM calibration.} Methods joined by a bar are not significantly different.}
    \label{fig:cdplotkmcalsmall}
\end{figure}

\begin{figure}[h!]
    \centering
    \includegraphics[width=\linewidth]{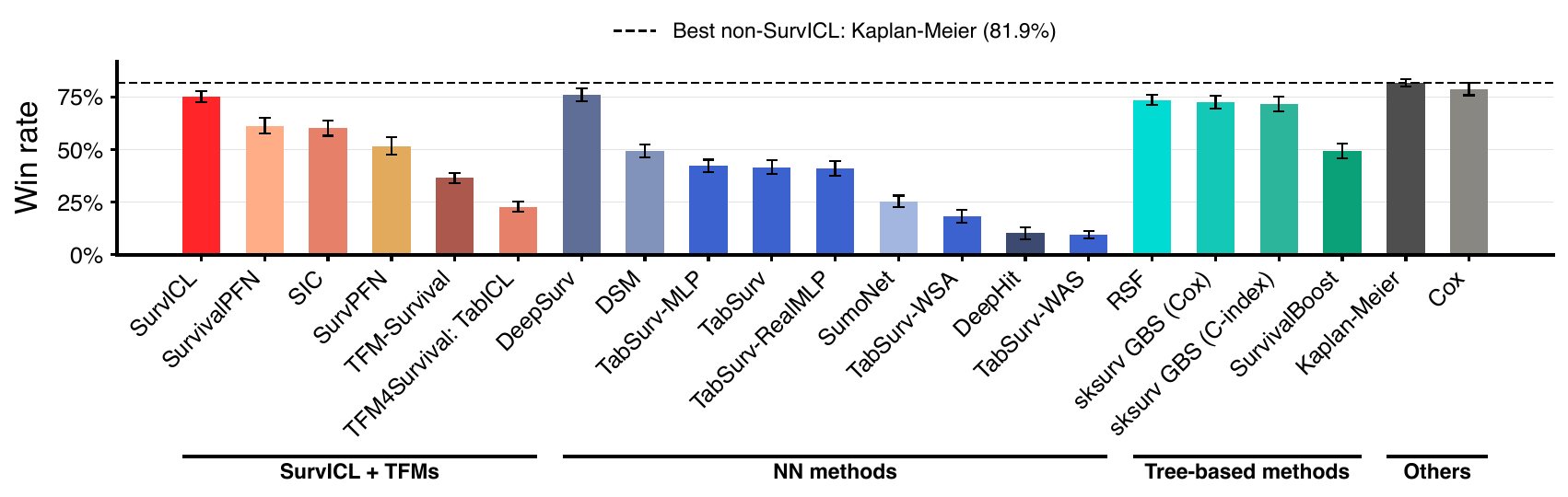}
    \caption{\textbf{Pairwise win rate in the KM-calibration on the small held-out datasets:} the fraction of (dataset, competitor) pairs on which a method achieves a better KM-calibration.}
    \label{fig:winplotkmcalsmall}
\end{figure}

Results for the KM-calibration on small held-out datasets are shown in Figures \ref{fig:boxplotkmcalsmall}, \ref{fig:cdplotkmcalsmall}, \ref{fig:winplotkmcalsmall}.

\FloatBarrier
\newpage

\newpage
\subsubsection{MAE}

\begin{figure}[h!]
    \centering
    \includegraphics[width=0.5\linewidth]{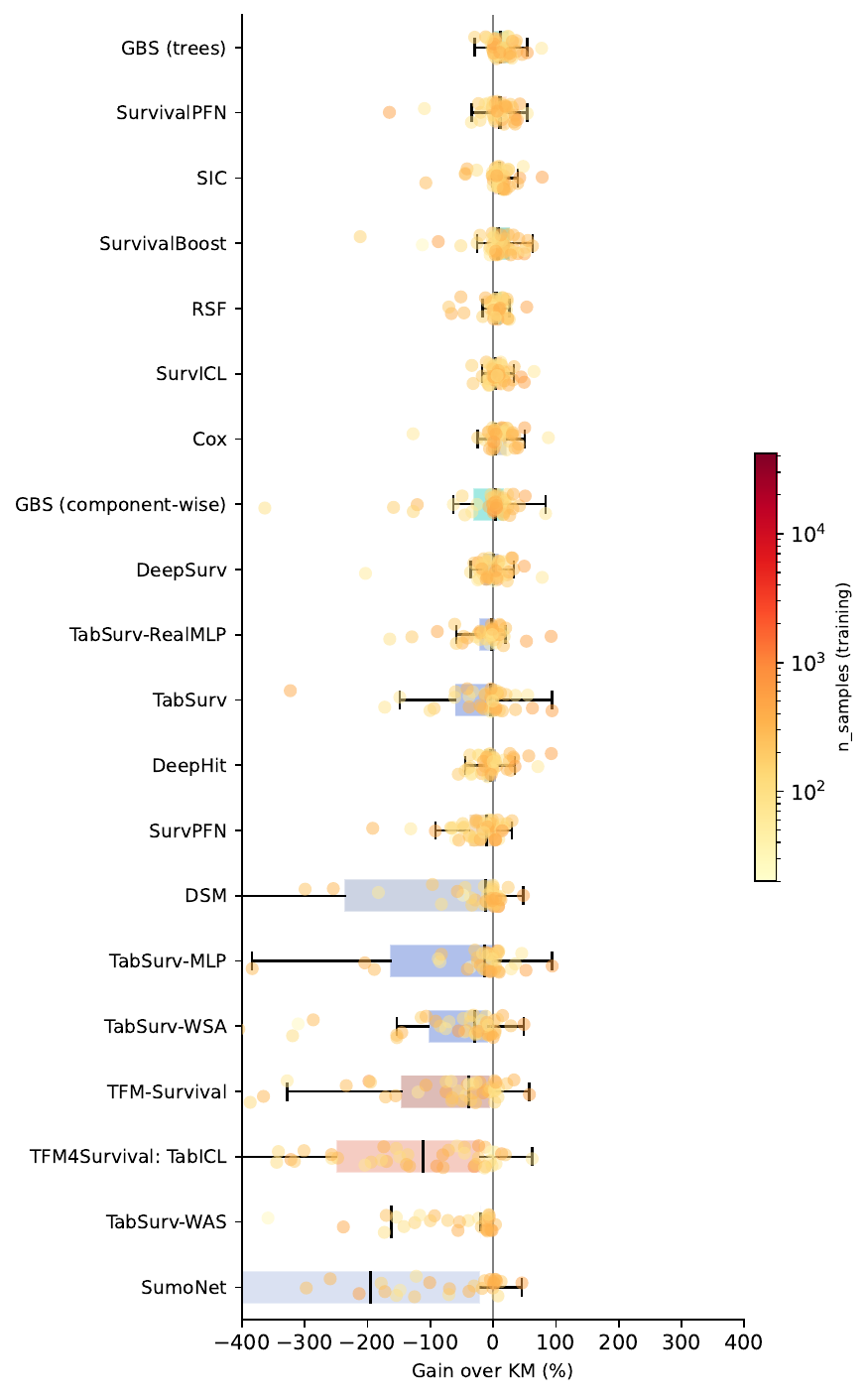}
    \caption{\textbf{MAE on the small held-out datasets, compared to the MAE obtained with a Kaplan-Meier estimator (higher is better).} The points show individual dataset performances, and a boxplot summarizes the overall distribution. The dot color represents the dataset size.}
    \label{fig:boxplotmaesmall}
\end{figure}

\begin{figure}[h!]
    \centering
    \includegraphics[width=\linewidth]{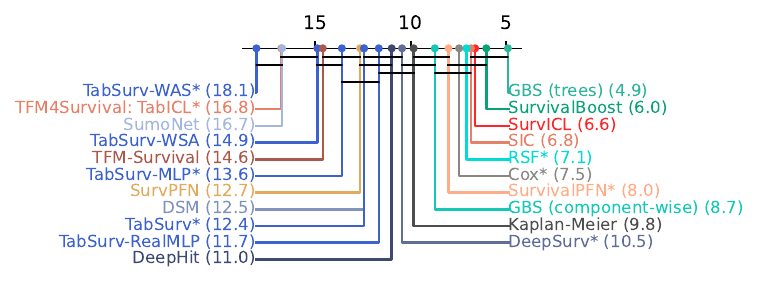}
    \caption{\textbf{Critical difference diagram of average ranks on the small held-out datasets (lower is better) for the MAE.} Methods joined by a bar are not significantly different.}
    \label{fig:cdplotmaesmall}
\end{figure}

\begin{figure}[h!]
    \centering
    \includegraphics[width=\linewidth]{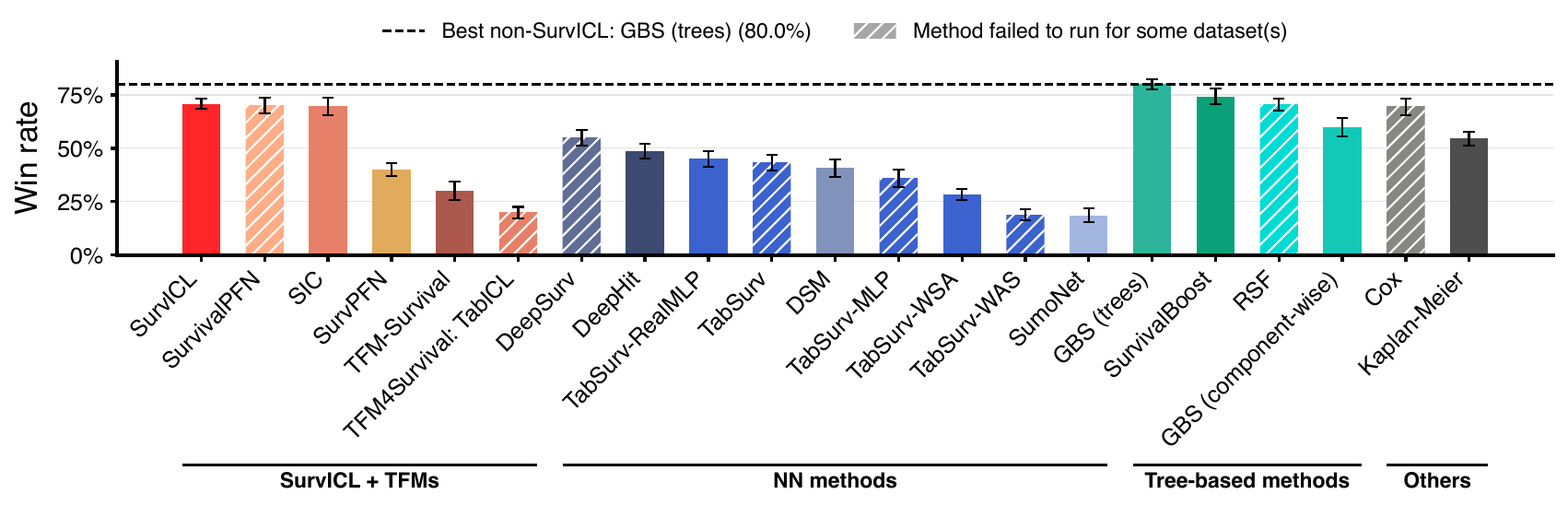}
    \caption{\textbf{Pairwise win rate in the MAE on the small held-out datasets:} the fraction of (dataset, competitor) pairs on which a method achieves a lower MAE.}
    \label{fig:winplotmaesmall}
\end{figure}

\begin{figure}[h!]
    \centering
    \includegraphics[width=0.9\linewidth]{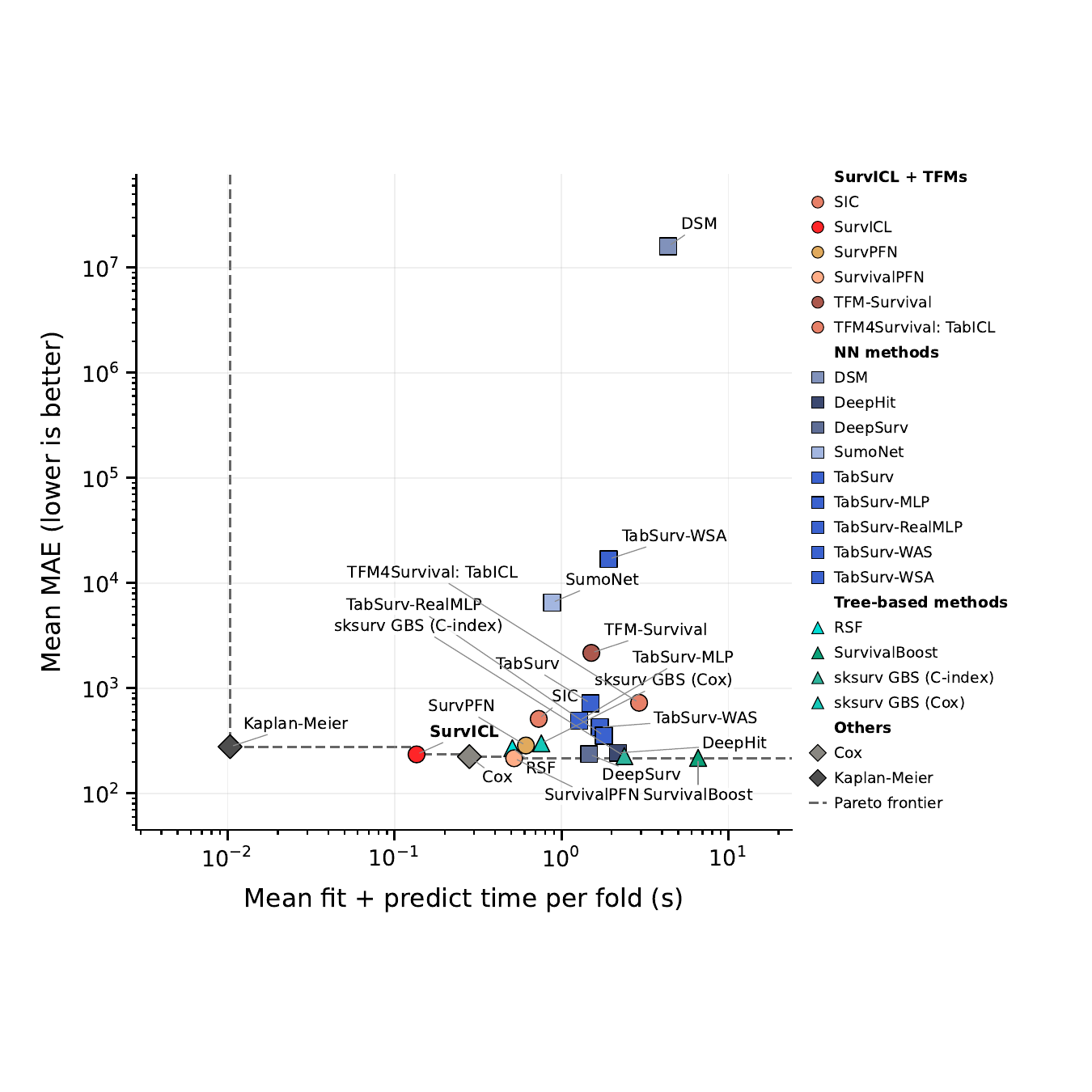}
    \caption{\textbf{Pareto front of mean MAE (lower is better) and fit $+$ predict time.} We report average performances over the small held-out datasets.}
    \label{fig:paretomaesmall}
\end{figure}

Results for the MAE on small held-out datasets are shown in Figures \ref{fig:boxplotmaesmall}, \ref{fig:cdplotmaesmall}, \ref{fig:winplotmaesmall}, \ref{fig:paretomaesmall}.

\FloatBarrier
\newpage

\newpage
\subsection{Held-out datasets with more than 1000 samples }
\subsubsection{IBS}

\begin{figure}[h!]
    \centering
    \includegraphics[width=0.5\linewidth]{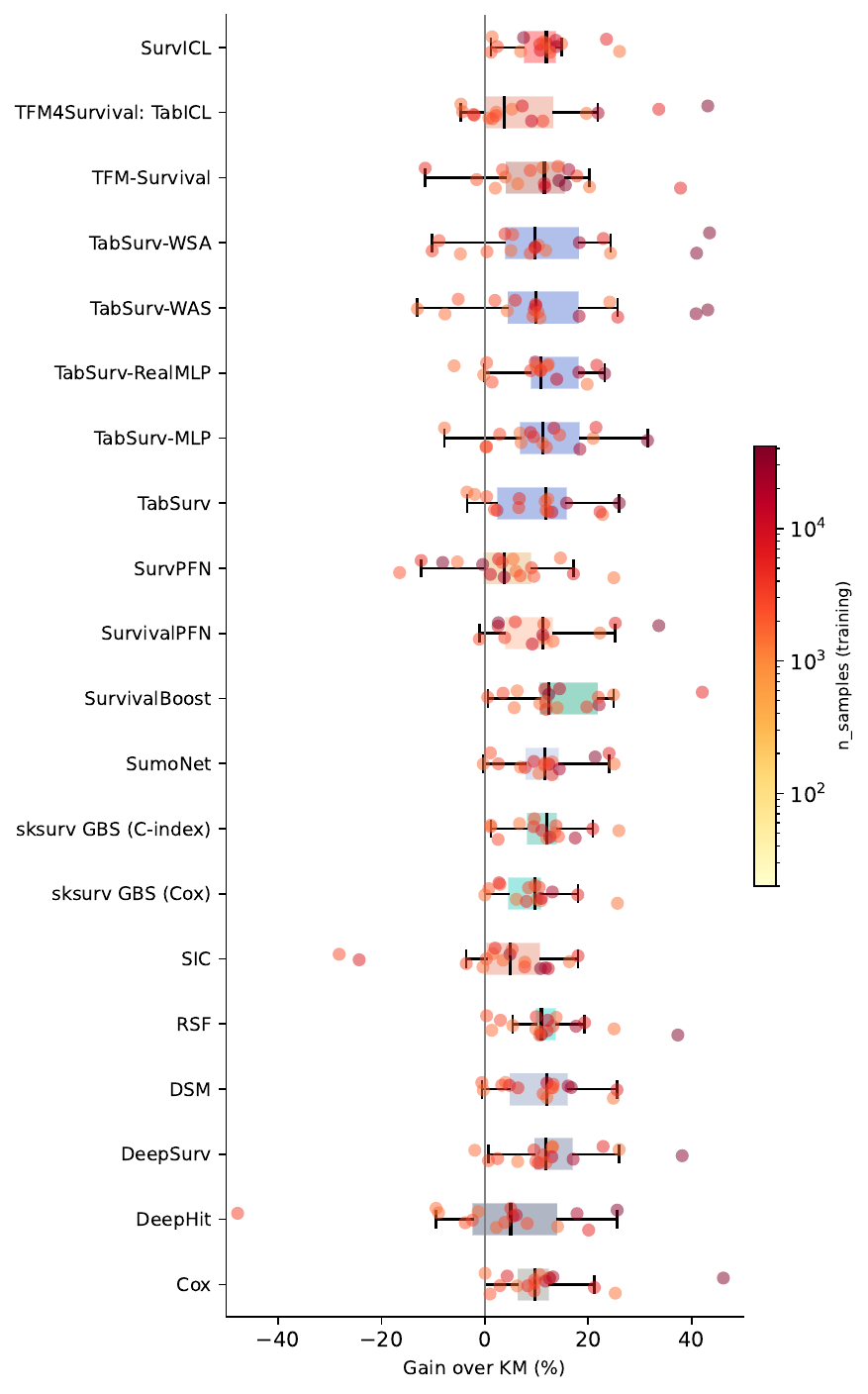}
    \caption{\textbf{IBS on the large held-out datasets, compared to the IBS obtained with a Kaplan-Meier estimator (higher is better).} The points show individual dataset performances, and a boxplot summarizes the overall distribution. The dot color represents the dataset size.}
    \label{fig:boxplotibslarge}
\end{figure}

\begin{figure}[h!]
    \centering
    \includegraphics[width=\linewidth]{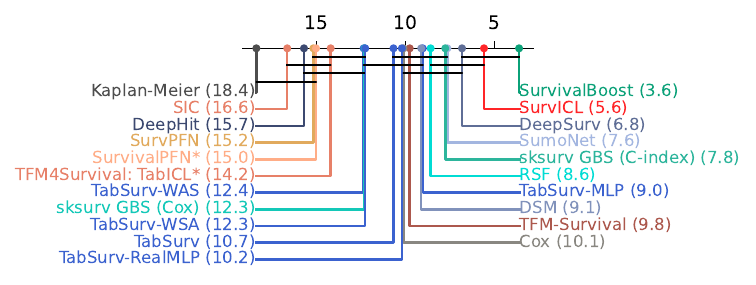}
    \caption{\textbf{Critical difference diagram of average ranks on the large held-out datasets (lower is better) for the IBS.} Methods joined by a bar are not significantly different.}
    \label{fig:cdplotibslarge}
\end{figure}

\begin{figure}[h!]
    \centering
    \includegraphics[width=\linewidth]{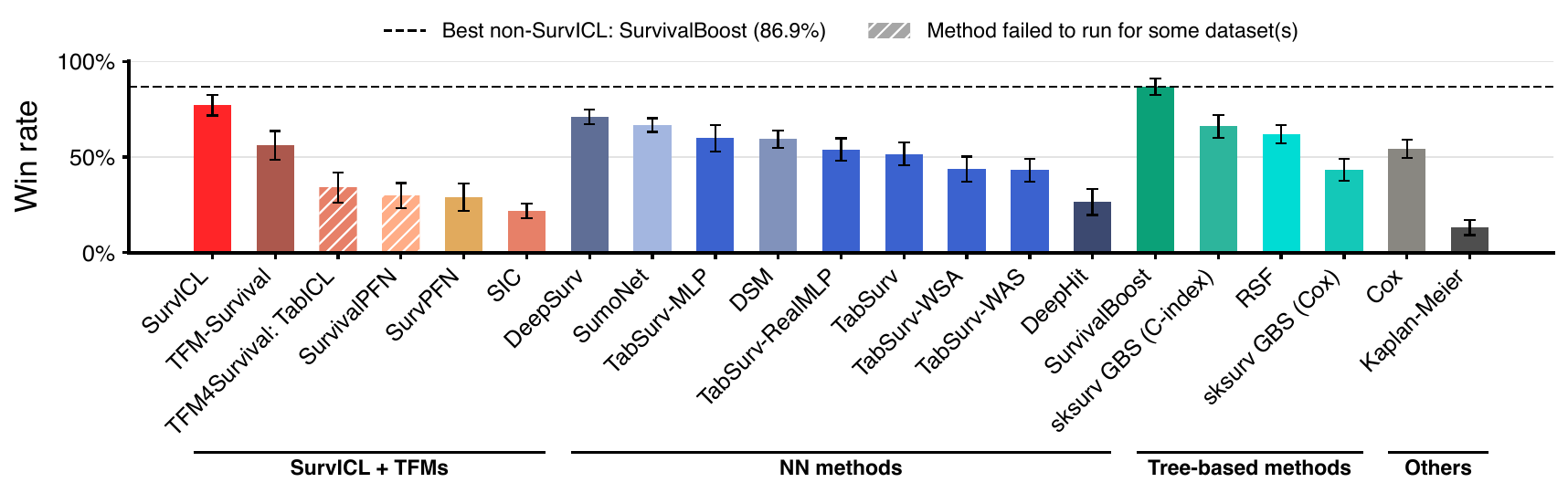}
    \caption{\textbf{Pairwise win rate in the IBS on the large held-out datasets:} the fraction of (dataset, competitor) pairs on which a method achieves a lower IBS.}
    \label{fig:winplotibslarge}
\end{figure}

\begin{figure}[h!]
    \centering
    \includegraphics[width=0.9\linewidth]{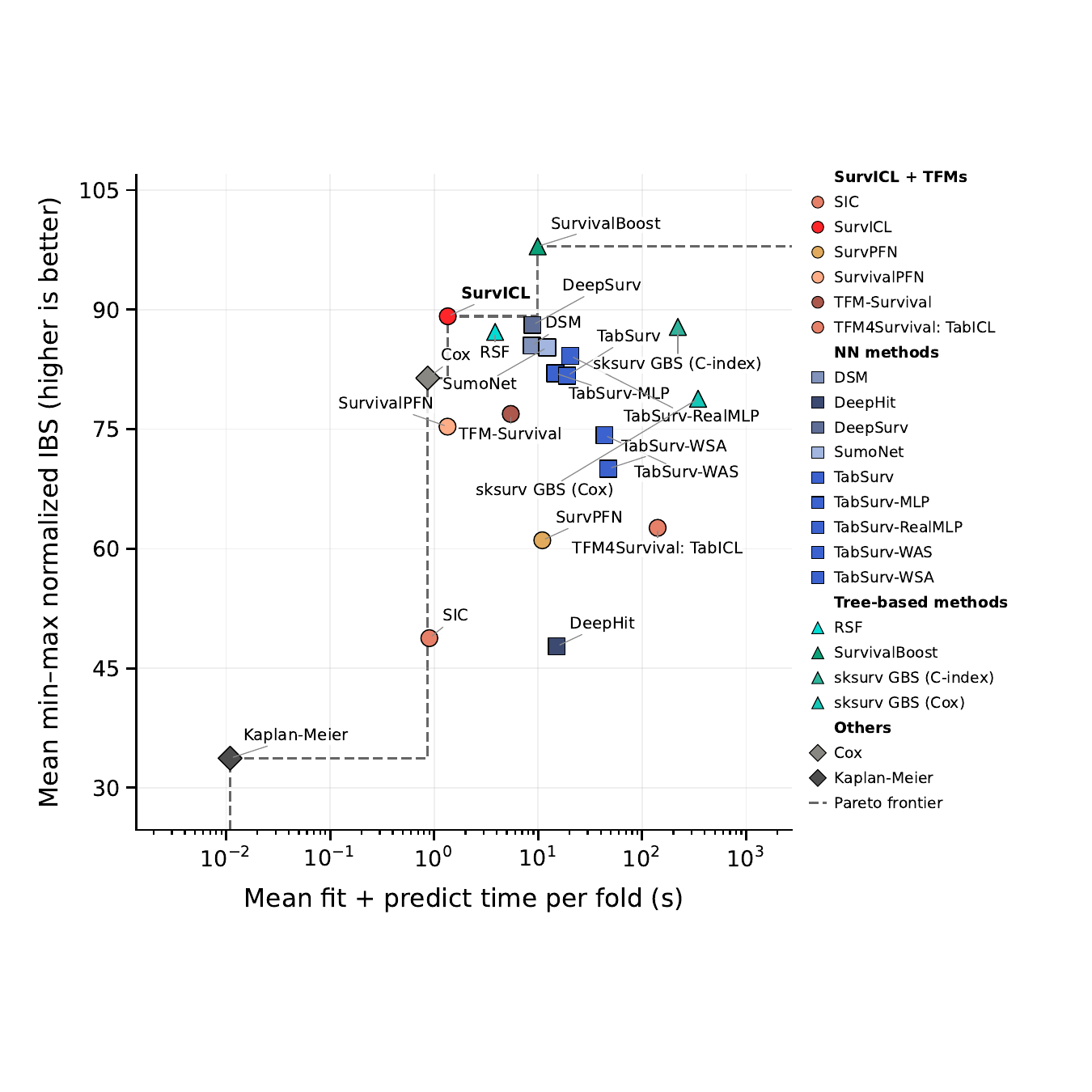}
    \caption{\textbf{Pareto front of mean normalized IBS (higher is better) and fit $+$ predict time.} We report average performances over the large held-out datasets. IBS was normalized to provide comparability across all datasets.}
    \label{fig:paretoibslarge}
\end{figure}

Results for the IBS on large held-out datasets are shown in Figures \ref{fig:boxplotibslarge}, \ref{fig:cdplotibslarge}, \ref{fig:winplotibslarge}, \ref{fig:paretoibslarge}.

\FloatBarrier
\newpage

\newpage
\subsubsection{C-index}

\begin{figure}[h!]
    \centering
    \includegraphics[width=0.5\linewidth]{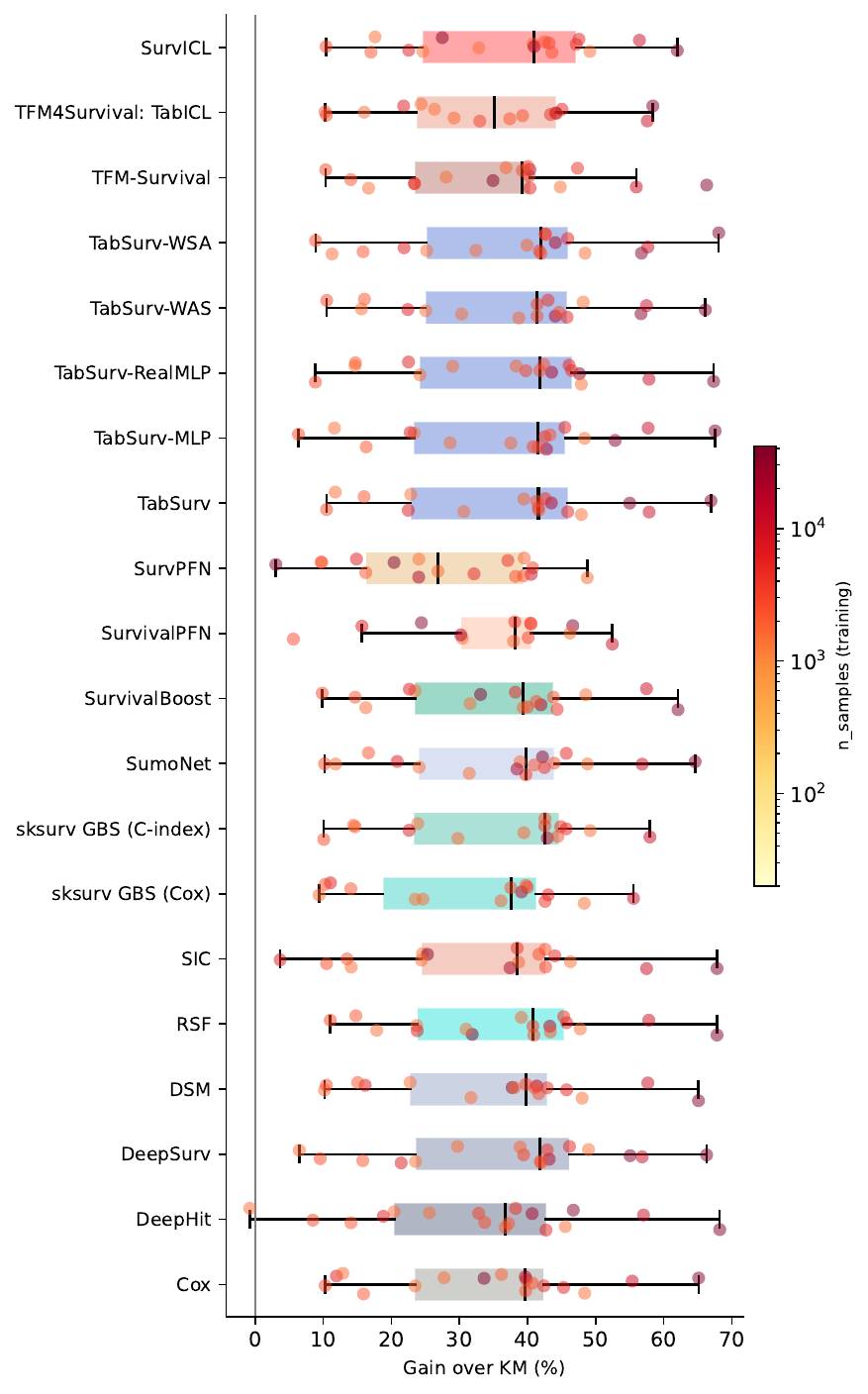}
    \caption{\textbf{C-index at the 50th quantile on the large held-out datasets, compared to the C-index obtained with a Kaplan-Meier estimator (higher is better).} The points show individual dataset performances, and a boxplot summarizes the overall distribution. The dot color represents the dataset size.}
    \label{fig:boxplotcindexlarge}
\end{figure}

\begin{figure}[h!]
    \centering
    \includegraphics[width=\linewidth]{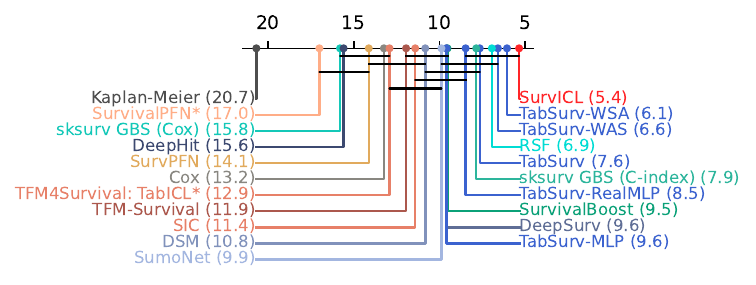}
    \caption{\textbf{Critical difference diagram of average ranks on the large held-out datasets (lower is better) for the C-index at the 50th quantile.} Methods joined by a bar are not significantly different.}
    \label{fig:cdplotcindexlarge}
\end{figure}

\begin{figure}[h!]
    \centering
    \includegraphics[width=\linewidth]{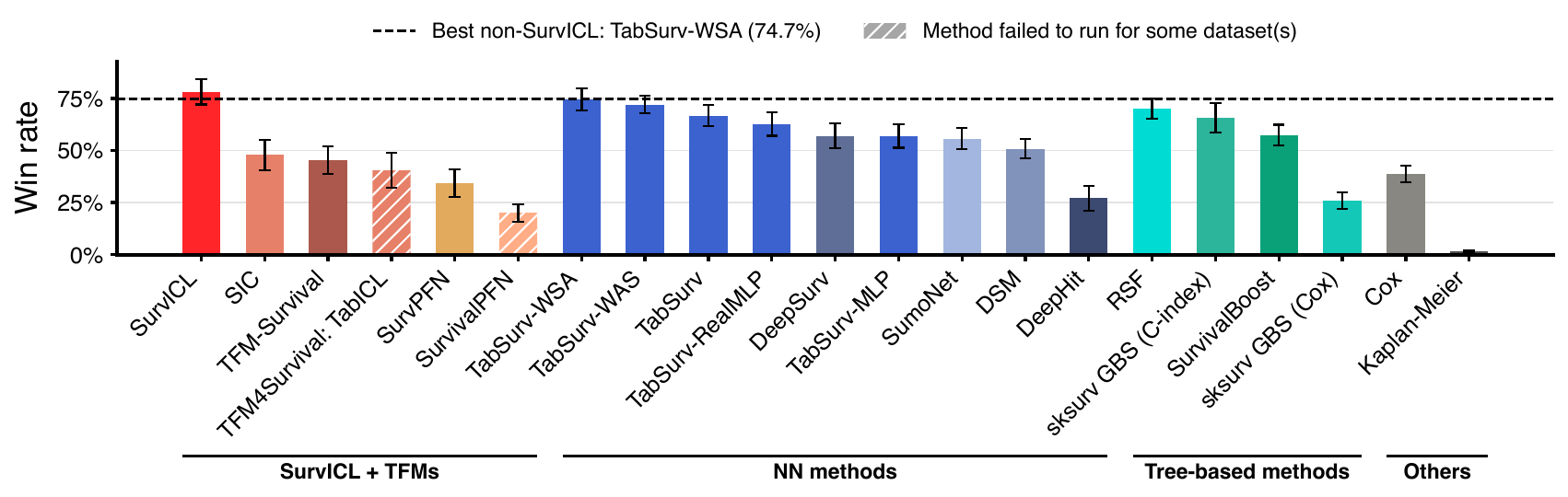}
    \caption{\textbf{Pairwise win rate in the C-index at the 50th quantile on the large held-out datasets:} the fraction of (dataset, competitor) pairs on which a method achieves a higher C-index.}
    \label{fig:winplotcindexlarge}
\end{figure}

\begin{figure}[h!]
    \centering
    \includegraphics[width=0.9\linewidth]{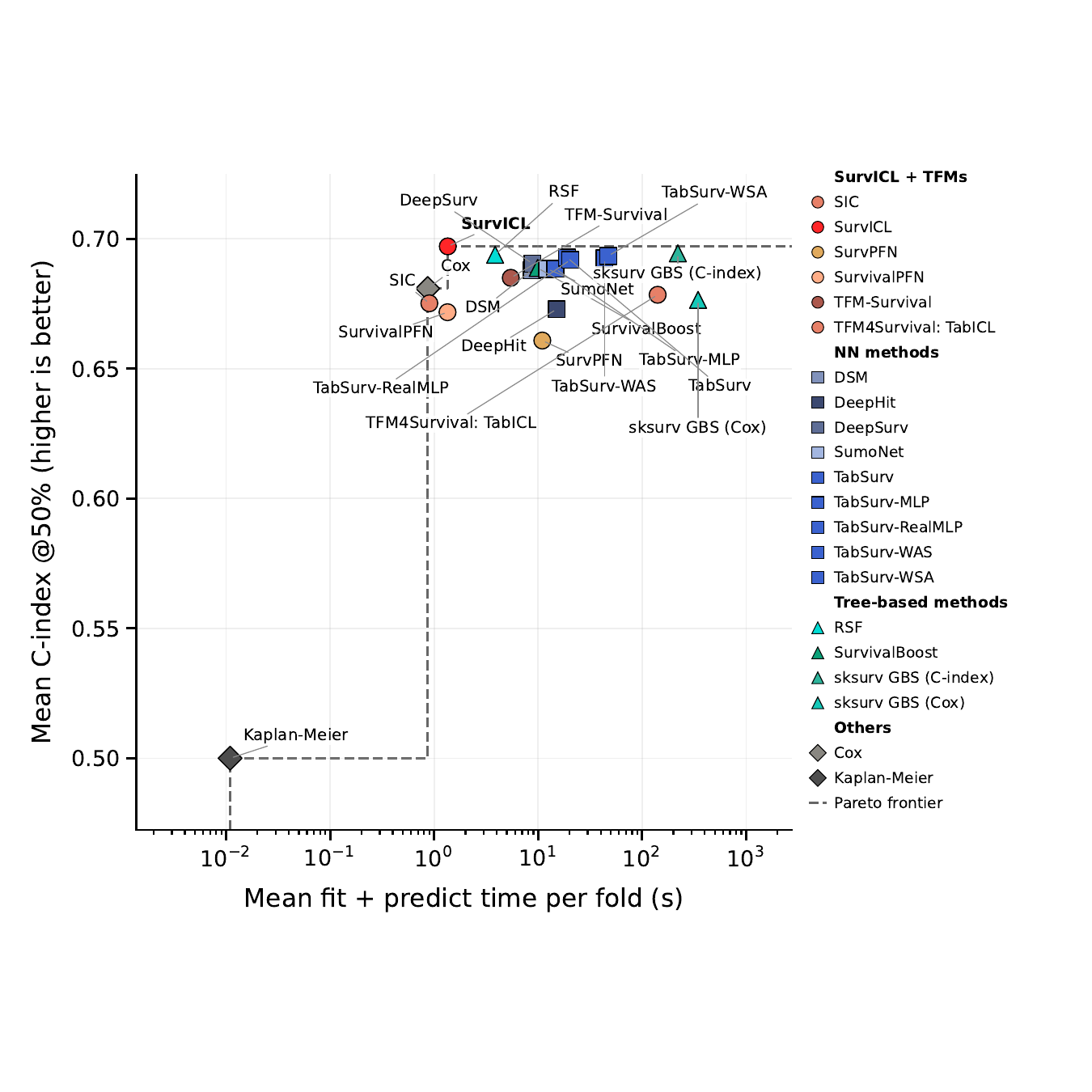}
    \caption{\textbf{Pareto front of mean  C-index on the 50th quantile (higher is better) and fit $+$ predict time.} We report average performances over the large held-out datasets.}
    \label{fig:paretocindexlarge}
\end{figure}

Results for the C-index on large held-out datasets are shown in Figures \ref{fig:boxplotcindexlarge}, \ref{fig:cdplotcindexlarge}, \ref{fig:winplotcindexlarge}, \ref{fig:paretocindexlarge}.

\FloatBarrier
\newpage

\newpage
\subsubsection{KM-calibration}

\begin{figure}[h!]
    \centering
    \includegraphics[width=0.5\linewidth]{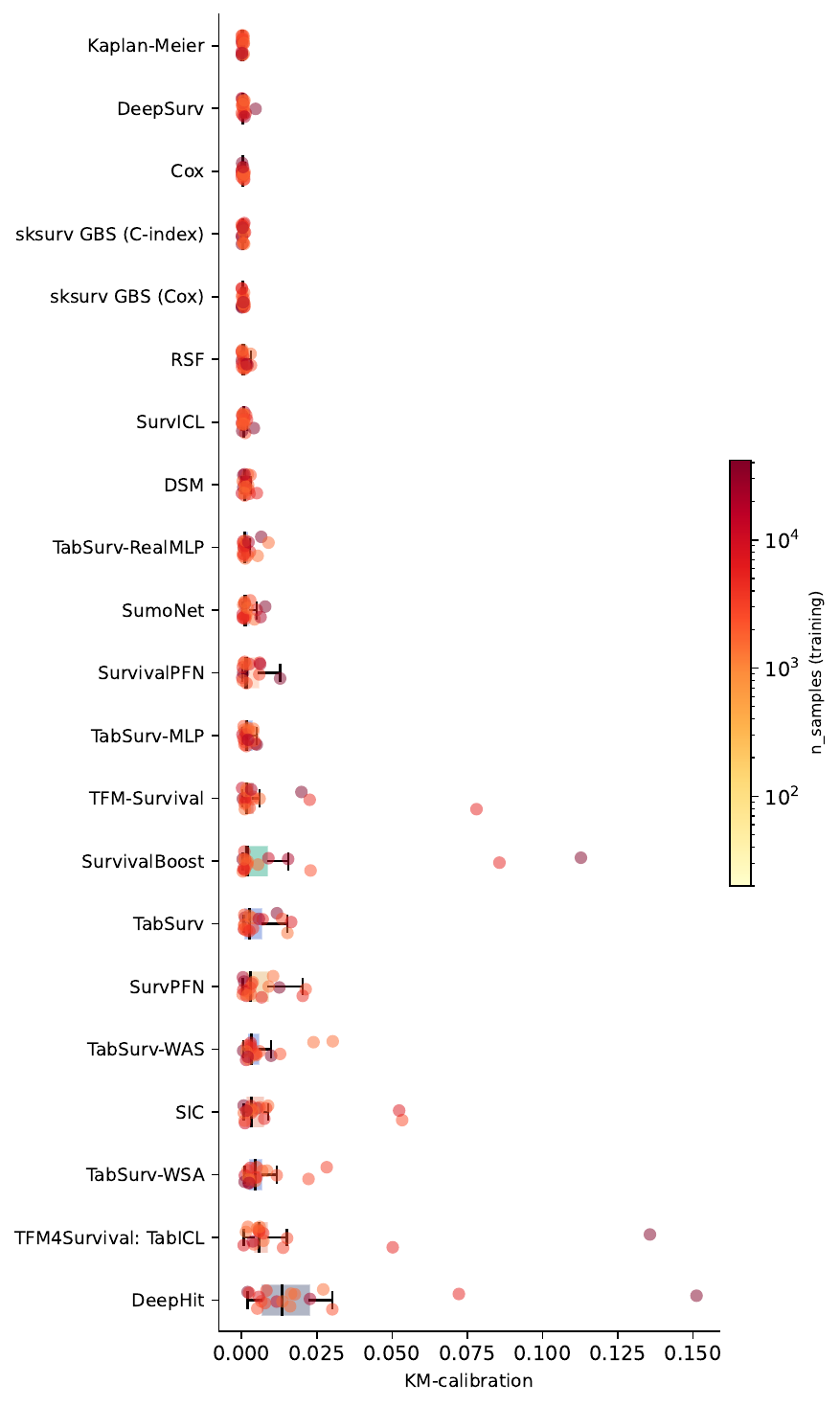}
    \caption{\textbf{KM calibration on the large held-out datasets (lower is better).} The points show individual dataset performances, and a boxplot summarizes the overall distribution. The dot color represents the dataset size.}
    \label{fig:boxplotkmcallarge}
\end{figure}

\begin{figure}[h!]
    \centering
    \includegraphics[width=\linewidth]{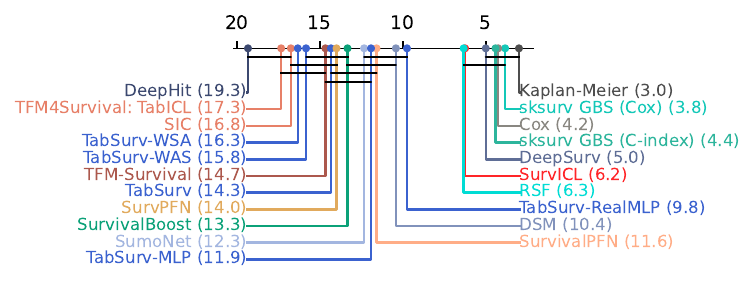}
    \caption{\textbf{Critical difference diagram of average ranks on the large held-out datasets (lower is better) for the KM calibration.} Methods joined by a bar are not significantly different.}
    \label{fig:cdplotkmcallarge}
\end{figure}

\begin{figure}[h!]
    \centering
    \includegraphics[width=\linewidth]{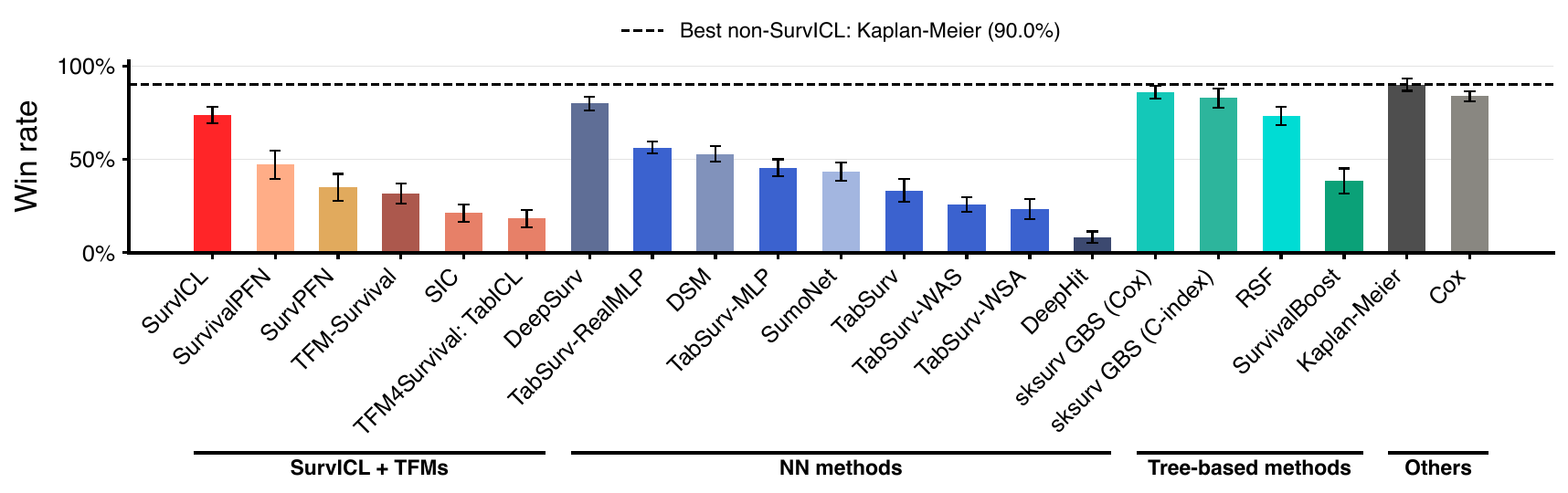}
    \caption{\textbf{Pairwise win rate in the KM calibration on the large held-out datasets:} the fraction of (dataset, competitor) pairs on which a method achieves a better KM calibration.}
    \label{fig:winplotkmcallarge}
\end{figure}

Results for the KM-calibration on large held-out datasets are shown in Figures \ref{fig:boxplotkmcallarge}, \ref{fig:cdplotkmcallarge}, \ref{fig:winplotkmcallarge}.

\FloatBarrier
\newpage

\newpage
\subsubsection{MAE}

\begin{figure}[h!]
    \centering
    \includegraphics[width=0.5\linewidth]{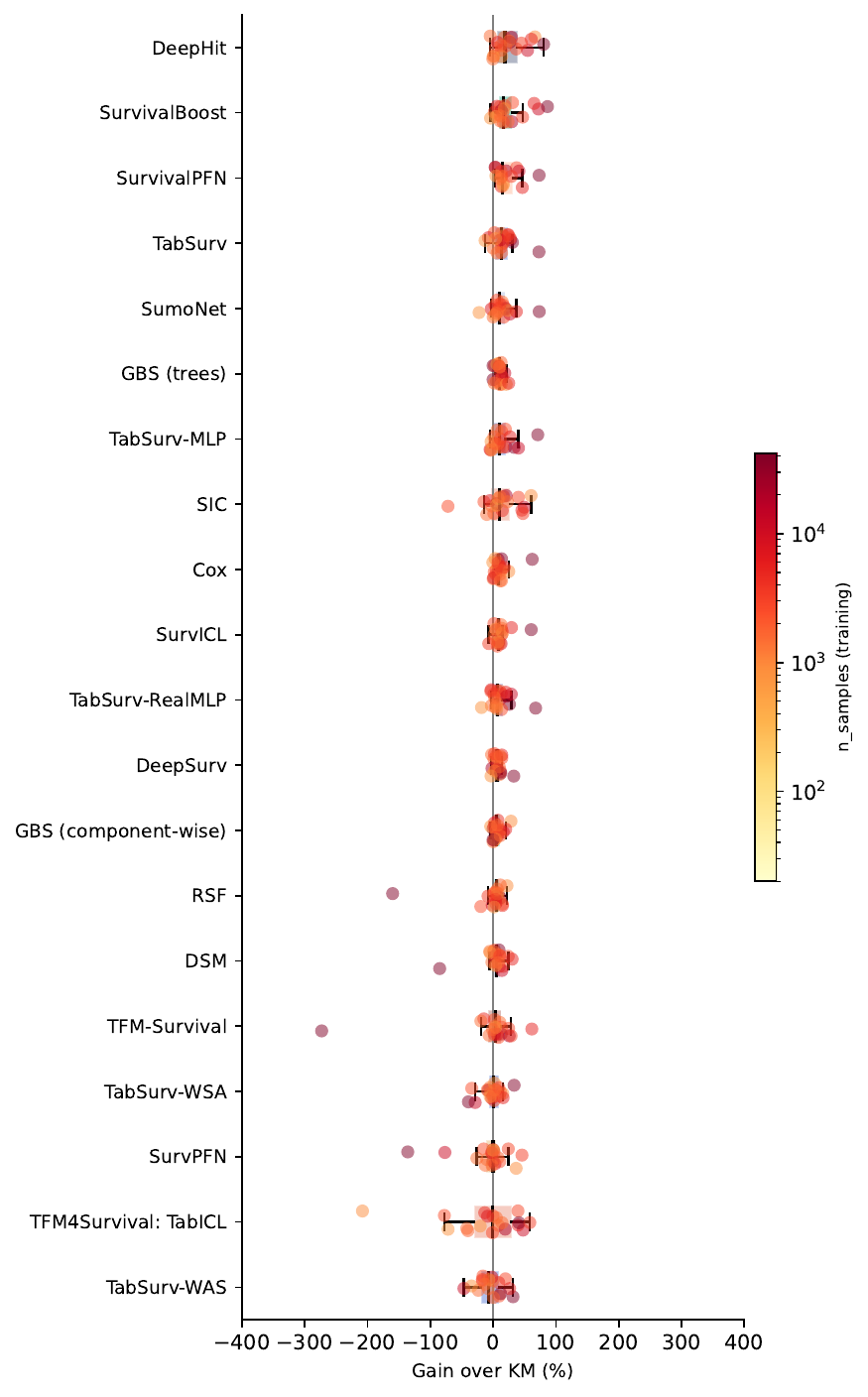}
    \caption{\textbf{MAE on the large held-out datasets, compared to the MAE obtained with a Kaplan-Meier estimator (higher is better).} The points show individual dataset performances, and a boxplot summarizes the overall distribution. The dot color represents the dataset size.}
    \label{fig:boxplotmaelarge}
\end{figure}

\begin{figure}[h!]
    \centering
    \includegraphics[width=\linewidth]{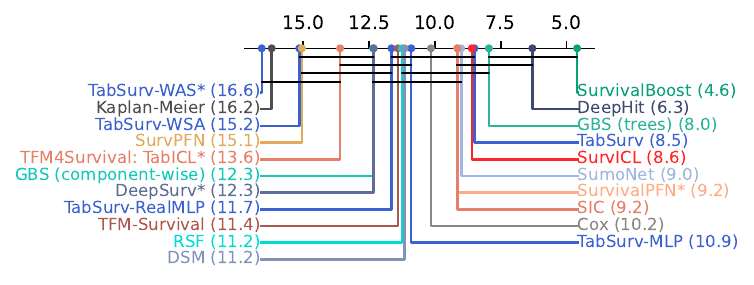}
    \caption{\textbf{Critical difference diagram of average ranks on the large held-out datasets (lower is better) for the MAE.} Methods joined by a bar are not significantly different.}
    \label{fig:cdplotmaelarge}
\end{figure}

\begin{figure}[h!]
    \centering
    \includegraphics[width=\linewidth]{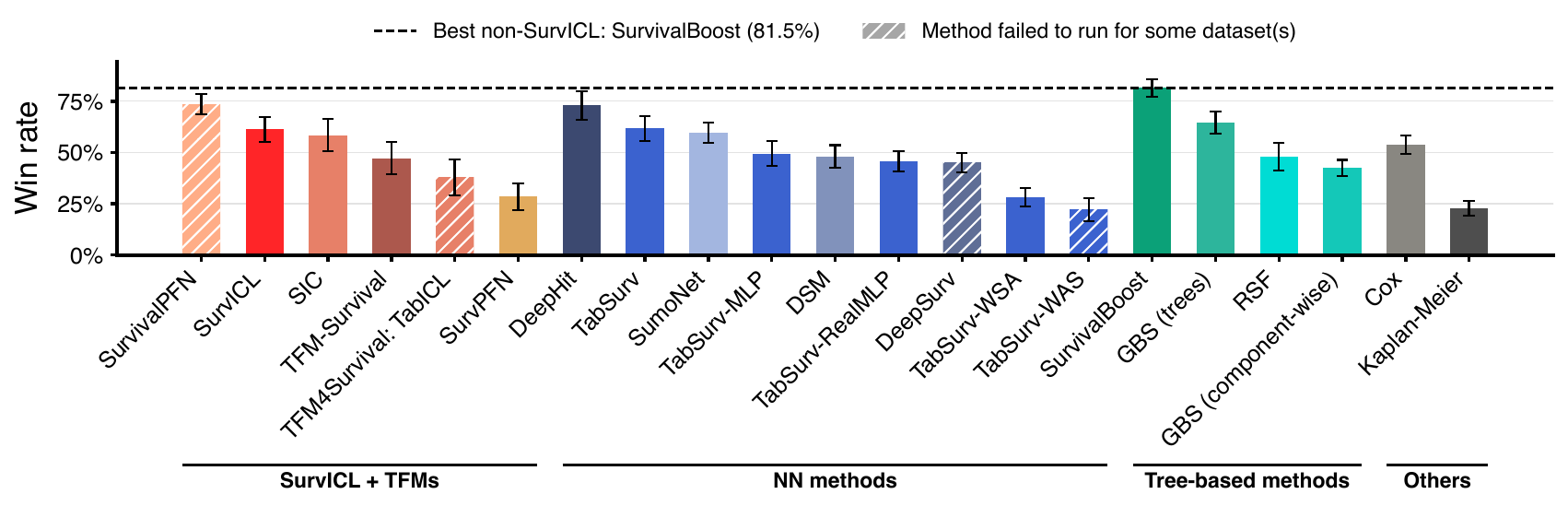}
    \caption{\textbf{Pairwise win rate in the MAE on the large held-out datasets:} the fraction of (dataset, competitor) pairs on which a method achieves a lower MAE.}
    \label{fig:winplotmaelarge}
\end{figure}

Results for the MAE on large held-out datasets are shown in Figures \ref{fig:boxplotmaelarge}, \ref{fig:cdplotmaelarge}, \ref{fig:winplotmaelarge}.

\begin{figure}[h!]
    \centering
    \includegraphics[width=0.9\linewidth]{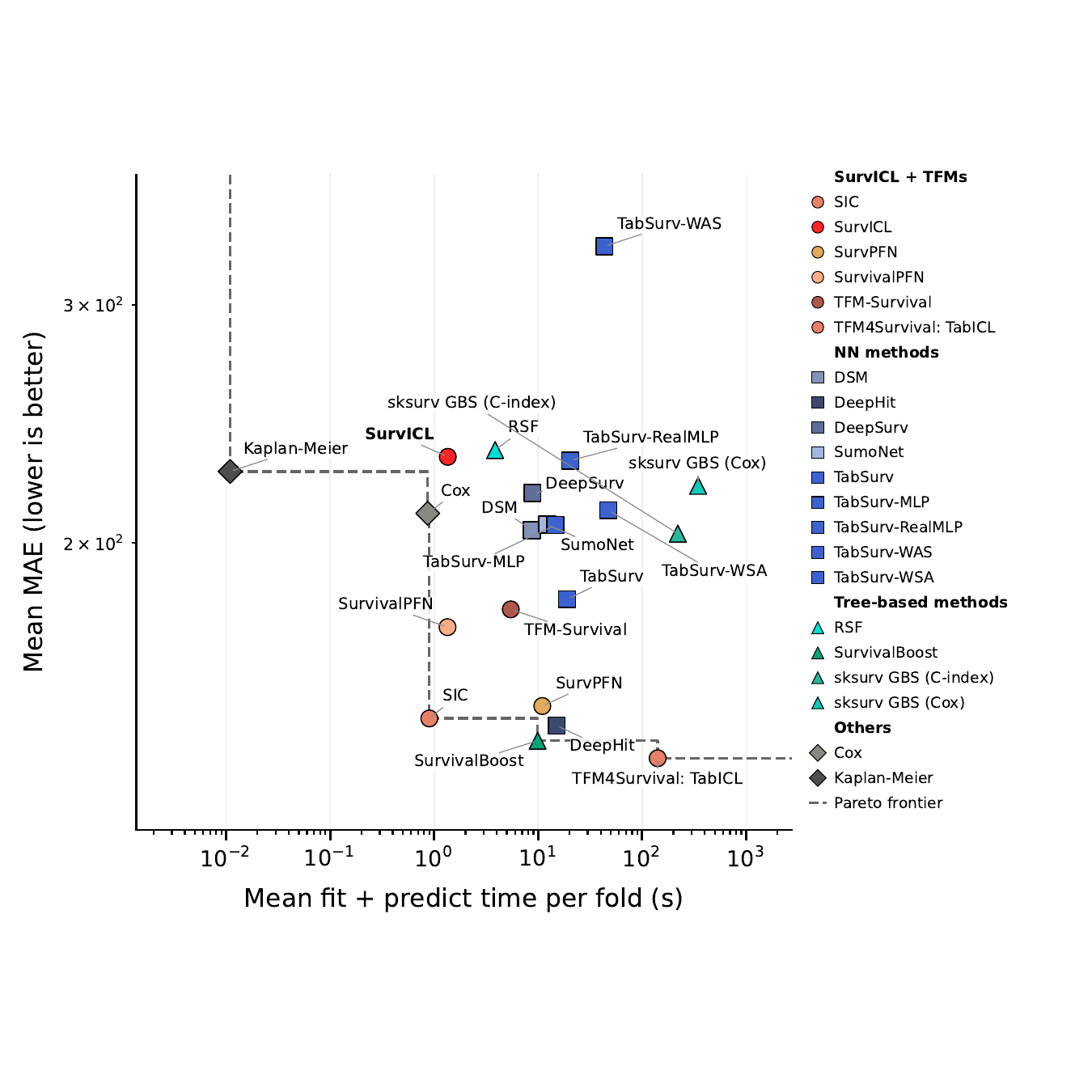}
    \caption{\textbf{Pareto front of mean MAE (lower is better) and fit $+$ predict time.} We report average performances over the large held-out datasets.}
    \label{fig:paretomaelarge}
\end{figure}

\end{document}